\documentclass[10pt,logo,copyright]{nvidiatechreport}
\usepackage[numbers,sort]{natbib}
\definecolor{citecolor}{HTML}{0071BC}
\definecolor{linkcolor}{HTML}{ED1C24}

\makeatletter
\@ifpackageloaded{hyperref}{}{
    \usepackage[pagebackref=false,breaklinks=true,colorlinks,citecolor=citecolor,linkcolor=linkcolor,bookmarks=false]{hyperref}
}
\makeatother

\usepackage{epsfig}
\usepackage{graphicx}
\usepackage{amsmath,mathtools}
\usepackage{amssymb}
\usepackage{amsthm}
\usepackage{nicefrac}       %
\usepackage{microtype}      %
\usepackage{amsmath} 
\usepackage{float}
\usepackage{placeins}
\usepackage{amsfonts}
\usepackage{bm}
\usepackage{enumitem}
\usepackage{mathtools}
\usepackage{multirow}
\usepackage{siunitx}

\usepackage{fontawesome}
\usepackage{pifont}
\usepackage{color}
\usepackage[algo2e,inoutnumbered,linesnumbered,algoruled,vlined]{algorithm2e}
\usepackage{algpseudocode}
\usepackage{tabularx}
\usepackage{url}
\usepackage{bbm}
\usepackage{threeparttable}
\usepackage{tikz}
\usepackage{wrapfig}
\usetikzlibrary{backgrounds}
\usetikzlibrary{decorations.pathreplacing}
\usetikzlibrary{arrows.meta,positioning,fit}
\usepackage[outline]{contour}
\usepackage{subcaption}
\usepackage{xcolor}
\usepackage{caption}
\usepackage{wrapfig}


\theoremstyle{definition}

\DeclarePairedDelimiterX{\infdivx}[2]{(}{)}{%
  #1\;\delimsize\|\;#2%
}

\definecolor{darkblue}{RGB}{49,130,189}
\definecolor{stanfordgrey}{RGB}{46,45,41}
\definecolor{cardinalred}{RGB}{253,141,60}

\makeatletter
\DeclareRobustCommand\onedot{\futurelet\@let@token\@onedot}
\def\@onedot{\ifx\@let@token.\else.\null\fi\xspace}

\makeatother

\usepackage[capitalize]{cleveref}

\crefname{section}{Sec.}{Secs.}
\Crefname{section}{Sec.}{Secs.}
\crefname{subsection}{Sec.}{Secs.}
\Crefname{subsection}{Sec.}{Secs.}
\crefname{subsubsection}{Sec.}{Secs.}
\Crefname{subsubsection}{Sec.}{Secs.}
\crefname{conj}{Conj.}{Conj.}

\Crefname{assumption}{\textbf{H}\hspace{-3pt}}{\textbf{H}\hspace{-3pt}}
\crefname{assumption}{\textbf{H}}{\textbf{H}}

\crefname{algorithm}{Alg.}{Alg.}
\crefname{assumption}{\textbf{H}}{\textbf{H}}
\crefname{equation}{Eq.}{Eqs.}
\Crefname{equation}{Eq.}{Eqs.}
\crefname{definition}{Dfn.}{Dfn.}
\crefname{lemma}{Lemma}{Lemma}
\crefname{dfn}{Dfn.}{Dfn.}
\crefname{thm}{Thm.}{Thm.}
\crefname{table}{Tab.}{Tabs.}
\Crefname{table}{Tab.}{Tabs.}
\crefname{figure}{Fig.}{Figs.}
\Crefname{figure}{Fig.}{Figs.}

\definecolor{mygreen}{RGB}{159, 200, 59}
\definecolor{myred}{RGB}{223, 135, 102}

\newcommand{\MethodName}{\textit{Asset Harvester}\xspace}

\definecolor{policy_green}{rgb}{0.2, 0.59, 0.2}
\definecolor{policy_yellow}{rgb}{0.8, 0.8, 0.2}

\usepackage{amsmath,amsfonts,bm,mathtools}

\def\eqref#1{\cref{#1}}

\def\1{\bm{1}}

\DeclareMathAlphabet{\mathsfit}{\encodingdefault}{\sfdefault}{m}{sl}
\SetMathAlphabet{\mathsfit}{bold}{\encodingdefault}{\sfdefault}{bx}{n}

\def\@onedot{\ifx\@let@token.\else.\null\fi\xspace}

\renewcommand{\MethodName}{\textit{Foundation4D}\xspace}

\newcommand{\method}{\MethodName}
\definecolor{draftred}{HTML}{B42318}
\definecolor{sauronblue}{HTML}{1F5AA6}
\definecolor{sauronteal}{HTML}{168A8A}
\definecolor{sauronorange}{HTML}{D97706}

\title{Foundation4D: Towards a unified perception and reconstruction model for autonomous vehicles}
\author{NVIDIA\textsuperscript{\ensuremath{\ddagger}}}

\begin{document}
\maketitle
\begingroup
\renewcommand{\thefootnote}{\ensuremath{\ddagger}}
\footnotetext{A full list of authors can be found in the Appendix \ref{AppA}. }
\endgroup

\vspace{-12pt}
\begin{abstract}

We introduce \method, a unified feed-forward model for geometry estimation, 3D detection, semantic occupancy, and Gaussian reconstruction from multi-camera driving video. A shared backbone uses factorized spatiotemporal attention to efficiently combine observations across cameras and time. Our video inference pipeline uses pose graph optimization to refine camera trajectories and assembles object tracks across input windows. We also develop a data annotation pipeline and release PAI--Foundation4D, a benchmark with dense depth and semantic occupancy labels for 1,253 driving clips. With benchmark-specific fine-tuning, \method achieves the lowest depth AbsRel on Waymo and nuScenes, the highest detection mAP on nuScenes, and the highest occupancy mIoU on OpenOccupancy among the compared methods. Our task-combination study shows benefits for both detection and geometry: co-training detection with geometry and occupancy improves top-class mAP by 1.2 points over detection-only training, while co-training geometry with detection reduces camera-to-rig translation error by $13.4\%$ relative to geometry-only training. We further use the predicted geometry and trajectories as priors for lidarless NuRec, supporting reconstruction of driving scenes for simulation.
\end{abstract}

{\hbadness=10000\abscontent\par}

\begingroup
\setlength{\intextsep}{6pt}
\captionsetup{font=footnotesize,skip=2pt}
\begin{center}
\includegraphics[width=\textwidth,trim=0 8mm 0 0,clip]{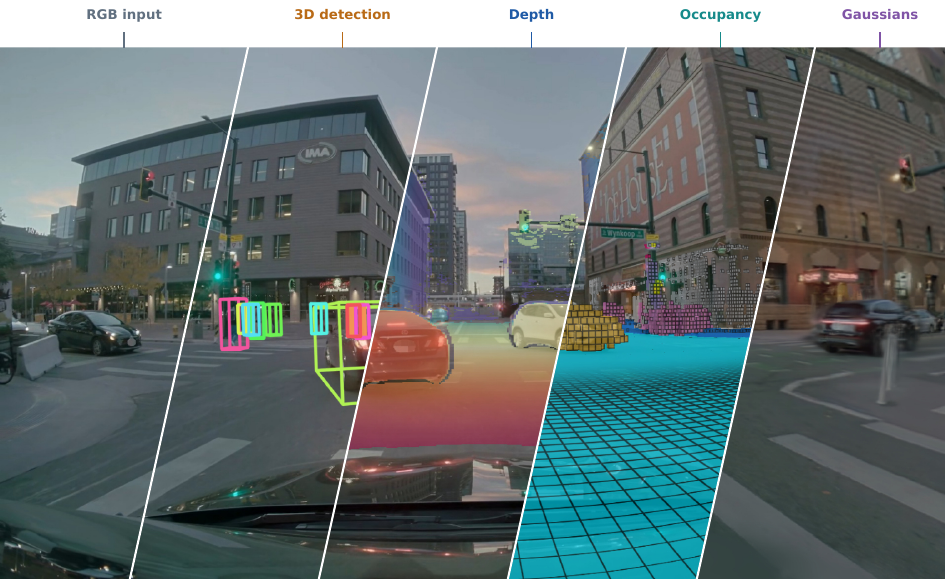}
\captionof{figure}{\textbf{From camera video to complementary scene representations.} A shared view illustrates RGB input, 3D detection, depth, semantic occupancy, and Gaussian rendering. %
}
\label{fig:teaser}
\end{center}
\endgroup

\clearpage
\begingroup
\setcounter{tocdepth}{2}
\renewcommand{\baselinestretch}{1}\small
\setlength{\parskip}{0pt}
\tableofcontents
\endgroup
\clearpage

\section{Introduction}
\label{sec:introduction}

Two central goals in the development of autonomous vehicle systems are understanding the driving environment and reconstructing recorded scenes for simulation. Both require recovering scene geometry, objects, their motion, and the space they occupy. Unified perception models learn these capabilities through shared visual representations, avoiding the repeated processing of the same observations by separate networks. UniAD and PARA-Drive connect perception with prediction and planning~\cite{hu2023uniad,weng2024paradrive}, while GeoUP jointly predicts depth, 3D objects, and semantic occupancy from driving video~\cite{xu2026geoup}. These outputs capture much of the structure and motion needed for scene reconstruction. Simulation additionally requires recovering scene appearance so that the environment and its moving actors can be rendered from novel viewpoints. Unifying perception and reconstruction would therefore enable a shared model to both understand real driving scenes and recreate them for simulation.

Recent progress in feed-forward reconstruction offers a basis for this integration. VGGT~\cite{wang2025vggt} jointly recovers camera parameters and dense geometry from multiple images, while Gaussian reconstruction models such as Instant NuRec~\cite{nvidia2026instantnurec} predict renderable driving scenes without per-scene optimization. These advances make it possible to infer scene structure and appearance directly from images, but extending them to a shared model for multi-camera video requires efficient use of observations across views and time. In particular, the global attention used in VGGT grows quadratically with the total number of image tokens, making longer sequences costly to process.

We present \method, a feed-forward model for driving perception and reconstruction. A shared backbone extracts features from multi-camera video, and task-specific decoders predict camera poses and metric depth, 3D objects, semantic occupancy, and Gaussian attributes in a single pass over an input window. Factorized spatiotemporal attention reduces the cost of combining information across cameras and time. To use these predictions over a complete driving recording, geometric estimates must remain aligned and object identities must be maintained across windows. Our video-level inference pipeline assembles the window predictions into scene geometry and camera and object trajectories over long sequences. These outputs support Gaussian scene reconstruction and also provide priors for lidarless NuRec~\cite{nvidia2026nurec}, demonstrating their use in a downstream neural reconstruction workflow.

Training and evaluating these capabilities requires dense annotations that remain aligned with the changing
scene. Sparse LiDAR observations can be combined to improve surface coverage, but vehicle and object motion
make it difficult to align these measurements with camera images. Asynchronous sensing and rolling-shutter
exposure further complicate this alignment. We provide a pipeline for the creation of accurate depth images and semantic occupancy. 
We release the PAI–Foundation4D benchmark~\cite{nvidia2026physicalai-av-f4d}, 1,253 clips of the Physical AI Autonomous Vehicles (PAI) dataset~\cite{nvidia2025physicalai-av} that support geometry, detection, and occupancy evaluation on the same driving scenes. By providing
dense annotations together with documented evaluation protocols, the benchmark gives the community a
common basis for comparing methods and measuring progress across these tasks.

We evaluate \method on established public benchmarks and PAI--Foundation4D. With benchmark-specific fine-tuning, it achieves the lowest depth error on Waymo~\cite{sun2020waymo} and nuScenes~\cite{caesar2020nuscenes}, the highest detection accuracy on nuScenes, and the highest semantic occupancy accuracy on OpenOccupancy~\cite{wang2023openoccupancy} among the compared methods. Our evaluations show that \method can be adapted effectively across driving datasets. Longer temporal context improves detection without retraining, while pose graph optimization (PGO) with a motion prior improves camera trajectory accuracy over sequential stitching on PAI (\cref{tab:detection_temporal_context,tab:geometry_pai_benchmark}). Our task-combination study also reveals benefits and trade-offs of joint perception learning, with gains that depend on the tasks being combined.

Our reconstruction experiments show that using predicted ego poses preserves overall rendering quality close to that obtained with ground-truth ego poses (\cref{tab:gaussian_rgb}). The predicted geometry and actor trajectories also provide useful priors for lidarless NuRec~\cite{nvidia2026nurec}, demonstrating their practical value in reconstructing driving scenes for simulation.

Our contributions are:
\begin{itemize}
    \item \method, a feed-forward perception and reconstruction model that shares multi-camera video features across task-specific decoders for geometry, semantics, objects, occupancy, and Gaussian reconstruction.
    \item A video-level inference pipeline that aligns window-level geometry and assembles object trajectories, improving trajectory estimation over long driving sequences.
    \item A data annotation pipeline that accounts for sensor timing, ego and actor motion, and visibility to generate depth and semantic occupancy labels from driving recordings.
    \item The PAI--Foundation4D benchmark, with dense annotations and evaluation protocols for comparing geometry, detection, and occupancy methods on a common set of driving scenes.
\end{itemize}

\section{Related work}
\label{sec:related}

\method brings geometric recovery, semantic perception, and Gaussian reconstruction into a shared model for multi-camera driving video. It builds on two closely related research directions: multi-task learning for autonomous driving, perception and reconstruction tasks.

\subsection{Multi-task learning for autonomous driving}
\label{sec:related_shared}
Multi-task learning allows driving models to share features across different descriptions of the same scene. PnPNet~\cite{liang2020pnpnet} and ViP3D~\cite{gu2023vip3d} jointly model detection, tracking, and motion forecasting, while BEVFormer~\cite{li2022bevformer} and PETRv2~\cite{liu2023petrv2} explore shared representations for detection and segmentation. UniVision~\cite{hong2024univision} jointly learns 3D detection and occupancy through interacting BEV and voxel features. Planning-oriented approaches further connect scene understanding to driving decisions~\cite{casas2021mp3,hu2022stp3,jiang2023vad}. UniAD~\cite{hu2023uniad} passes information between perception, prediction, and planning through task queries; PARA-Drive~\cite{weng2024paradrive} studies these dependencies and co-trains parallel task heads through shared BEV features. The benefits of joint learning depend on which tasks are combined~\cite{standley2020tasks}, making task interactions an important consideration alongside the range of outputs a model supports.

Pretraining offers a complementary route to representations that support multiple tasks. UniPAD~\cite{yang2024unipad} learns transferable features through volumetric rendering with image and depth reconstruction objectives, while DriveWorld~\cite{min2024driveworld} uses temporal scene and action prediction for downstream adaptation. Geometric foundation models such as VGGT~\cite{wang2025vggt} learn multi-image features for recovering cameras and dense scene structure. These reconstruction objectives provide a geometric basis for extending a shared backbone to driving perception.

Recent driving models extend geometric representations to perception and planning. DVGT-2~\cite{zuo2026dvgt2} couples streaming geometry reconstruction with trajectory planning. Closely related to our approach, GeoUP~\cite{xu2026geoup} adapts VGGT to calibrated multi-camera video through factorized temporal and cross-view attention, conditioning its features on known camera intrinsics and poses. It jointly predicts metric depth, 3D objects, and semantic occupancy, with auxiliary camera-pose supervision and multi-dataset training that accommodates missing task annotations. USR-Drive~\cite{chen2026usrdrive} connects object detection and renderable reconstruction by jointly denoising 3D box and Gaussian latents from multi-view video with known camera poses. \method combines geometric and semantic perception with Gaussian reconstruction in a shared feed-forward model, while explicitly estimating ego motion and camera-to-rig calibration. We train the shared model on large-scale, partially annotated driving data and examine transfer and interference between geometry, detection, and occupancy.

\subsection{Perception and reconstruction tasks}
\label{sec:related_tasks}

\paragraph{Geometric reconstruction.}
Feed-forward geometry models learn to recover 3D structure and camera geometry from images. DUSt3R and MASt3R establish pointmap- and correspondence-based reconstruction~\cite{wang2024dust3r,leroy2024mast3r}, while VGGT~\cite{wang2025vggt} jointly predicts cameras and dense geometry from multiple images. Related models explore alternative geometric parameterizations~\cite{wang2026pi3,lin2026depthanything3}, optional camera and depth priors~\cite{jang2025pow3r,keetha2026mapanything}, and efficient multi-image inference~\cite{yang2025fast3r,wang2026vggtomega,burzio2026dejaview}. Temporal extensions maintain scene state or cached features for incremental reconstruction~\cite{wang2025cut3r,zhuo2026streamvggt,chen2026lingbotmap}, while dynamic models recover time-varying structure and point trajectories~\cite{zhang2026d4rt,luo20264rc,jiang2026omnix4d}.

Driving adaptations exploit the fixed relationships between vehicle-mounted cameras and the motion of the rig. DVGT~\cite{zuo2026dvgt} predicts metric pointmaps and ego poses from multi-camera images without supplied calibration. Rig3R~\cite{li2025rig3r} supports optional rig metadata and inference of missing camera geometry. With calibration available, DriveVGGT~\cite{jia2026drivevggt} uses calibration-guided attention and factorized heads for rig geometry and ego motion. TRIG~\cite{liao2026trig} instead separates trajectory and rig priors in feature encoding and supervision while retaining the VGGT prediction heads. LiAuto-GeoX~\cite{lian2026liautogeox} addresses inference efficiency by distilling dense geometry and cross-view relations into a compact student.

Scaling reconstruction to long videos requires managing memory and accumulated drift. DVGT-2~\cite{zuo2026dvgt2} carries temporal information through a sliding-window cache, while Drive3R~\cite{yang2026drive3r} combines block-wise causal attention with confidence-aware re-anchoring under known camera calibration. Complementary systems assemble local reconstructions through geometric alignment and global optimization. VGGT-Long~\cite{deng2026vggtlong} aligns overlapping chunks with confidence-weighted similarity transforms and uses loop constraints to correct drift. VGGT-SLAM~\cite{maggio2025vggtslam} optimizes submap alignments on $\mathrm{SL}(4)$ to account for projective ambiguity. VGGT-SLAM~2.0~\cite{maggio2026vggtslam2} refines keyframe-level geometry through constraints within and between submaps, and uses VGGT attention to verify loop closures. \method adapts this local-to-global strategy to driving video, where multiple cameras move as one rig but may capture images at different times. We separate camera-to-rig calibration from ego motion so that overlapping windows can constrain a shared trajectory, whose corrections are propagated to each camera at its own capture time (\cref{sec:inference_system}).

\paragraph{3D Object detection and tracking.}
Camera-based 3D object detection localizes and classifies surrounding objects as oriented 3D cuboids from multi-camera images, while multi-object tracking links these detections into identity-consistent trajectories over time.
Query-based detectors build on the set prediction of DETR~\cite{carion2020detr}, decoding a sparse set of object queries that are matched one-to-one to the ground truth without anchors or non-maximum suppression.
They differ mainly in how queries gather multi-view image features.
Dense-attention methods make image features 3D-aware for global cross-attention. PETR~\cite{liu2022petr} adds position embeddings computed from camera frustums, while 3DPPE~\cite{shu20233dppe} replaces them with 3D point embeddings derived from predicted depth.
Sparse methods instead sample features around 3D anchors or BEV pillar queries across views and frames~\cite{lin2022sparse4d,liu2023sparsebev}.
Temporal fusion further improves localization and velocity estimation. PETRv2~\cite{liu2023petrv2} aligns past-frame coordinates with ego motion, StreamPETR~\cite{wang2023exploring} propagates object queries through a memory queue for efficient streaming detection, Stream3DPPE~\cite{shu20233dppe} and RoPETR~\cite{ji2025ropetr} extend StreamPETR with point embeddings and spatio-temporal rotary position embeddings, respectively.
RayDN~\cite{liu2024ray} improves training by extending query denoising~\cite{li2022dndetr} with hard negatives sampled along camera rays, which suppresses duplicate detections at wrong depths.
These detectors aggregate views and frames inside the detection head, using ego poses to align past frames.
GeoUP~\cite{xu2026geoup} instead fuses them in a VGGT-initialized backbone conditioned on camera calibration and ego poses (\cref{sec:related_shared}) and decodes boxes with a RayDN-style head, reporting state-of-the-art nuScenes results.
\method likewise aggregates images from all cameras across a temporal window in its geometry-pretrained backbone and decodes PETR-style queries into 3D boxes and velocities through deformable attention~\cite{zhu2021deformable}.
Unlike GeoUP, which requires ground-truth ego poses, \method predicts them directly from multi-view images.

Tracking methods aim to obtain temporally consistent object trajectories and differ in how they associate detections over time.
Following the tracking by detection paradigm, SORT~\cite{bewley2016sort} combines Kalman-filter motion prediction with Hungarian matching, AB3DMOT~\cite{weng2020ab3dmot} extends this design to 3D boxes, and CenterPoint~\cite{yin2021centerpoint} matches detections using their predicted velocities.
Query-based trackers such as TrackFormer~\cite{meinhardt2022trackformer} and MOTR~\cite{zeng2022motr} instead propagate track queries from frame to frame, an approach that MUTR3D~\cite{zhang2022mutr3d} and PF-Track~\cite{pang2023standing} bring to multi-camera 3D tracking.
These online trackers rely on past observations, whereas offline methods optimize association over many frames at once. MPNTrack~\cite{braso2020learning} learns it with message passing on a detection graph. SUSHI~\cite{cetintas2023sushi} scales learned association to long clips through a hierarchy of graphs over increasingly long time spans. \method takes inspiration from the offline literature, but adopts a simple yet effective association mechanism from online methods. It utilizes Kalman-filter consuming velocities predicted by our detection head, and uses and Hungarian association for frame-by frame association, similar to SORT and AB3DMOT. However, it processes the clip offline, reconnecting fragmented tracks in a second pass and refining each trajectory with a Rauch--Tung--Striebel (RTS) smoother that also uses later observations.

\paragraph{Semantic occupancy.}

Semantic occupancy prediction tackles scene geometry beyond visible surfaces, completing the surrounding geometry and assigning semantic labels to occupied regions. Existing approaches can be grouped according to their latent scene representations into \textit{plane-based}, \textit{dense volumetric}, and \textit{sparse} methods, with hybrid methods connecting these families. Plane-based approaches encode scenes using two-dimensional feature grids. BEVFormer~\cite{li2022bevformer} learns BEV features through spatiotemporal attention, while TPVFormer~\cite{huang2023tpvformer} uses three orthogonal feature planes. SelfOcc~\cite{huang2024selfocc} investigates BEV and TPV representations under self-supervised training, and POP-3D~\cite{vobecky2023pop3d} uses TPV features for open-vocabulary occupancy prediction. Dense volumetric approaches maintain features on regular 3D voxel grids. MonoScene~\cite{cao2022monoscene} and SurroundOcc~\cite{wei2023surroundocc} use 3D convolutions, OccFormer~\cite{zhang2023occformer} uses a dual-path transformer, and OccMamba~\cite{li2025occmamba} applies state space models. OccGen~\cite{wang2024occgen} uses diffusion-based refinement, PanoOcc~\cite{wang2024panoocc} aggregates multi-view and temporal features through voxel queries, and COTR~\cite{ma2024cotr} learns compact geometry-aware volumetric features. Allocating features to empty regions introduces memory and computational costs, motivating sparse alternatives. SparseOcc~\cite{tang2024sparseocc} operates on sparse voxel tensors. OPUS~\cite{wang2024opus} predicts occupied locations and semantic classes directly from learned queries, SPSC~\cite{guo2026spsc} progressively prunes unoccupied queries during coarse-to-fine refinement, and GaussianFormer~\cite{huang2024gaussianformer} represents scenes with sparse semantic Gaussian primitives. Hybrid designs combine these representation choices. VoxFormer~\cite{li2023voxformer} expands depth-guided sparse voxel queries into a dense volume, while CONet~\cite{wang2023openoccupancy} refines coarse occupied voxels at higher resolution. Symphonies~\cite{jiang2024symphonies} connects contextual instance queries with image and voxel features, and CGFormer~\cite{yu2024cgformer} combines voxel and tri-perspective features. \method adopts a BEV representation for semantic occupancy at 20$\,\mathrm{cm}$ resolution and up to 100$\,\mathrm{m}$ ahead of the ego vehicle. This compact BEV representation limits memory use, allowing us to support semantic occupancy alongside the other output tasks within a unified multi-camera, multi-timestamp model.

\paragraph{Gaussian scene reconstruction.}
3D Gaussian Splatting (3DGS)~\cite{kerbl2023gaussiansplatting} represents a scene with explicit anisotropic primitives for novel-view rendering. Feed-forward methods learn to predict these primitives across scenes, reducing the need to optimize each new scene separately. Representative approaches include pixelSplat~\cite{charatan2024pixelsplat} and MVSplat~\cite{chen2024mvsplat} for sparse-view reconstruction, and GS-LRM~\cite{zhang2024gslrm} for transformer-based Gaussian prediction. DepthSplat~\cite{xu2025depthsplat} connects multi-view depth estimation with Gaussian reconstruction, while NoPoSplat~\cite{ye2025noposplat} predicts Gaussians from unposed images. Geometry foundation models provide a further route to feed-forward rendering: AnySplat~\cite{jiang2025anysplat} extends a VGGT-based geometry transformer with Gaussian prediction and distills camera and geometry priors from pretrained VGGT. TokenGS~\cite{ren2026tokengs} instead uses learnable Gaussian tokens to decouple primitive prediction from the input pixel grid.

Driving scenes additionally require wide camera coverage and a representation of moving objects. DrivingGaussian~\cite{zhou2024drivinggaussian}, Street Gaussians~\cite{yan2024streetgaussians}, and OmniRe~\cite{chen2025omnire} reconstruct static backgrounds and dynamic actors through per-sequence optimization. Feed-forward approaches such as STORM~\cite{yang2025storm} and DrivingForward~\cite{tian2025drivingforward} address sparse temporal observations and surround-view inputs, while DGGT~\cite{chen2026dggt} jointly estimates cameras, Gaussians, and scene motion. Camera modeling is also important for driving capture: 3DGUT~\cite{wu2025dgut} supports nonlinear camera projection and rolling shutter. Following the layered scene representation of Instant NuRec~\cite{nvidia2026instantnurec}, \method integrates Gaussian reconstruction with geometry, detection, and occupancy through shared features and joint training. At inference, our Gaussian branch consumes camera poses derived from the model's predicted ego motion and actor tracks assembled from its detections, rather than requiring ground-truth poses or tracks (\cref{sec:gaussian_decoder}).

\section{Foundation4D}
\label{sec:method}

\begin{figure*}[!tp]
\centering
\includegraphics[width=\textwidth]{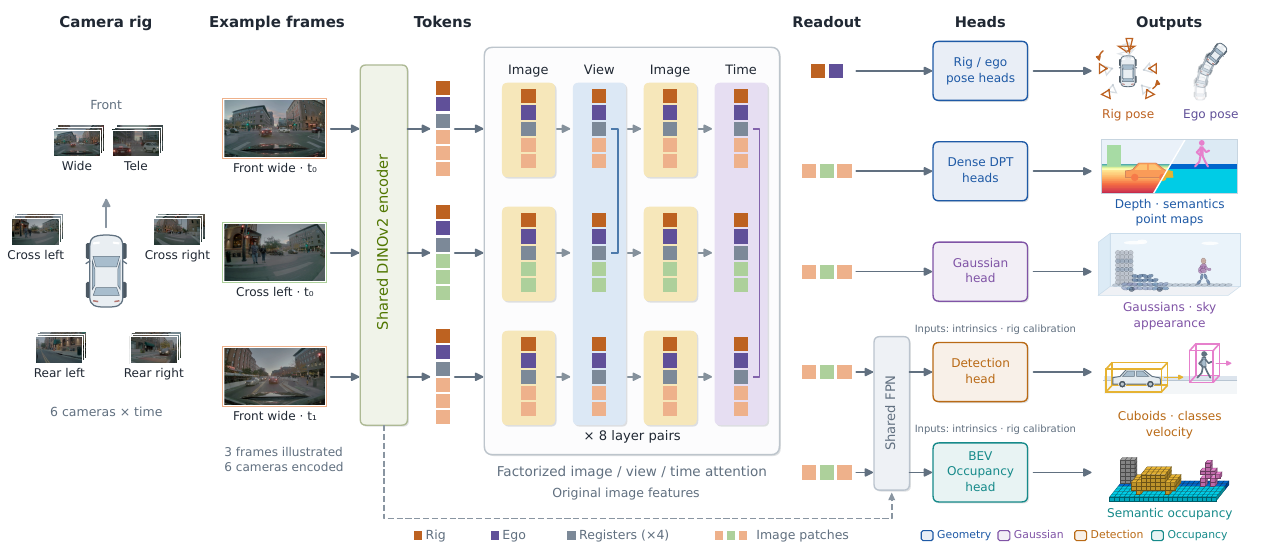}
\caption{\textbf{\method architecture.} \method unifies driving perception and reconstruction in a feed-forward model with a shared representation of multi-camera video. A common image encoder and factorized spatiotemporal attention efficiently combine observations across views and time. Task-specific decoders recover camera-to-rig calibration, ego motion, and metric depth; predict 3D objects, velocities, and high-resolution semantic occupancy; and reconstruct scene appearance with 3D Gaussians.}

\label{fig:architecture}
\end{figure*}

\subsection{Problem formulation}
\label{sec:problem_formulation}

Given a video clip with $T$ temporal samples from $C$ cameras, \method estimates camera $z$-depth maps and pointmaps, ego motion, 3D bounding boxes, semantic occupancy, and 3D Gaussians for rendering. We denote each input image by $\mathbf{I}_{t,c}\in\mathbb{R}^{H\times W\times3}$, where $t$ and $c$ index time and camera, and $H$ and $W$ are the image dimensions. Camera $c=1$ defines the reference timestamps $\tau_t=\tau_{t,1}$. Each image retains its capture time $\tau_{t,c}$, which may differ from $\tau_t$ for asynchronous inputs.

\subsection{Overview}
\label{sec:method_overview}
\method uses a shared image encoder and a factorized spatiotemporal aggregator to combine information across cameras and time (\cref{fig:architecture}). The geometry and Gaussian branches decode the resulting contextual features into geometric predictions and a renderable scene representation. Detection and occupancy share a feature pyramid that combines these features with the image encoder's finer spatial detail.

\subsection{Shared image and spatiotemporal representation}
\label{sec:shared_representation}
Following VGGT~\cite{wang2025vggt}, a shared DINOv2 ViT-L/14 encoder~\cite{oquab2023dinov2} extracts patch features independently from each image, with feature width $d=1024$. We use two spatial resolutions to reduce aggregation cost while retaining image detail. A $1036\times1036$ image produces a $74\times74$ encoder grid, which we bilinearly downsample to $37\times37$ before aggregation. The aggregator thus processes one quarter of the spatial tokens, while the original encoder features remain available to the detection and occupancy feature pyramid.

VGGT alternates within-image self-attention with global self-attention across all images. Global attention scales quadratically with the total number of tokens, making it expensive for multi-camera videos. We instead factor this operation across cameras and time. Each layer applies within-image self-attention followed by self-attention across cameras at a fixed temporal index or across time for a fixed camera. These camera and temporal groups alternate across layers, combining spatial and temporal context without attending over the full window at once (\cref{fig:architecture}).

Each image patch sequence also carries a rig token for relative camera calibration, an ego token for rig motion, and four register tokens for auxiliary context. All tokens participate in attention. After 16 aggregation layers, we concatenate the outputs of the final within-image and temporal attention blocks along the channel dimension. This gives a contextual token sequence $\mathbf{F}_{t,c}$ of width $2d$, whose patch features are denoted by $\mathbf{F}^{\mathrm{patch}}_{t,c}$.

\subsection{Geometry decoders}
\label{sec:geometry_decoder}
The geometry branch predicts camera poses and depth. We also train two auxiliary decoders: an image-semantic decoder that provides masks for video-level inference, and a pointmap decoder that supplies additional geometric supervision during training. We express camera-to-world poses $\mathbf{T}_{t,c}\in\mathrm{SE}(3)$ and pointmaps in the window reference frame, anchored to camera $c=1$ at the first reference timestamp $\tau_1$.

\paragraph{Pose prediction.}
We factor camera poses into a fixed rig transform $\mathbf{T}^{\mathrm{rig}}_c$ for each camera and a time-varying reference-camera trajectory $\mathbf{T}^{\mathrm{ego}}_t$. The rig head reads rig-token features from all cameras in the first temporal group, while the ego head reads ego-token features from the reference camera across time. Separate transformer heads process these sequences, following VGGT's camera-head architecture~\cite{wang2025vggt}, and regress translation and quaternion parameters through an MLP. The heads predict world-to-camera encodings, which are converted to camera-to-world transforms before pose composition.

Each rig transform maps a camera into the reference-camera frame and is reused throughout the window. We interpolate the ego trajectory at each image's capture time, using linear interpolation for translation and shortest-arc spherical interpolation for rotation. Denoting this trajectory by $\mathbf{T}^{\mathrm{ego}}(\tau)$, we construct the camera pose as
\begin{equation}
\mathbf{T}_{t,c}
=\mathbf{T}^{\mathrm{ego}}(\tau_{t,c})\mathbf{T}^{\mathrm{rig}}_c.
\label{eq:pose_composition}
\end{equation}
For synchronized images, the trajectory is evaluated at the shared timestamp $\tau_t$. Rig supervision uses relative poses from the first temporal group: these targets represent static calibration for synchronized inputs, but can include ego motion between exposures for asynchronous inputs. The composition above treats the predicted rig factor as fixed, and interpolating the ego trajectory does not remove this mismatch in the supervision target. Each image geometry remains associated with its own acquisition time; interpolation does not synchronize the images or compensate for independently moving objects. At inference, capture times outside the ego trajectory's sampled interval extend the nearest segment's motion, while a single trajectory sample yields a constant pose.

\paragraph{Dense prediction.}
Independently parameterized DPT decoders~\cite{ranftl2021dpt} transform the contextual patch features $\mathbf{F}^{\mathrm{patch}}_{t,c}$ into camera $z$-depth $\mathbf{D}_{t,c}$, an auxiliary pointmap $\mathbf{P}^{\mathrm{aux}}_{t,c}$ in the window reference frame, and auxiliary semantic logits $\mathbf{S}_{t,c}$. Each decoder reassembles tokens into spatial feature maps and fuses multiple scales through convolutional refinement, producing a $518\times518$ output grid. Positive confidence maps accompany the depth and auxiliary pointmap predictions. The auxiliary semantic decoder outputs 21 class logits per pixel through a final $1\!\times\!1$ convolution; the highest-scoring class gives the predicted label.

The final pointmap is constructed from predicted depth and camera poses rather than taken from the auxiliary decoder. After restoring metric units, we unproject depth through the calibrated camera model, accounting for lens distortion and the output-grid resolution. For one image, let $d(u)$ be the depth at pixel $u$, $\mathbf{r}(u)\in\mathbb{S}^2$ its unit camera ray, and $(\mathbf{R},\mathbf{t})$ the camera-to-world rotation and translation. For valid pixels with $[\mathbf{r}(u)]_z>0$, the corresponding point in the window reference frame is
\begin{equation}
\mathbf{p}(u)
=\mathbf{R}\frac{d(u)}{[\mathbf{r}(u)]_z}\mathbf{r}(u)+\mathbf{t}.
\label{eq:pointmap_lifting}
\end{equation}
The division by $[\mathbf{r}(u)]_z$ converts camera $z$-depth to distance along the ray.

\paragraph{Supervision.}
The geometry objective combines dense geometry, rig and ego poses, temporal relative motion, and auxiliary image semantics:
\begin{equation}
\mathcal{L}_{\mathrm{geo}}
=\lambda_D\mathcal{L}_D+\lambda_P\mathcal{L}_P
+\lambda_{\mathrm{rig}}\mathcal{L}_{\mathrm{rig}}
+\lambda_{\mathrm{ego}}\bigl(\mathcal{L}_{\mathrm{ego}}+\lambda_{\mathrm{rel}}\mathcal{L}_{\mathrm{rel}}\bigr)
+m_{\mathrm{sem}}\lambda_{\mathrm{sem}}\mathcal{L}_{\mathrm{sem}},
\label{eq:geometry_loss}
\end{equation}
We use $(\lambda_D,\lambda_P,\lambda_{\mathrm{rig}},\lambda_{\mathrm{ego}},\lambda_{\mathrm{rel}},\lambda_{\mathrm{sem}})=(1,1,10,10,1,1)$, with $m_{\mathrm{sem}}$ enabling semantic supervision only when annotations are available. Depth, point coordinates, and translations are divided by a fixed reference length of $40$\,m before supervision. In the losses below, hats denote predictions and unmarked quantities denote targets.

Depth and auxiliary pointmap supervision combine direct regression with a confidence-weighted term. Over valid pixels $i\in\mathcal{M}$, the residuals are $e_i^D=|\hat d_i-d_i|$ and $e_i^P=\|\hat{\mathbf{p}}^{\mathrm{aux}}_i-\mathbf{p}_i\|_2$. For $X\in\{D,P\}$ and predicted confidence $q_i^X$, we use
\begin{equation}
\mathcal{L}_X
=\mathcal{R}_{0.98}\!\left(\{e_i^X\}_{i\in\mathcal{M}}\right)
+\mathcal{R}_{0.98}\!\left(\{q_i^X e_i^X-\beta\log q_i^X\}_{i\in\mathcal{M}}\right),
\qquad \beta=0.2.
\label{eq:dense_geometry_loss}
\end{equation}
Here $\mathcal{R}_{0.98}$ applies robust averaging independently to each term. The logarithmic penalty regularizes confidence, while the unweighted term retains direct supervision.

Rig and ego losses supervise the heads' native world-to-camera encodings. We write the translation as $\mathbf{t}^{\mathrm{w2c}}$ to distinguish it from the camera-to-world translation used in unprojection, and denote the quaternion by $\mathbf{q}$. The error for one pose is
\begin{equation}
\ell_{\mathrm{pose}}
=\frac{1}{3}\|\hat{\mathbf{t}}^{\mathrm{w2c}}-\mathbf{t}^{\mathrm{w2c}}\|_1
+\frac{1}{4}\min_{\sigma\in\{-1,1\}}
\|\hat{\mathbf{q}}-\sigma\mathbf{q}\|_1.
\label{eq:pose_encoding_loss}
\end{equation}
The sign minimization accounts for the two quaternion representations of the same rotation. $\mathcal{L}_{\mathrm{rig}}$ and $\mathcal{L}_{\mathrm{ego}}$ average this error over their respective valid observations.

The relative-pose term additionally constrains motion between nearby temporal samples. For each unique pair $i<j$ with $j-i<5$, we form $\Delta\hat{\mathbf{T}}_{ji}=(\hat{\mathbf{T}}^{\mathrm{ego}}_j)^{-1}\hat{\mathbf{T}}^{\mathrm{ego}}_i$, which maps camera coordinates from time $i$ to time $j$; targets are constructed identically from ground-truth poses. For each pair, we sum the mean absolute errors over the three translation components and four quaternion components, then average across pairs to obtain $\mathcal{L}_{\mathrm{rel}}$. Relative quaternions are unit quaternions with nonnegative scalar components. The loss is zero when no pair is available.

For annotated images, the auxiliary $\mathcal{L}_{\mathrm{sem}}$ sums cross-entropy, soft Dice, and Lov\'{a}sz--Softmax losses with equal weights, excluding unknown pixels from all three terms. Cross-entropy uses uniform class weights and label smoothing of $0.05$. Soft Dice averages overlap over the 20 known classes, adding $1$ to both its numerator and denominator. Lov\'{a}sz--Softmax is evaluated over classes present in each image and then averaged across images.

For robust dense regression, $\mathcal{R}_{0.98}$ caps loss values at $100$ and excludes values at or above the $98$th percentile when sufficient valid observations are available; small sets use the untrimmed mean. Pose supervision requires more than $100$ valid pixels per observation when dense geometry labels are available.

\subsection{3D object detection} \label{sec:detection_decoder}

The object branch predicts a sparse set of classified 3D cuboids at each timestamp in the vehicle rig coordinate system, covering a region of interest of 
$[-108,150]\times[-108,108]\times[-3,5]\,\mathrm{m}$. 
Following PETR~\cite{liu2022petr}, we represent object hypotheses with queries anchored at learned 3D reference points. Each query predicts class scores and a cuboid described by its 3D center, dimensions, yaw, and planar velocity. The shared backbone provides multi-camera temporal context, while the object decoder produces detections for each timestamp. An offline tracker subsequently assembles these detections into trajectories over the complete clip.

The internal data detector uses 12 classes: automobile, heavy truck, bus, train or tram car, other vehicle, person, rider, trailer, protruding object, trolley bus, stroller, and animal. Public-benchmark classifiers are adapted to their target label sets.

\paragraph{Feature pyramid.}
A feature pyramid shared with occupancy fuses the high-resolution image features with the aggregator's contextual patch features (\cref{sec:shared_representation}). Lateral projections align their channels, after which the contextual features are upsampled and added to the image features through a top-down pathway.

\paragraph{Decoder architecture.}
The decoder follows the DETR3D  architecture~\cite{wang2021detr3d}, alternating query self-attention, multi-scale deformable cross-attention, and feed-forward layers. Query content is initialized to zero, while positional embeddings are computed from the learned 3D reference points. In each cross-attention layer, query features predict 3D offsets around their reference points. The resulting sampling locations are projected into the camera views using the supplied intrinsics, lens model, and extrinsics. We sample the feature pyramid at these locations and combine the samples with query-dependent attention weights over cameras, scales, and sampling points. Classification and regression heads read the refined queries at each decoder layer. The reference configuration uses 900 queries and six decoder layers of width 256.

\paragraph{Supervision.}
We use one-to-one Hungarian matching between predicted queries and annotated cuboids, with a cost combining focal classification and cuboid regression. Following PETR~\cite{liu2022petr}, classification uses sigmoid focal loss; matched queries receive a weighted $L_1$ regression loss. Unmatched queries are supervised as background. The regression representation contains metric center coordinates, logarithmic dimensions, sine and cosine of yaw angle, and planar velocity. The center predictions are parameterized as offsets from the normalized reference points and converted back to metric coordinates for supervision and readout. Planar velocities are later used to estimate object motion and obtain temporally consistent tracks, as explained in ~\cref{sec:mot}.

The detection objective combines supervision at every decoder layer with query denoising and an auxiliary image-space detection loss:
\begin{equation}
\mathcal{L}_{\mathrm{det}}
=\sum_{\ell=1}^{6}
\left(
\lambda_{\mathrm{cls}}\mathcal{L}_{\mathrm{focal}}^{(\ell)}
+\lambda_{\mathrm{box}}\mathcal{L}_{\mathrm{box}}^{(\ell)}
+\lambda_{\mathrm{DN}}\mathcal{L}_{\mathrm{DN}}^{(\ell)}
\right)
+\mathcal{L}_{\mathrm{2D}}.
\label{eq:detection_loss}
\end{equation}
Here $\mathcal{L}_{\mathrm{box}}$ includes coordinate-wise regression weights. We use $(\lambda_{\mathrm{cls}},\lambda_{\mathrm{box}},\lambda_{\mathrm{DN}})=(2,0.25,1)$; the denoising term contains the same weighted classification and regression losses. Focal loss uses $\alpha=0.25$ and $\gamma=2$. Intermediate decoder outputs receive the same forms of supervision as the final output.

Query denoising follows the training principle of DN-DETR~\cite{li2022dndetr}. During training, we append groups of queries whose reference points are obtained by perturbing ground-truth cuboid centers in proportion to object size. Their target assignments are known, bypassing Hungarian matching for these queries; strongly perturbed queries receive background classification targets. Attention masks prevent ordinary detection queries from attending to denoising queries and isolate the denoising groups from one another.

An auxiliary YOLOX-style head~\cite{ge2021yolox} predicts per-camera 2D boxes, classes, objectness, and projected cuboid centers from the pyramid features. Projected cuboid annotations provide its targets. Its loss is
\begin{equation}
\mathcal{L}_{\mathrm{2D}}
=\mathcal{L}_{\mathrm{cls}}^{\mathrm{2D}}
+\mathcal{L}_{\mathrm{obj}}^{\mathrm{2D}}
+5\mathcal{L}_{\mathrm{IoU}}^{\mathrm{2D}}
+\mathcal{L}_{L_1}^{\mathrm{2D}}
+\mathcal{L}_{\mathrm{center}}^{\mathrm{2D}},
\label{eq:detection_2d_loss}
\end{equation}
where classification and objectness use binary cross-entropy, box localization combines squared-IoU and $L_1$ losses, and projected centers use an $L_1$ loss. Query denoising and the 2D head are used only during training.

\subsection{Semantic occupancy}
\label{sec:occupancy_decoder}
The occupancy branch predicts binary occupancy and semantic labels on a metric voxel grid of size $V_x\times V_y\times V_z$ in the vehicle rig coordinate system, covering a region of interest of $[-20,100]\times[-30,30]\times[-3,5]\,\mathrm{m}$, at a voxel resolution of $0.2\,\mathrm{m}$. To limit memory usage, intermediate features remain in BEV, with the voxel height dimension introduced at the outputs.

\paragraph{BEV projection module.}

\begin{figure}[!tp]
\centering
\edef\bevPanelHeight{\the\dimexpr0.147\linewidth\relax}
\begin{minipage}[t]{0.392\linewidth}
\centering
\includegraphics[height=\bevPanelHeight]{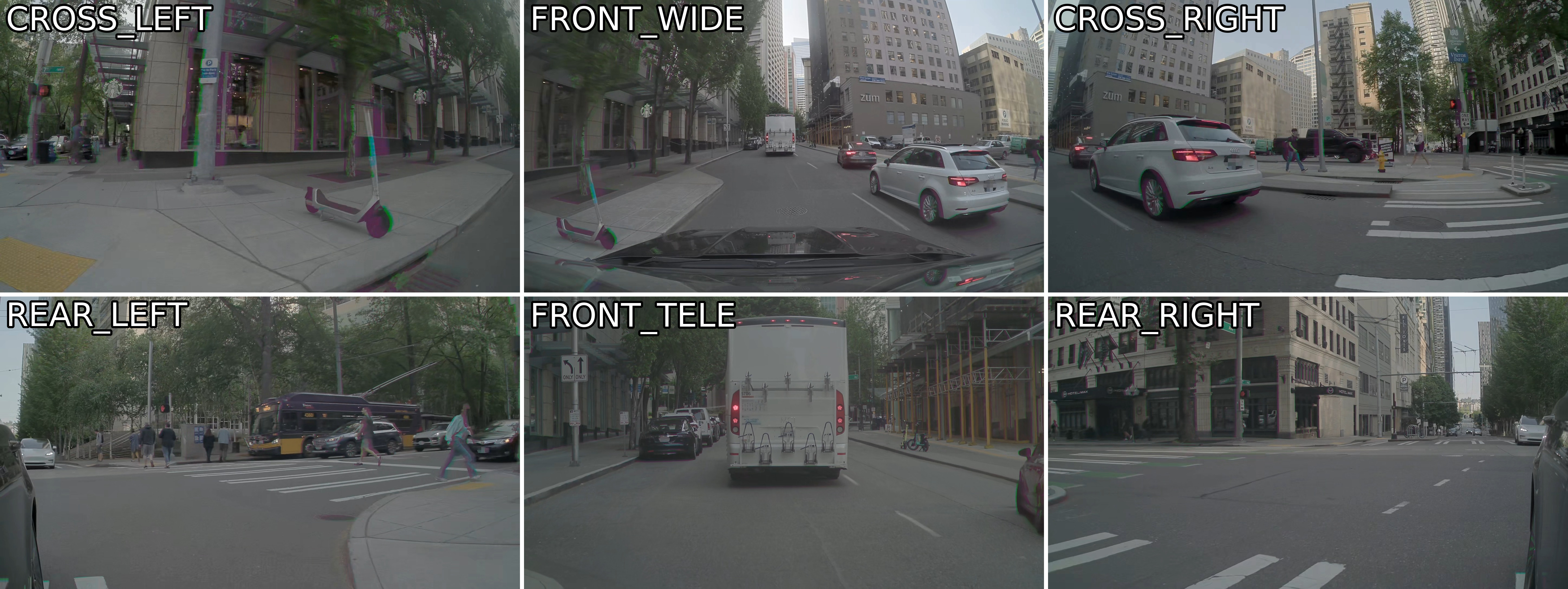}
\par\vspace{3pt}
{\footnotesize (a) Input camera frames\par}
\end{minipage}\hfill%
\begin{minipage}[t]{0.294\linewidth}
\centering
\includegraphics[height=\bevPanelHeight]{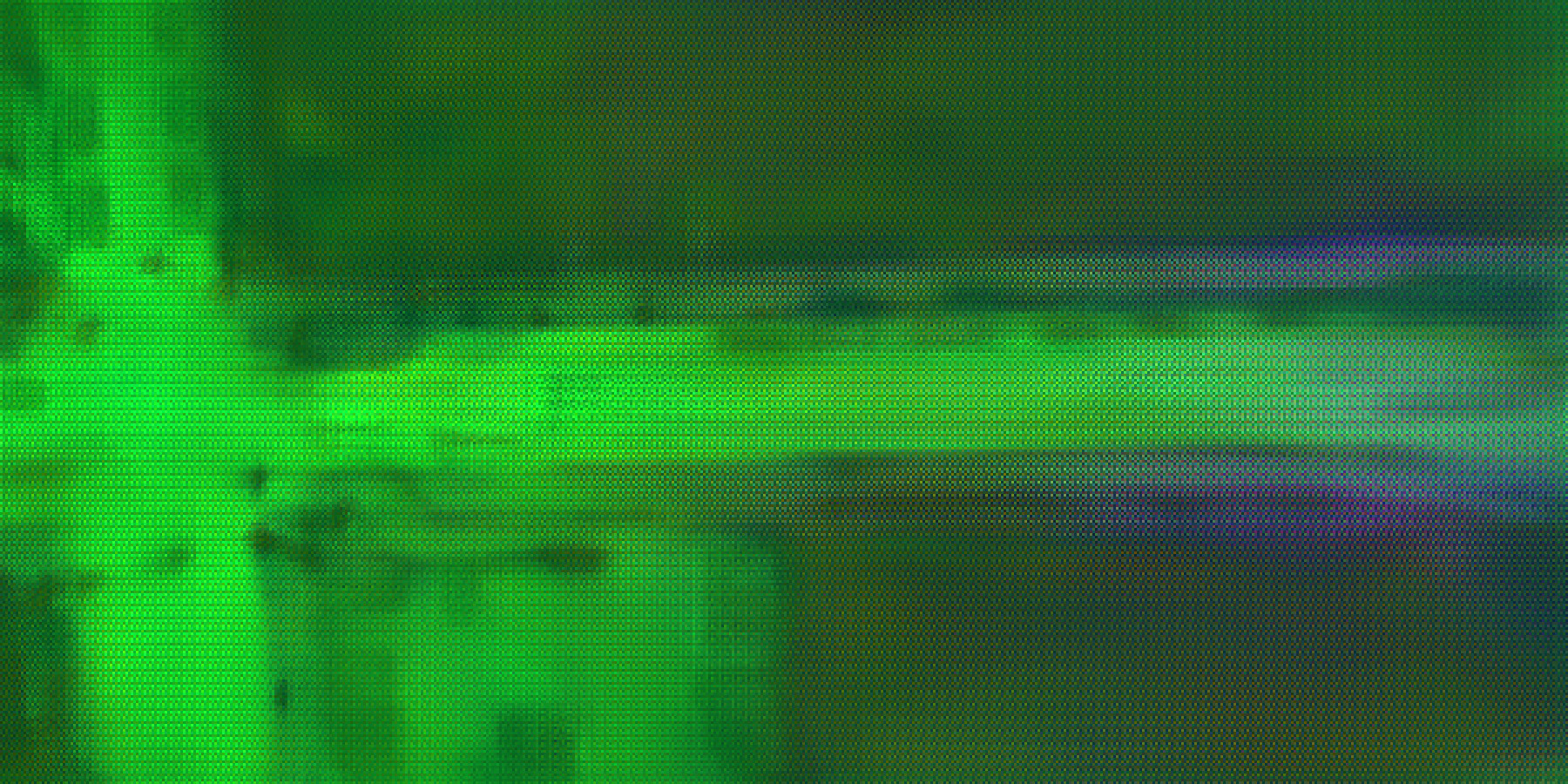}
\par\vspace{3pt}
{\footnotesize (b) BEV features\par}
\end{minipage}\hfill%
\begin{minipage}[t]{0.294\linewidth}
\centering
\includegraphics[height=\bevPanelHeight]{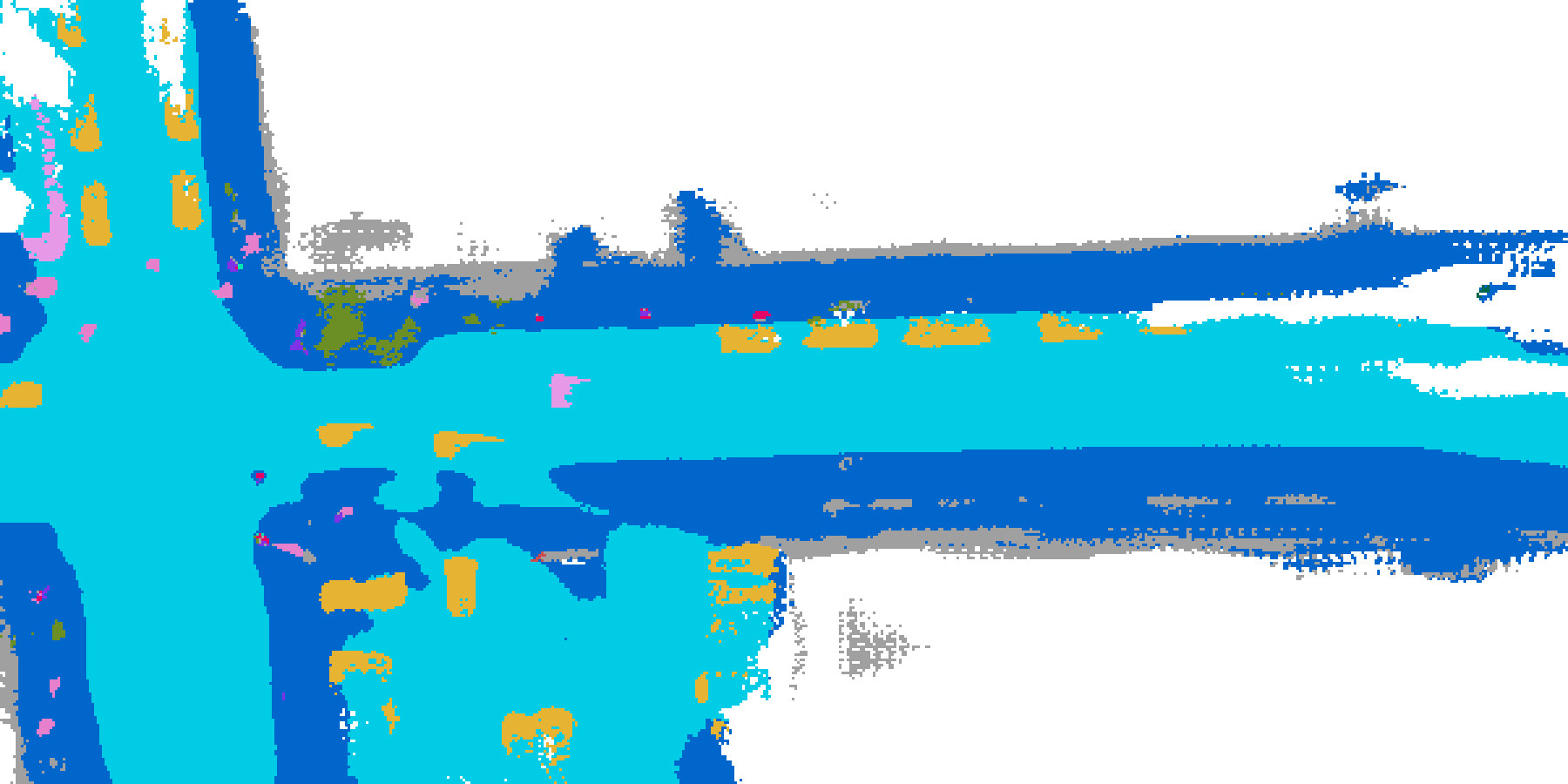}
\par\vspace{3pt}
{\footnotesize (c) BEV of \method predictions\par}
\end{minipage}
\caption{\textbf{\method semantic occupancy BEV features.} PCA of the upsampled features reveals a separation between occupied and empty regions. All panels show the same sample; PCA colors encode feature variation, while prediction colors denote semantic classes.}
\label{fig:bev_feature_visualization}
\end{figure}

Our BEV projector is a Deformable DETR~\cite{zhu2021deformable} inspired module that aggregates multi-scale image features produced by the shared feature pyramid neck into a spatially grounded BEV map for each timestamp (\cref{fig:bev_feature_visualization}). For each BEV cell, we place four reference points along its vertical pillar at uniformly spaced heights within the target height range. We encode each point's normalized 3D coordinates and use separate learned layers to initialize its query feature and positional embedding. Transformer layers refine these queries through self-attention and deformable cross-attention, while positional embeddings are used in self-attention and in predicting the deformable cross-attention weights. Cross-attention projects 3D sampling points into the camera views using camera intrinsics and static rig-to-camera extrinsics, which we assume are known and part of the vehicle rig. This projection requires no ego-motion estimates, since the reference points and BEV grid are defined in the rig coordinate system of each timestamp. A learned merging layer combines the refined pillar features into the BEV map $\mathbf{F}^{\mathrm{BEV}}_t$.

\paragraph{BEV upsampling and output heads.}
We use a fully convolutional decoder and progressively upsample $\mathbf{F}^{\mathrm{BEV}}_t$ through 2D transposed convolutions. Let $K_{\mathrm{occ}}$ denote the number of occupancy semantic classes. Separate transposed-convolution heads output $V_z$ binary channels and $V_zK_{\mathrm{occ}}$ semantic channels. Channel-to-height reshaping~\cite{yu2023flashocc} converts these outputs into binary logits $\mathbf{O}^{\mathrm{bin}}_t\in\mathbb{R}^{V_x\times V_y\times V_z}$ and semantic logits $\mathbf{O}^{\mathrm{sem}}_t\in\mathbb{R}^{V_x\times V_y\times V_z\times K_{\mathrm{occ}}}$. At inference time, the final semantic occupancy prediction is obtained by thresholding the sigmoid probabilities of $\mathbf{O}^{\mathrm{bin}}_t$ and assigning each occupied voxel the highest-scoring semantic class from $\mathbf{O}^{\mathrm{sem}}_t$. The remaining voxels are labeled as empty.

\paragraph{Supervision.}
We supervise occupancy using LiDAR-derived geometry and image-derived semantic labels, with an auxiliary BEV segmentation head attached to the upsampled BEV features. The training objective is
\begin{equation}
\mathcal{L}_{\mathrm{occ}}
=\lambda_{\mathrm{BCE}}\mathcal{L}_{\mathrm{BCE}}
+\lambda_{\mathrm{Lov}}\mathcal{L}_{\mathrm{Lovasz}}^{\mathrm{bin}}
+\lambda_{\mathrm{CE}}\mathcal{L}_{\mathrm{CE}}^{\mathrm{occ}}
+\lambda_{\mathrm{Dice}}\mathcal{L}_{\mathrm{Dice}}^{\mathrm{occ}}
+\lambda_{\mathrm{BEV}}\mathcal{L}_{\mathrm{BEV}}.
\label{eq:occupancy_loss}
\end{equation}
Here, $\mathcal{L}_{\mathrm{BCE}}$ and $\mathcal{L}_{\mathrm{Lovasz}}^{\mathrm{bin}}$ are binary cross-entropy and Lov\'asz hinge losses, while $\mathcal{L}_{\mathrm{CE}}^{\mathrm{occ}}$ and $\mathcal{L}_{\mathrm{Dice}}^{\mathrm{occ}}$ are cross-entropy and soft Dice losses for voxel semantics. The auxiliary BEV loss is $\mathcal{L}_{\mathrm{BEV}}=\lambda_{\mathrm{CE}}\mathcal{L}_{\mathrm{CE}}^{\mathrm{BEV}}+\lambda_{\mathrm{Dice}}\mathcal{L}_{\mathrm{Dice}}^{\mathrm{BEV}}$, combining cross-entropy and soft Dice supervision with the same weights used for voxel semantics. BEV targets are obtained by selecting the topmost semantic voxel in each vertical column. We use $(\lambda_{\mathrm{BCE}},\lambda_{\mathrm{Lov}},\lambda_{\mathrm{CE}},\lambda_{\mathrm{Dice}},\lambda_{\mathrm{BEV}})=(1,0.1,1,1,0.5)$ in our experiments. Loss terms include sums over the supervised projection-decoder layers.

\subsection{Gaussian reconstruction}
\label{sec:gaussian_decoder}

Following Instant NuRec~\cite{nvidia2026instantnurec}, we represent a driving scene with static Gaussians, actor-associated Gaussians, and a sky cubemap. The reconstruction branch uses shared contextual features alongside provided camera calibrations. During training, we utilize ground-truth camera-to-world poses and actor cuboid tracks, whereas during inference, we rely on predicted poses and trajectories. The Gaussian branch does not predict actor motion itself. 

For each source pixel $i=(t,c,u)$ on the decoder output grid, we predict a Gaussian with center $\boldsymbol{\mu}_i$, RGB color $\mathbf{c}_i$, scales $\mathbf{s}_i$, unit quaternion $\mathbf{q}_i$, and opacity $\alpha_i$, as well as a unit surface normal for auxiliary supervision. Its covariance is $\boldsymbol{\Sigma}_i=\mathbf{R}_i\operatorname{diag}(\mathbf{s}_i^2)\mathbf{R}_i^\top$, where $\mathbf{R}_i=\operatorname{Rot}(\mathbf{q}_i)$. Centers, orientations, and normals are expressed in the clip world frame defined by the predicted trajectory.

\paragraph{Gaussian decoding.}
DPT decoders~\cite{ranftl2021dpt} predict depth, color residuals, normals, and Gaussian attributes from contextual patch features. The Gaussian depth decoder is initialized from the geometry depth head but has separate parameters trained with reconstruction losses; the geometry head retains its own supervision.

We obtain each unit camera ray $\mathbf{r}_i$ by calibrated inverse projection,
accounting for lens distortion and image resizing.
Let $\mathbf{E}_c$ denote the known camera-to-rig transform.
At inference, the source camera pose is
$\widehat{\mathbf{T}}_{t,c}
=\widehat{\mathbf{T}}_{\mathrm{rig}}(\tau_{t,c})\mathbf{E}_c$,
where $\widehat{\mathbf{T}}_{\mathrm{rig}}$ is the assembled predicted
rig-to-world trajectory. Given predicted metric camera $z$-depth $d_i$
and the rotation and translation
$(\widehat{\mathbf{R}}_{t,c},\widehat{\mathbf{t}}_{t,c})$
of this camera pose, the center of each valid Gaussian is
\begin{equation}
\boldsymbol{\mu}_i
=\widehat{\mathbf{R}}_{t,c}
\frac{d_i}{[\mathbf{r}_i]_z}
\mathbf{r}_i+\widehat{\mathbf{t}}_{t,c}.
\label{eq:gaussian_lifting}
\end{equation}

The factor $d_i/[\mathbf{r}_i]_z$ converts camera depth to ray distance. Gaussian color adds a bounded learned residual to the resampled source RGB. To suppress Gaussians in sky, ego-vehicle, and unknown regions, we multiply their opacities by a mask from the geometry branch's semantic predictions. The mask uses soft probabilities during training with gradients stopped and discrete labels at inference.

\paragraph{Scene rendering.}
Supplied or predicted cuboid tracks associate source Gaussians with actors. To render a target timestamp, we interpolate each actor's predicted trajectory and rigidly transform its Gaussians from their source capture times, while static Gaussians remain fixed. Differentiable splatting produces accumulated RGB $\mathbf{c}_{\mathrm{G}}(u)$, accumulated opacity $A(u)$, and ray distance at target pixel $u$. A scene decoder predicts a world-space sky cubemap from contextual features, source RGB, and calibrated rays; its color $\mathbf{c}_{\mathrm{sky}}(u)$ along the target ray fills the residual transmittance. A separate decoder uses contextual features to predict an affine RGB transform for each camera, with $\mathbf{M}_c\in\mathbb{R}^{3\times3}$ and $\mathbf{b}_c\in\mathbb{R}^3$, to account for appearance differences. Applying the target camera's transform and clipping to $[0,1]$ gives the pixel colors of the rendered image $\mathbf{I}$:
\begin{equation}
\mathbf{c}(u)
=\operatorname{clip}_{[0,1]}\!\left[
\mathbf{M}_c\bigl(\mathbf{c}_{\mathrm{G}}(u)+(1-A(u))\mathbf{c}_{\mathrm{sky}}(u)\bigr)
+\mathbf{b}_c\right].
\label{eq:gaussian_composition}
\end{equation}

\paragraph{Supervision.}
Pose and cuboid-track annotations provide conditioning during reconstruction training, while RGB, depth, and available semantic annotations supervise the corresponding losses. This training setup is distinct from the predicted-input inference configuration described above: annotated poses and actor tracks are not required when reconstructing a clip at inference. We supervise rendering at two timestamps excluded from the context: one inside the context interval and one within $0.5$\,s after its last sample, with camera observations aligned to the target times. The rendering objective is
\begin{equation}
\mathcal{L}_{\mathrm{render}}
=\mathcal{L}_{\mathrm{RGB}}+0.2\mathcal{L}_{\mathrm{LPIPS}}
+0.1\mathcal{L}_{\mathrm{inv-dist}}+2\mathcal{L}_{\mathrm{opacity}},
\label{eq:gaussian_render_loss}
\end{equation}
where the RGB, inverse-ray-distance, and accumulated-opacity terms use mean squared error; the opacity target is the non-sky mask. Pixel losses average over valid pixels, while LPIPS averages over valid target images after masked pixels are replaced by target colors. The full task objective adds source RGB consistency, normal supervision, sky consistency, and affine regularization:
\begin{equation}
\begin{aligned}
\mathcal{L}_{\mathrm{gs}}&=\mathcal{L}_{\mathrm{render}}
+0.01\mathcal{L}_{\mathrm{src\text{-}RGB}}
+0.2\mathcal{L}_{\mathrm{normal}}\\
&\quad+0.01\mathcal{L}_{\mathrm{sky}}
+0.01\mathcal{L}_{\mathrm{affine}}.
\end{aligned}
\label{eq:gaussian_task_loss}
\end{equation}

Source RGB consistency uses an $L_1$ loss, and normal supervision uses cosine distance to normals derived from depth annotations. The sky loss is $L_1$ against observed sky colors, using a smoothed prediction as the target in unobserved regions. Affine transforms are regularized by their mean squared deviation from identity. Supervision masks come from semantic annotations on NRE and PlanNet and detached semantic predictions from the geometry branch on MLMCF; the latter are not semantic training targets. These supervision masks are distinct from the predicted mask applied to Gaussian opacity.

\section{Training the unified model}
\label{sec:training}

\subsection{Training data}
We train on three internal driving datasets: NRE, PlanNet, and MLMCF, comprising approximately 68k, 76k, and 60k clips, respectively, with some overlap between collections. Table~\ref{tab:training_data_distribution} summarizes their scene
composition. Clips average approximately $12$\,s in length, amounting to roughly $695$ hours of driving in total. The sensor rigs use six polynomial $f$-theta cameras~\cite{nvidia_ncore_ftheta} with different fields of view, covering both wide-angle and narrower views of the scene. Images are captured at 30\,FPS, and a 128-beam LiDAR operates at 10\,Hz. Training windows subsample these recordings to cover different temporal spans.

\begin{table}[t]
    \centering
    \small
    \setlength{\tabcolsep}{3pt}
    \begin{tabular}{@{}lrrrrrr@{}}
        \toprule
        Dataset & Clips & Urban & Residential & Highway & Rural & Other \\
        \midrule
        NRE     & 67,970 & 16.5\% &  9.1\% & 64.5\% &  7.6\% & 2.3\% \\
        PlanNet & 75,921 & 42.5\% & 17.3\% & 20.8\% & 16.5\% & 2.9\% \\
        MLMCF   & 59,999 & 36.5\% & 10.9\% & 42.9\% &  6.9\% & 2.8\% \\
        \bottomrule
    \end{tabular}
    \caption{\textbf{Scene composition of the internal training splits.} NRE is dominated by highway driving, while PlanNet emphasizes urban
and residential scenes. MLMCF provides a more balanced mix of urban
and highway driving, complementing the other two collections.
Counts and percentages refer to unique clips within each dataset.}
    \label{tab:training_data_distribution}
\end{table}

The datasets provide depth, camera motion, cuboid tracks, and, where available, semantic annotations. LiDAR measurements are used to generate depth through the annotation pipeline described in \cref{sec:pai_benchmark}. Semantic occupancy labels are obtained through an internal auto-labeling pipeline providing annotations for 10 semantic classes. Detection annotations cover $12$ classes, with LiDAR-based cuboid autolabels in NRE and manually annotated cuboids in PlanNet and MLMCF.

\subsection{Training strategy}
We train the six-camera unified model in three stages, increasing temporal context before introducing Gaussian reconstruction. The \textit{first} stage jointly trains the image encoder, spatiotemporal aggregator, and geometry, detection, and occupancy heads on short windows, reducing the cost of each update. The \textit{second} stage extends the windows to provide more temporal context for pose and motion estimation. We freeze the image encoder at this point and continue training the aggregator and the three task heads.

The \textit{third} stage adds Gaussian reconstruction to the trained perception model. We first warm up the Gaussian heads with the existing model frozen, then jointly optimize the aggregator and all four task heads while keeping the image encoder frozen. This schedule defers the cost of Gaussian decoding and rendering until the perception model has been trained.

At each stage, we optimize a weighted sum of the active task losses defined in \cref{sec:method}:
\begin{equation}
\mathcal{L}
=\lambda_{\mathrm{geo}}\mathcal{L}_{\mathrm{geo}}
+\lambda_{\mathrm{det}}\mathcal{L}_{\mathrm{det}}
+\lambda_{\mathrm{occ}}\mathcal{L}_{\mathrm{occ}}
+\lambda_{\mathrm{gs}}\mathcal{L}_{\mathrm{gs}}.
\label{eq:joint_loss}
\end{equation}
For joint fine-tuning with Gaussian reconstruction, we use $(\lambda_{\mathrm{geo}},\lambda_{\mathrm{det}},\lambda_{\mathrm{occ}},\lambda_{\mathrm{gs}})=(1,0.1,0.2,4)$. A loss contributes only when its task is active and the required supervision is available; individual losses also exclude invalid pixels, objects, or voxels. All three datasets contribute to geometry, detection, and Gaussian reconstruction, whereas occupancy uses only NRE and PlanNet. Only NRE is used to supervise the auxiliary image-semantic head. The single-task and pairwise recipes used to study task transfer are described separately in \cref{sec:joint_ablation}.

\subsection{Implementation details}
\label{sec:training_implementation}
All stages use six cameras with images resized to $1036\times1036$. During temporal refinement and final joint fine-tuning, we sample NRE, PlanNet, and MLMCF in a $1:2:1$ ratio. The first stage uses four timestamps per camera and random frame strides up to eight, and runs for approximately 170k updates. Temporal refinement increases the window to eight timestamps and the maximum stride to 16 for a further 45k updates. Gaussian training retains this window configuration, with head warm-up followed by 10k joint updates.

The first two stages use AdamW with a main learning rate of $3\times10^{-4}$ and weight decay of 0.01. During the first stage, the image encoder uses one tenth of this learning rate. Both stages use 500 updates of linear learning-rate warm-up. The first then keeps the learning rate constant, while the second holds it until update 10k before applying cosine decay. The Gaussian heads are warmed up with a learning rate of $10^{-4}$. Final joint fine-tuning uses a main learning rate of $10^{-5}$ and weight decay of 0.01, with 500 warm-up updates followed by cosine learning-rate decay to zero.

We train with BF16 mixed precision, retaining FP32 computation in the geometry and Gaussian decoders and in numerically sensitive detection and occupancy operations. On NVIDIA H100 GPUs, FlashAttention-3 (FA3)~\cite{shah2024flashattention3} implements attention in the shared spatiotemporal backbone. Custom CUDA kernels apply two-dimensional rotary position embeddings (RoPE)~\cite{su2024roformer} to queries and keys directly in the packed query--key--value layout, avoiding separate tensor repacking before attention. On 64 H100 GPUs, representative geometry, detection, and occupancy training windows took approximately 3.8 seconds per update with four timestamps per camera and 8.3 seconds with eight.

\section{Multi-camera video inference system}
\label{sec:inference_system}

To reconstruct videos beyond the model's temporal context, we process overlapping windows and assemble their predicted camera trajectories and dense geometry in a common coordinate frame. The overlap provides repeated observations that connect the windows while keeping the network's context length fixed. We consider two alternative assembly methods: sequential window stitching and global pose graph optimization (PGO). Both use the predicted geometry and known camera intrinsics without updating network parameters. This formulation is related to overlapping-window and submap reconstruction in VGGT-Long~\cite{deng2026vggtlong}, VGGT-SLAM~\cite{maggio2025vggtslam}, and VGGT-SLAM~2.0~\cite{maggio2026vggtslam2}. \cref{fig:teaser_3d} illustrates the assembled geometry alongside occupancy and detections.
The resulting camera trajectories also place 3D object detections in the common reconstruction frame, where an offline multi-object tracker associates them across time to recover object trajectories over the entire clip.

\subsection{Sequential window stitching}
\label{sec:stitching}
Stitching aligns each new window to the preceding, already aligned window with a single $\mathrm{Sim}(3)$ transform. We pool depth-derived pointmap correspondences across cameras and overlapping frames: each pair $\mathbf{x}_a,\mathbf{y}_a$ describes the same camera, frame, and pixel in the new and preceding windows, respectively. These correspondences can include foreground objects because they refer to the same captured observation. Over the set $\mathcal{O}_k$ of correspondences for window $k$, we estimate scale $s_k$, rotation $\mathbf{R}_k$, and translation $\mathbf{t}_k$ by iteratively reweighted least squares:
\begin{equation}
\mathcal{E}_k(s,\mathbf{R},\mathbf{t})
=\sum_{a\in\mathcal{O}_k}m_aw_a\,
\rho_{\mathrm{H}}\!\left(\left\|s\mathbf{R}\mathbf{x}_a+\mathbf{t}-\mathbf{y}_a\right\|_2\right),
\label{eq:multicamera_sim3_alignment}
\end{equation}
Here, $\rho_{\mathrm{H}}$ is the Huber penalty, $m_a$ excludes correspondences classified as sky by the model's 2D semantic predictions in either window, and $w_a$ is the geometric mean of the two depth confidences. If masks or confidences are unavailable, the corresponding masks or weights are set to one.

We apply the estimated similarity transform to all camera poses in the window and scale the corresponding depth maps by $s_k$, placing both in the common reconstruction frame defined by the first window. Previously assembled predictions are retained in overlapping frames, and only new temporal samples are appended. Sequential alignment can accumulate rotation, translation, and scale errors: later windows do not revise earlier poses, and a single transform cannot correct inconsistencies within a window's trajectory. This motivates jointly reconciling the pose estimates across all windows.

\subsection{Global pose graph optimization}
\label{sec:pgo}
PGO estimates a single reference-camera trajectory for the entire video from the raw window predictions. It assumes a common metric scale across windows, supported by the fixed-scale supervision described in \cref{sec:geometry_decoder}. After restoring metric units, we optimize poses in $\mathrm{SE}(3)$ without estimating window scales or rescaling depth. Each unique reference-camera sample is a graph node, shared by all windows containing that sample. Within window $k$, the predicted camera-to-window poses $\widehat{\mathbf{T}}^{(k)}_i$ define relative measurements $\mathbf{Z}^{(k)}_{ij}=(\widehat{\mathbf{T}}^{(k)}_i)^{-1}\widehat{\mathbf{T}}^{(k)}_j=(\mathbf{R}^{(k)}_{ij},\mathbf{t}^{(k)}_{ij})$, which map coordinates from sample $j$ to sample $i$. These measurements remove the window's reference rotation and translation while retaining its predicted scale. Measurements of the same pair from different windows remain separate factors, allowing the windows to constrain a shared trajectory. The graph uses predicted relative poses and timestamps, without external pose factors or loop-closure constraints.

We solve for the trajectory in two stages, first optimizing rotations and then translations. Let $\mathbf{T}_i=(\mathbf{R}_i,\mathbf{t}_i)$ map reference-camera coordinates at node $i$ to the video coordinate frame, with the first pose fixed to the identity. Over the multiset $\mathcal{E}$ of window factors, the objectives are
\begin{equation}
\begin{aligned}
\{\mathbf{R}_i^\star\}
&=\underset{\mathbf{R}_i\in\mathrm{SO}(3),\,\mathbf{R}_1=\mathbf{I}_3}{\arg\min}
\frac{1}{\Omega}\sum_{(k,i,j)\in\mathcal{E}}
\omega^{(k)}_{ij}
\left\|\mathbf{R}_i^\top\mathbf{R}_j-\mathbf{R}^{(k)}_{ij}\right\|_F^2,\\
\{\mathbf{t}_i^\star\}
&=\underset{\mathbf{t}_1=\mathbf{0}}{\arg\min}
\frac{1}{\Omega}\sum_{(k,i,j)\in\mathcal{E}}
\omega^{(k)}_{ij}
\left\|(\mathbf{R}_i^\star)^\top(\mathbf{t}_j-\mathbf{t}_i)-\mathbf{t}^{(k)}_{ij}\right\|_2^2
+\lambda_{\mathrm{motion}}\mathcal{L}_{\mathrm{motion}},
\end{aligned}
\label{eq:clip_pgo}
\end{equation}
where $\omega^{(k)}_{ij}$ weights each factor and $\Omega=\sum_{(k,i,j)\in\mathcal{E}}\omega^{(k)}_{ij}$. With rotations fixed, translation estimation is a linear least-squares problem. The optional motion prior $\mathcal{L}_{\mathrm{motion}}$ penalizes changes between incoming and outgoing translation velocities at each interior node, using the actual reference-camera timestamp intervals. Each velocity difference is multiplied by the median sampling interval before taking its squared Euclidean norm, and the penalty is averaged over interior nodes.

The optimized trajectory $\mathbf{T}_i^\star=(\mathbf{R}_i^\star,\mathbf{t}_i^\star)$ then provides a correction for every camera at its acquisition time. For each sample, we retain depth and semantic predictions from the window in which it lies closest to the temporal center. For an image captured at $\tau_{t,c}$ in selected window $k$, we interpolate the source and optimized reference-camera trajectories separately and apply
\begin{equation}
\widehat{\mathbf{T}}^{\mathrm{clip}}_{t,c}
=\mathbf{T}^\star(\tau_{t,c})
\left[\widehat{\mathbf{T}}^{(k)}(\tau_{t,c})\right]^{-1}
\widehat{\mathbf{T}}^{(k)}_{t,c}.
\label{eq:clip_capture_time_correction}
\end{equation}
This preserves each camera's predicted relation to the source reference trajectory while transferring the optimized motion to its actual capture time. The retained depth is unchanged. For either assembly method, final pointmaps are obtained by unprojecting the retained depth with the assembled camera poses using \cref{eq:pointmap_lifting}. \cref{tab:geometry_inference_ablation} compares the two methods under a matched checkpoint and window schedule.

In our PGO implementation, factors connect samples separated by 1, 2, 4, or 8 sampled rows within each window. Their weights decrease linearly with the pair's midpoint position, from 1 at the window's beginning to 0.1 at its end. Rotations are initialized by chaining averaged adjacent-frame measurements and refined with L-BFGS using rotation-vector updates. We use $\lambda_{\mathrm{motion}}=30$; setting it to zero disables the motion prior, which is also omitted for trajectories with fewer than three nodes.

\subsection{Offline multi-object tracking} \label{sec:mot}
We collect 3D detections over the complete clip and use the assembled ego trajectory to transform their centers, orientations, and velocities from timestamp-specific rig frames into a common world frame. We then associate detections into tracks in temporal order. A constant-velocity Kalman filter maintains each track’s planar position and velocity, using both positions and velocities predicted by the detection model as measurements. At each observation time, the filter predicts track positions, and Hungarian assignment matches detections to these predictions by minimizing planar distance subject to a distance gate. Matched detections update the filter state, unmatched detections initialize new tracks, and unmatched tracks remain eligible for association for a bounded time.

Tracking proceeds in two phases. The first uses a shorter track lifetime to form conservative trajectories; we then interpolate gaps and smooth the trajectories before a second pass with a longer lifetime reconnects remaining fragments. Final planar positions and velocities are refined with a Rauch--Tung--Striebel (RTS) backward smoother, allowing later observations to improve earlier state estimates. Height and yaw are smoothed separately, with yaw unwrapped before interpolation and smoothing, and each trajectory is assigned median cuboid dimensions.

\section{Physical AI Autonomous Vehicles Foundation4D}
\label{sec:pai_benchmark}
Physical AI Autonomous Vehicles (PAI)~\cite{nvidia2025physicalai-av} provides multi-sensor driving recordings for autonomous vehicle research, including multi-camera RGB videos, LiDAR point clouds, and radar detections.
The dataset also provides sensor calibration, ego-motion trajectories, and automatically generated 3D cuboid tracks with object poses, dimensions, semantic classes, and track identifiers.
We augment the PAI sequences used in our benchmark with LiDAR-derived depth maps and occupancy grids with per-voxel semantic and velocity labels for dense 3D perception evaluation~\cite{nvidia2026physicalai-av-f4d}.
Our geometry, detection, and semantic occupancy evaluations use the same 1,253 PAI test clips and six cameras. Task-specific protocols are described in \cref{sec:experiments}.

\begin{figure}[!tp]
\centering
\includegraphics[width=\linewidth]{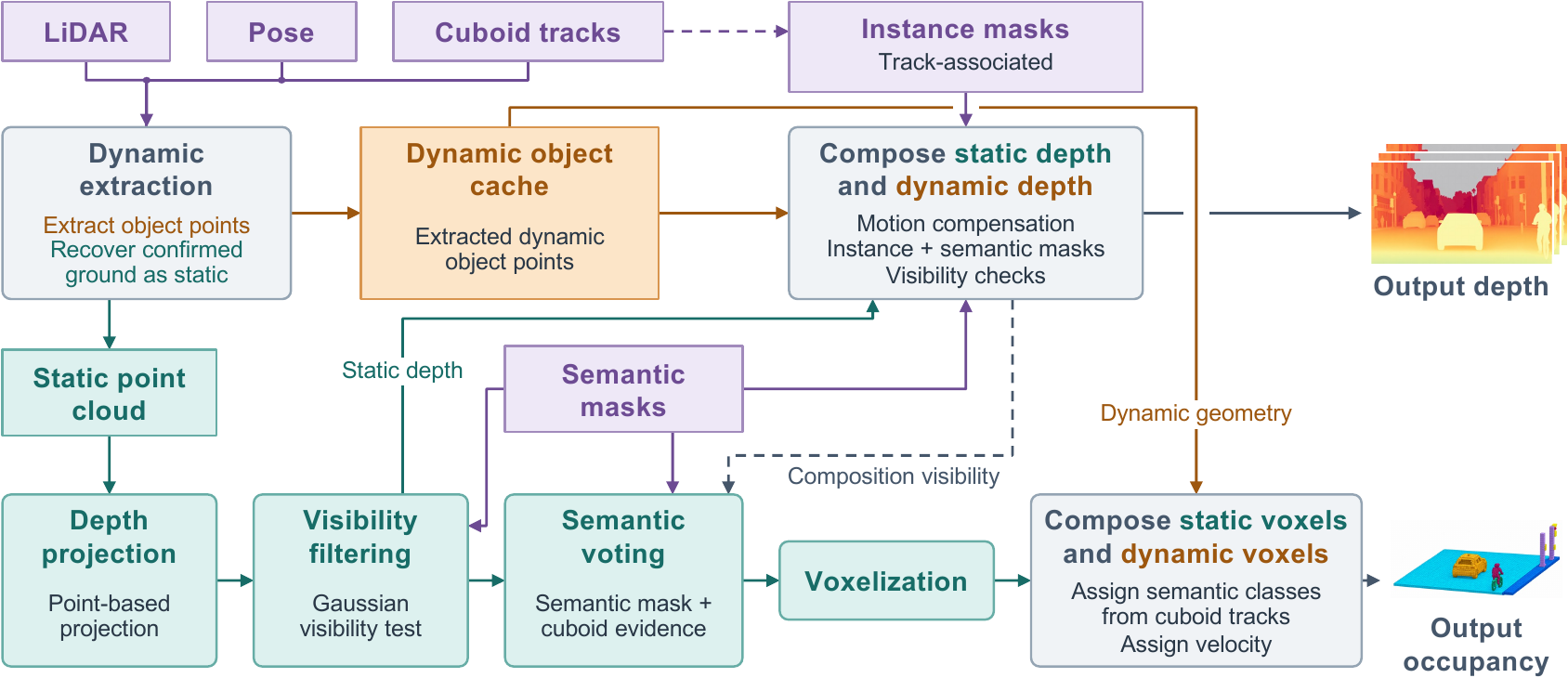}
\caption{\textbf{PAI depth and occupancy annotation pipeline.} The pipeline combines temporally aggregated LiDAR, cuboid tracks, and image masks to generate camera-view depth maps and vehicle-frame occupancy grids with per-voxel semantic and velocity labels.}
\label{fig:pai_label_pipeline}
\end{figure}

\begin{figure}[!tp]
\centering
\includegraphics[width=\linewidth]{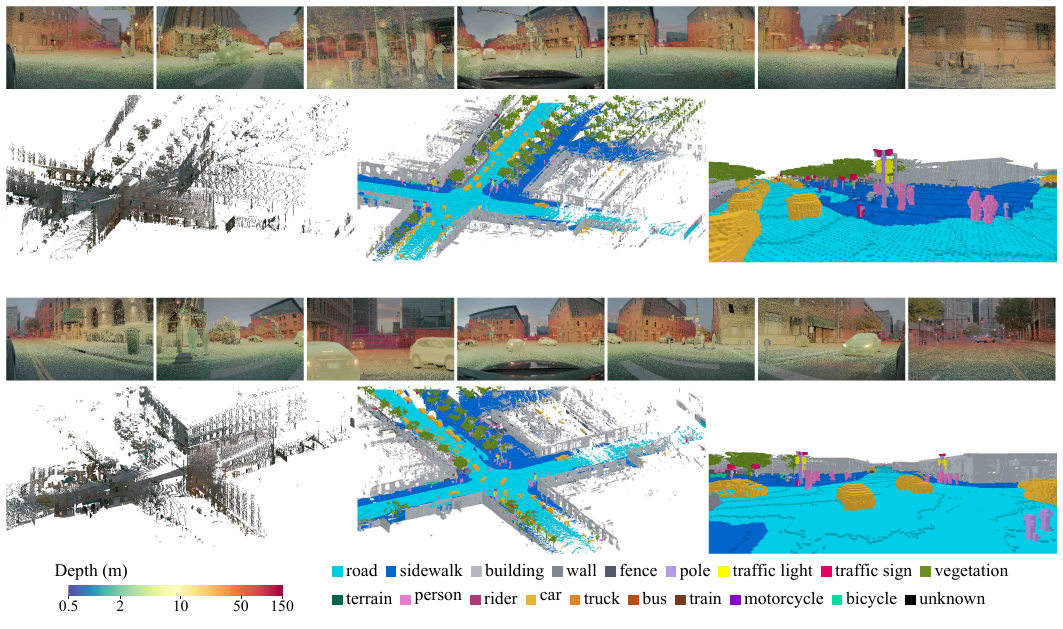}
\caption{\textbf{Generated depth and semantic occupancy annotations on PAI.} Each example shows multi-camera RGB images overlaid with depth (top), followed by unprojected depth colored by RGB, a side view of semantic occupancy, and a forward view of semantic occupancy (bottom, left to right). The shared legend indicates depth in meters and semantic classes.}
\label{fig:pai_gt_depth_occupancy}
\end{figure}

The annotation pipeline aggregates static LiDAR points across time and combines motion-compensated dynamic points within local temporal windows. 
As auxiliary inputs, it uses semantic masks precomputed by an internal semantic segmentation model and instance masks generated by SAM~2~\cite{ravi2024sam2} using box prompts derived from projected 3D cuboids.
Camera calibration and sensor poses align LiDAR points across time and camera views, while cuboid tracks guide motion compensation and supplement image semantic predictions with cuboid class labels.
The pipeline generates two outputs: depth maps rendered in each camera view and semantic occupancy grids in the vehicle frame.
\cref{fig:pai_label_pipeline} summarizes the annotation pipeline, and \cref{fig:pai_gt_depth_occupancy} illustrates the resulting depth and semantic occupancy annotations on two PAI examples.

\subsection{Temporal LiDAR aggregation}
Single LiDAR scans sparsely sample the scene, so we accumulate measurements over time to obtain more complete surface coverage. We compensate each return for ego motion using its acquisition timestamp and transform it into a common world frame. Cuboid tracks identify actors that require motion compensation: static points are accumulated across the full clip, while dynamic points are collected from a local window of up to ten LiDAR scans and warped to the actor pose at the target timestamp. This separation increases coverage while preserving the positions of moving objects.

Smoothed cuboid trajectories are used to classify tracks as static or dynamic. 
For dynamic tracks, we interpolate the original cuboid poses to select points within padded cuboid crops and align them across time. 
Ground points within these crops are excluded from object reconstruction, with only high-confidence ground returns restored to the static cloud. 
The accumulated static cloud is then filtered by local density while preserving sparse planar surfaces supported by multiple scans.
The pipeline also supports optional intensity filtering, with separate thresholds for static and dynamic returns.

\subsection{Depth annotation generation}
We generate metric depth maps by rendering the accumulated static scene and motion-compensated actors into each camera view. 
Projection uses calibrated intrinsics and extrinsics at the camera's native resolution, accounting for camera motion during rolling-shutter exposure. 
Static depth is obtained from the nearest visible projected LiDAR point; dynamic points are rendered with fixed-radius Gaussian splats. 
Dynamic depth is integrated into the static depth map where instance masks overlap with dynamic semantic regions, subject to occlusion checks.
Objects without instance masks use direct point projection.

Visibility checks prevent occluded points from contributing to the depth map. We compare candidate depths against measured static depth and an auxiliary splatted reference, rendered from the static cloud, using the latter only where it is consistent with nearby measurements and semantic labels. Within masked dynamic regions, pixels without rendered dynamic depth are marked invalid instead of retaining background depth. Static depth is retained when a dynamic prediction is rejected by occlusion checks. Visible dynamic points from direct projection can fill missing depth. Each valid output stores distance along the camera viewing ray in meters. Missing depth, sky, and ego-vehicle pixels are set to zero.

\subsection{Semantic occupancy annotation generation}
We construct semantic occupancy annotations by assigning classes to the accumulated 3D points and voxelizing them in the target vehicle frame.
Static points receive depth-consistent semantic votes from precomputed image semantic segmentation maps across multiple cameras and frames, with greater weight given to nearer observations. These votes are supplemented by the classes of static cuboid tracks.
The highest accumulated weighted score determines each static point's label, while dynamic points inherit the class of their associated cuboid track. 
Traffic-sign and traffic-light labels are then refined by propagating them from dense labeled regions to nearby connected points in 3D. Unlabeled and sky points are removed before constructing the static occupancy map.

The static cloud is pre-voxelized in the world frame and cleaned to remove isolated components. At the end timestamp of each front wide-angle camera frame, we transform the static voxel centers and motion-compensated dynamic points into the vehicle frame. 
The exported grid uses $0.1$\,m cubic voxels over $x\in[-150,150)$\,m, $y\in[-100,100)$\,m, and $z\in[-3,5)$\,m. Static voxel centers and dynamic points are voxelized separately, with majority voting within each set. Dynamic voxels take precedence where the two sets overlap, except where static voxels represent road, sidewalk, or terrain.
Each occupied voxel also stores a 3D velocity in meters per second, representing world-relative motion expressed in vehicle coordinates. Here, static voxels have zero velocity while dynamic voxels inherit their track's translational velocity, estimated from successive cuboid centers.

\section{Experiments}
\label{sec:experiments}

We evaluate \method on geometry, 3D detection and tracking, semantic occupancy, and Gaussian reconstruction using internal datasets and public benchmarks\footnote{Bold values indicate the best result for each metric within the corresponding comparison group.}. Component and inference ablations examine the effects of model design, supervision, and video-level assembly. Finally, we compare different combinations of training tasks to assess the benefits and trade-offs of learning a shared representation.

\subsection{Geometry}
\label{sec:geometry_experiments}

Our geometry heads aim to provide a consistent video-level reconstruction of the 3D scene captured by multi-camera sensors. We evaluate this capability along three axes: (i) multi-camera video reconstruction on public and PAI benchmarks (\cref{sec:geometry_public_benchmark,sec:geometry_pai_benchmark}); (ii) the contribution of core model design choices; and (iii) the impact of our video inference system (\cref{sec:geometry_ablations}).

\paragraph{Evaluation protocol.}
\label{sec:geometry_metrics}
We jointly align the predicted multi-camera trajectories to their ground-truth counterparts with a single similarity transform $(s,\mathbf{R},\mathbf{t})\in\mathrm{Sim}(3)$ per evaluation sequence: one window for window-level evaluation or one complete clip for video-level evaluation. We compute trajectory errors from the aligned camera poses: ATE measures the mean camera-center error, while RTE and RRE measure relative translation and rotation errors between consecutive poses. We apply the same scale $s$ to all predicted depth maps, $\widetilde{\mathbf{D}}_{t,c}=s\widehat{\mathbf{D}}_{t,c}$, and evaluate AbsRel and $\delta_{1.25}$ against ground-truth depth over valid pixels. AbsRel is the mean absolute relative depth error; $\delta_{1.25}$ is the percentage of pixels whose predicted-to-ground-truth depth ratio lies between $\frac{1}{1.25}$ and $1.25$.
Rig RTE and RRE compare the recovered static inter-camera transforms with calibrated extrinsics, using the same scale $s$ for rig translations. Rig mAA@$K$ averages joint rotation and translation-direction accuracy over angular thresholds from $1^\circ$ to $K^\circ$. Alignment to ground truth is used only for evaluation.

\subsubsection{Public benchmarks}
\label{sec:geometry_public_benchmark}
We compare the geometric reconstruction performance of \method with recent driving geometry and unified perception models on the public Waymo~\cite{sun2020waymo} and nuScenes~\cite{caesar2020nuscenes} datasets, evaluating depth, trajectory, and camera-rig accuracy.

\paragraph{Baselines.}
We consider DVGT-1, DVGT-2, LiAuto-GeoX, and GeoUP as our baselines, using their publicly released checkpoints and codebases. DVGT-1~\citep{zuo2026dvgt} predicts metric pointmaps and ego poses from unposed multi-camera images, while DVGT-2~\citep{zuo2026dvgt2} predicts local pointmaps and relative ego motion through streaming inference. Both take images alone as model inputs; for evaluation, we use known camera-to-rig extrinsics to derive camera poses and depth from their ego-frame outputs. LiAuto-GeoX~\citep{lian2026liautogeox} is a compact model distilled from a pose-conditioned geometry teacher; we use its released student to predict depth from images alone. Although it also predicts camera poses, we lack sufficient implementation details to reproduce its reported pose performance.

GeoUP~\citep{xu2026geoup} jointly predicts depth, 3D objects, and semantic occupancy, assuming known camera intrinsics, camera-to-rig calibration, and ego poses. We evaluate all methods on depth estimation and compare camera pose accuracy only with DVGT-1. None explicitly estimates a shared, static camera-to-rig calibration as \method does.

\paragraph{Setup.}
Following all four baselines, we include Waymo and nuScenes in our training domain. We therefore fine-tune the backbone and geometry decoders of \method for 20,000 iterations on a mixture of our internal NRE data and the training splits of Waymo (798 sequences, five cameras) and nuScenes (700 scenes, six cameras), sampled with probabilities $0.30$, $0.35$, and $0.35$, respectively.

We evaluate on the official validation splits of Waymo and nuScenes, with ground-truth depth rendered from clip-aggregated LiDAR using the pipeline in \cref{sec:pai_benchmark}. We compare depth across baselines using window-level inference on all non-overlapping 48-frame windows and average metrics over all window predictions. Independent scale alignment within each window limits the effect of accumulated pose drift on depth evaluation. For ego-pose evaluation, we adopt the more practical video-level setting and process complete clips. For a fair comparison, we integrate both DVGT-1 and \method into our video inference system (\cref{sec:inference_system}), using the same 48-frame windows with a 5-frame overlap at each dataset's native frame rate.

\paragraph{Results.}
On both Waymo and nuScenes, the fine-tuned \method model achieves the lowest depth AbsRel and highest $\delta_{1.25}$ among the evaluated methods (\cref{tab:geometry_depth_public_benchmark}). It also reduces video-level trajectory ATE by $29$--$57\%$ relative to DVGT-1 (\cref{tab:geometry_all_public_benchmark}). Rig mAA@15 exceeds $83\%$ on Waymo and $99\%$ on nuScenes, showing that the model can estimate the camera arrangement alongside vehicle motion without supplied rig extrinsics or ego poses.

\begin{table}[!tp]
\centering
\footnotesize
\setlength{\tabcolsep}{3pt}

\caption{\textbf{Depth estimation on the Waymo and nuScenes validation splits.} Depth is evaluated independently in non-overlapping windows of 48 consecutive frames per camera, sampled at the native frame rates (Waymo: 10\,Hz; nuScenes: $\approx$12\,Hz).}

\label{tab:geometry_depth_public_benchmark}

\begin{tabular*}{\linewidth}{@{\extracolsep{\fill}}p{0.2\linewidth}rrrr@{}}

\toprule

\multirow{2}{*}{Model}
& \multicolumn{2}{c}{Waymo}
& \multicolumn{2}{c}{nuScenes} \\
\cmidrule(lr){2-3}\cmidrule(lr){4-5}
& AbsRel$\downarrow$ & $\delta_{1.25}\uparrow$ (\%)
& AbsRel$\downarrow$ & $\delta_{1.25}\uparrow$ (\%) \\

\midrule

DVGT-1~\citep{zuo2026dvgt}
    & 0.20 & 72.87 
    & 0.18 & 78.97 \\
DVGT-2~\citep{zuo2026dvgt2}
    & 0.26 & 70.93 
    & 0.22 & 75.32 \\
LiAuto-GeoX~\citep{lian2026liautogeox}
    & 0.19 & 75.72 
    & 0.27 & 62.75 \\
GeoUP~\citep{xu2026geoup}
    & 0.13 & 87.30 
    & 0.19 & 80.05 \\

\textbf{\method}
    & \textbf{0.09} & \textbf{91.97}  
    & \textbf{0.15} & \textbf{86.44}  \\

\bottomrule
\end{tabular*}
\end{table}

\begin{table}[!tp]
\centering
\footnotesize
\setlength{\tabcolsep}{3pt}

\caption{
\textbf{Video-level geometry comparison on the Waymo and nuScenes validation data described in \cref{sec:geometry_public_benchmark}, at native camera rates (10\,Hz and $\approx$12\,Hz).} Each clip uses one evaluation Sim(3), shared by poses and depth. Both models use our video inference system with 48-frame windows and a 5-frame overlap for stitching and PGO+LM. Rig metrics are omitted (--) because DVGT-1 does not estimate the rig.
}

\label{tab:geometry_all_public_benchmark}

\begin{tabular*}{\linewidth}{@{\extracolsep{\fill}}p{0.1\linewidth}p{0.12\linewidth}p{0.15\linewidth}rrrrrrrr@{}}

\toprule

\multirow{3}{*}{Dataset} 
& \multirow{3}{*}{Inference Mode} 
& \multirow{3}{*}{Model} 
& \multicolumn{3}{c}{Trajectory} 
& \multicolumn{3}{c}{Rig} 
& \multicolumn{2}{c}{Depth} \\

\cmidrule(lr){4-6}\cmidrule(lr){7-9}\cmidrule(lr){10-11}

&
&
& ATE$\downarrow$ & RTE$\downarrow$ & RRE$\downarrow$ 
& RTE$\downarrow$ & RRE$\downarrow$ & mAA@15$\uparrow$ 
& AbsRel$\downarrow$ & $\delta_{1.25}\uparrow$ \\

&
&
& (cm) & (cm) & ($^\circ$) 
& (cm) & ($^\circ$) & (\%) 
& & (\%) \\

\midrule

\multirow{4}{*}{Waymo} 
    & \multirow{2}{*}{Stitching} 
        & DVGT-1~\citep{zuo2026dvgt} 
        & 62.33 & 3.54 & 0.06
        & -- & -- & -- 
        & 0.20 & 73.04 \\
    &
        & \textbf{\method}
        & \textbf{32.70} & \textbf{3.20} & \textbf{0.06}
        & \textbf{0.96} & \textbf{1.02} & \textbf{83.37} 
        & \textbf{0.09} & \textbf{92.33} \\
\cmidrule(lr){2-11}
    & \multirow{2}{*}{PGO+LM} 
        & DVGT-1~\citep{zuo2026dvgt} 
        & 54.55 & 2.38 & \textbf{0.05} 
        & -- & -- & -- 
        & 0.20 & 73.26 \\
    &
        & \textbf{\method}
        & \textbf{23.62} & \textbf{1.56} & 0.06
        & \textbf{0.90} & \textbf{1.03} & \textbf{83.54} 
        & \textbf{0.09} & \textbf{92.94} \\

\midrule

\multirow{4}{*}{nuScenes} 
    & \multirow{2}{*}{Stitching} 
        & DVGT-1~\citep{zuo2026dvgt} 
        & 31.02 & 3.98 & 0.07 
        & -- & -- & -- 
        & 0.17 & 79.28 \\
    &
        & \textbf{\method}
        &\textbf{ 22.13} & \textbf{2.15} & \textbf{0.05}
        & \textbf{2.06} & \textbf{0.18} & \textbf{99.22} 
        & \textbf{0.15} & \textbf{86.58} \\
\cmidrule(lr){2-11}
    & \multirow{2}{*}{PGO+LM} 
        & DVGT-1~\citep{zuo2026dvgt}
        & 30.54 & 1.24 & 0.07
        & -- & -- & -- 
        & 0.17 & 79.28 \\
    &
        & \textbf{\method}
        & \textbf{17.48} & \textbf{1.19} & \textbf{0.05}
        & \textbf{2.02} & \textbf{0.18} & \textbf{99.22} 
        & \textbf{0.15} & \textbf{86.63} \\

\bottomrule
\end{tabular*}
\end{table}

\begin{table}[!tp]
\centering
\footnotesize
\setlength{\tabcolsep}{3pt}
\caption{\textbf{Geometry results on all 1,253 PAI test clips with six cameras.} $^\dagger$ denotes internal fine-tuning. Metrics and evaluation alignment follow \cref{sec:geometry_metrics}.}
\label{tab:geometry_pai_benchmark}
\begin{tabular*}{\linewidth}{@{\extracolsep{\fill}}p{0.17\linewidth}p{0.18\linewidth}rrrrrrrr@{}}
\toprule
\multirow{3}{*}{Model} & \multirow{3}{*}{Inference Mode} & \multicolumn{3}{c}{Trajectory} & \multicolumn{3}{c}{Rig} & \multicolumn{2}{c}{Depth} \\
\cmidrule(lr){3-5}\cmidrule(lr){6-8}\cmidrule(lr){9-10}
& & ATE$\downarrow$ & RTE$\downarrow$ & RRE$\downarrow$ & RTE$\downarrow$ & RRE$\downarrow$ & mAA@15$\uparrow$ & AbsRel$\downarrow$ & $\delta_{1.25}\uparrow$ \\
& & (cm) & (cm) & ($^\circ$) & (cm) & ($^\circ$) & (\%) & & (\%) \\
\midrule
VGGT$^\dagger$ & Stitching & 111.12 & 8.31 & 0.122 & 29.26 & 0.520 & 42.50 & 0.084 & 94.88 \\
\midrule
\multirow{2}{*}{\method} & Stitching & 78.06 & 4.69 & 0.070 & \textbf{3.90} & \textbf{0.404} & \textbf{88.72} & 0.081 & 94.98 \\
& PGO+LM & \textbf{52.75} & \textbf{2.28} & \textbf{0.066} & 3.92 & \textbf{0.404} & 88.71 & \textbf{0.079} & \textbf{95.03} \\
\bottomrule
\end{tabular*}
\end{table}

\begin{figure}[th!]
\centering
\includegraphics[width=\linewidth]{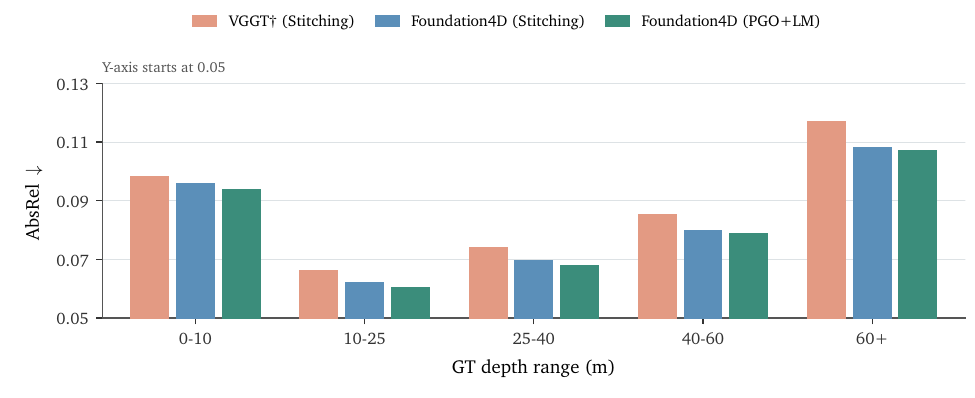}
\vspace{-10.00mm}
\caption{\textbf{Depth error by ground-truth depth range on PAI, averaged equally over 1,253 clips in each bin.} AbsRel is dimensionless; the y-axis starts at $0.05$. \method stitching uses sky masks; VGGT stitching and PGO+LM do not. Each configuration uses its shared full-clip Sim(3) evaluation scale, with no per-bin alignment. $^\dagger$ denotes internal fine-tuning.}
\label{fig:geometry_pai_depth_ranges}
\end{figure}

\subsubsection{PAI benchmark}
\label{sec:geometry_pai_benchmark}

We evaluate \method on the PAI benchmark introduced in \cref{sec:pai_benchmark}. The evaluation covers ego poses, camera-to-rig calibration, and depth across distance ranges, complemented by qualitative scene reconstructions.

\paragraph{Setup.}
We evaluate the unified \method model on geometry tasks using the full PAI test split of 1,253 clips and six cameras, comparing against an internally fine-tuned VGGT baseline. We fine-tune the original VGGT architecture on the same internal data using our best geometry specialist training schedule. We retain the official VGGT-Commercial initialization, which yields slightly better geometry performance than LingBot-Map initialization in our experiments. Because VGGT lacks semantic masks and predicts geometry up to scale, we evaluate it only with stitching without sky masking. \method uses sky-masked stitching and PGO+LM. Both models use the same window sampling schedule; metrics follow the shared geometry evaluation protocol and are averaged equally across clips.

\paragraph{Ego and rig calibration results.}
With stitching, \method reduces trajectory ATE by $29.8\%$ relative to the internally fine-tuned VGGT baseline and improves rig calibration (\cref{tab:geometry_pai_benchmark}). Rig translation error decreases by $86.7\%$, while rig mAA@15 rises from $42.50\%$ to $88.72\%$. PGO+LM further reduces trajectory ATE from $78.06$ to $52.75$\,cm.

\begin{figure}[t]
\centering
\includegraphics[width=0.98\linewidth]{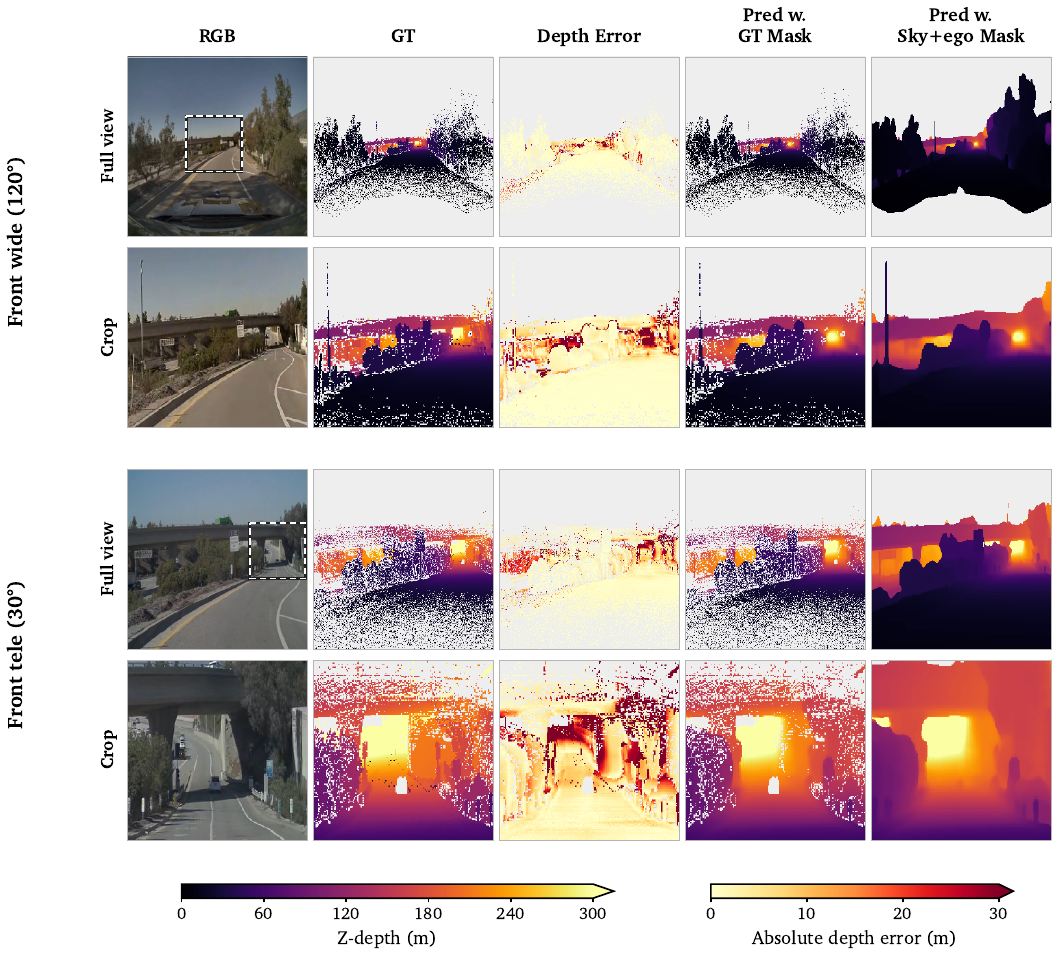}
\vspace{-2.00mm}
\caption{\textbf{\method depth predictions with PGO+LM from the front-wide and front-tele cameras of a PAI clip.} Each full view is followed by a $3.2\!\times$ enlargement of the region boxed in the RGB image. Absolute depth error and GT-masked predictions use valid GT support; the final column excludes predicted sky and ego-vehicle pixels from the dense prediction. Gray denotes masked pixels. Depth and error colors saturate at $300$\,m and $30$\,m, respectively. Predictions retain the shared full-clip pose-derived scale without local realignment.}
\label{fig:geometry_pai_far_range}
\end{figure}

\paragraph{Depth accuracy across ranges.}
Both \method inference modes reduce AbsRel in every depth range relative to VGGT (\cref{fig:geometry_pai_depth_ranges}). The relative improvement is larger at long range: with stitching, it grows from $2.2\%$ within $10$\,m to $7.7\%$ beyond $60$\,m; PGO+LM reaches an $8.4\%$ reduction in the far range. The example in \cref{fig:geometry_pai_far_range} shows multi-camera depth reconstruction of an overpass and the road beyond it, while highlighting residual errors around distant structures and depth discontinuities.

\begin{figure}[!tp]
\centering
\begingroup
\setlength{\tabcolsep}{0pt}
\renewcommand{\arraystretch}{1}
\newcommand{\paipointcloudrow}[3]{%
  \multicolumn{2}{l}{\small\bfseries #1}\\[2pt]
  \includegraphics[width=.493\linewidth]{assets/geometry_pai_pointclouds/#2} &%
  \includegraphics[width=.493\linewidth]{assets/geometry_pai_pointclouds/#3}\\[5pt]
}
\begin{tabular}{@{}c@{\hspace{.014\linewidth}}c@{}}
\small\bfseries RGB point cloud & \small\bfseries Semantic point cloud\\[5pt]
\paipointcloudrow{(a) City street}{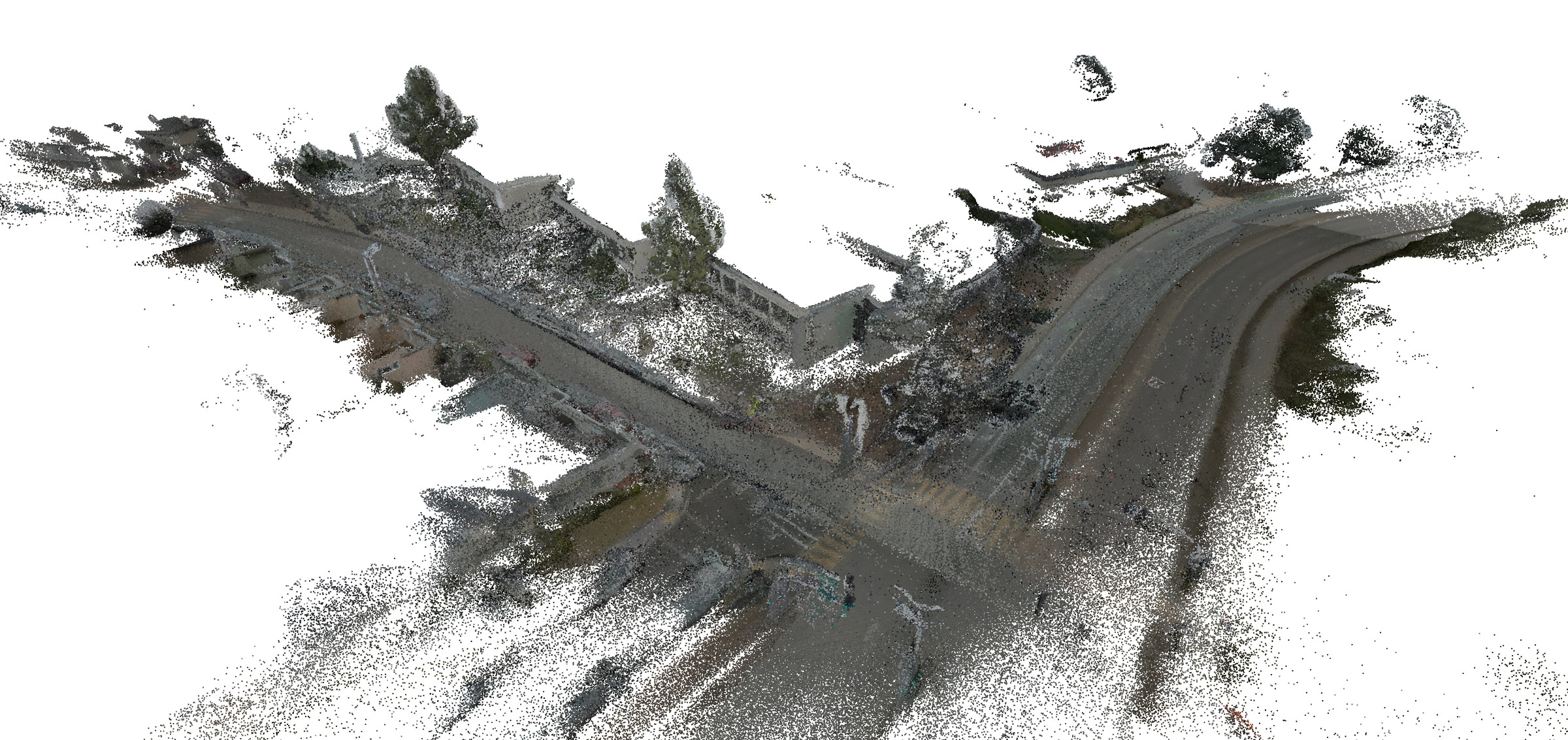}{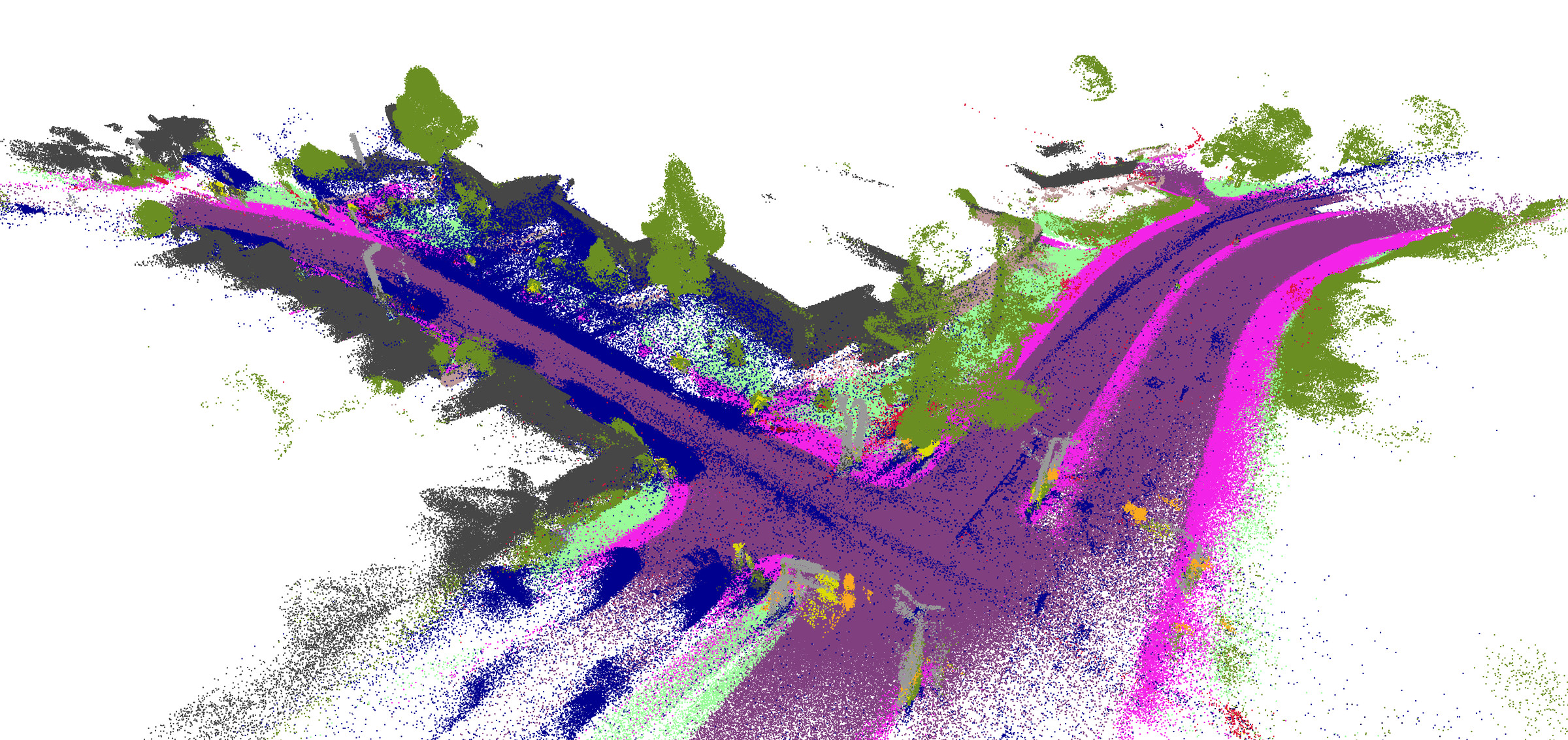}
\paipointcloudrow{(b) Under the overpass}{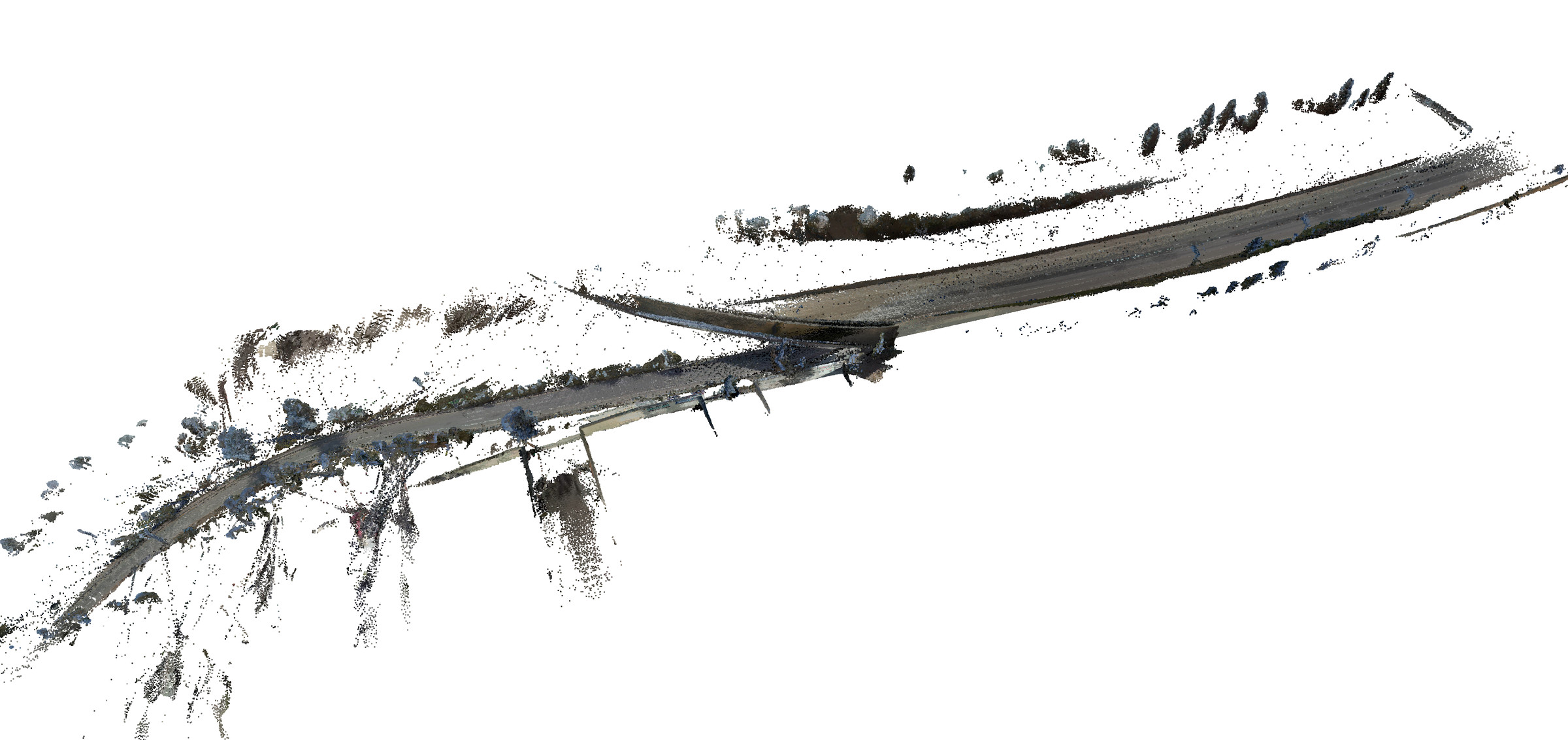}{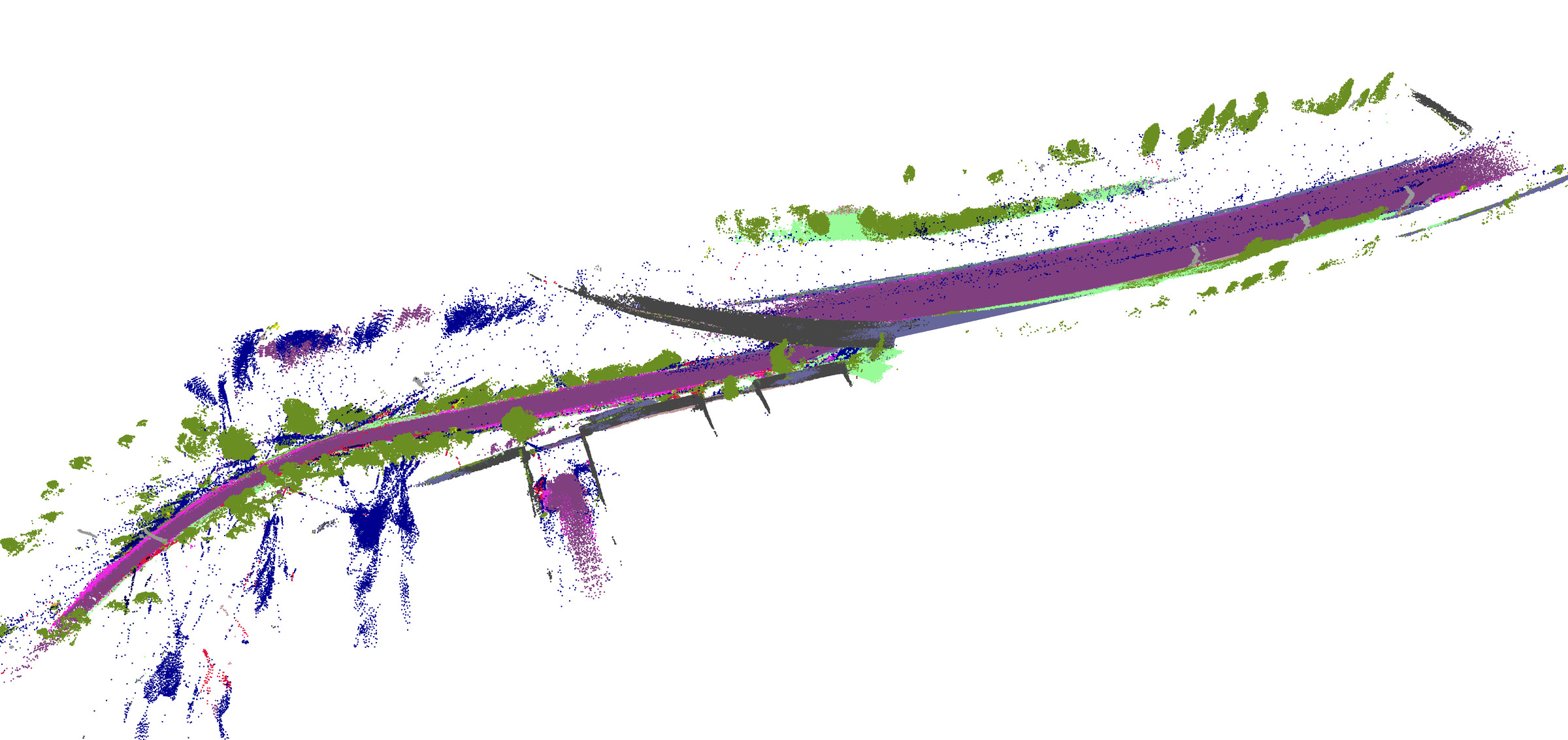}
\paipointcloudrow{(c) Commercial boulevard}{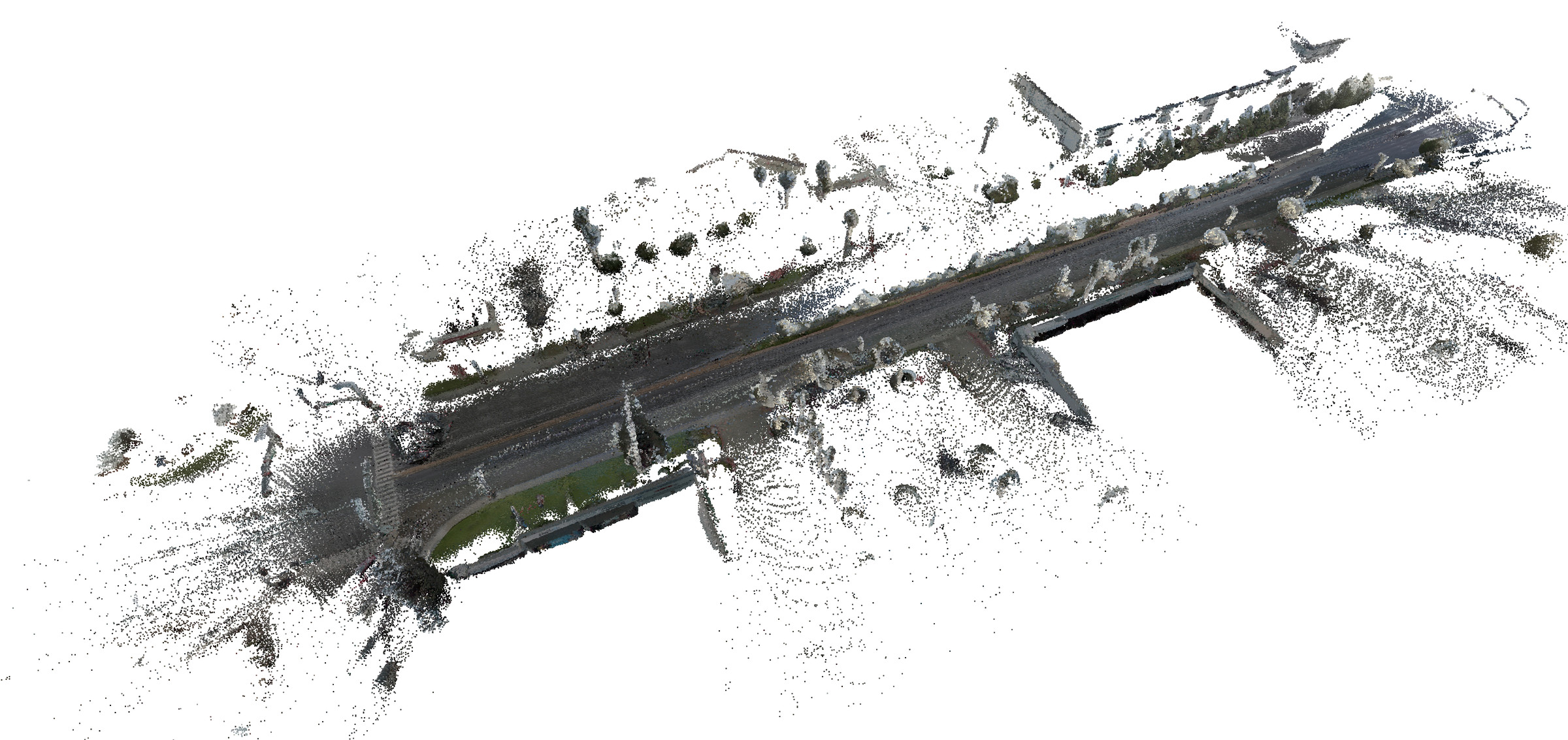}{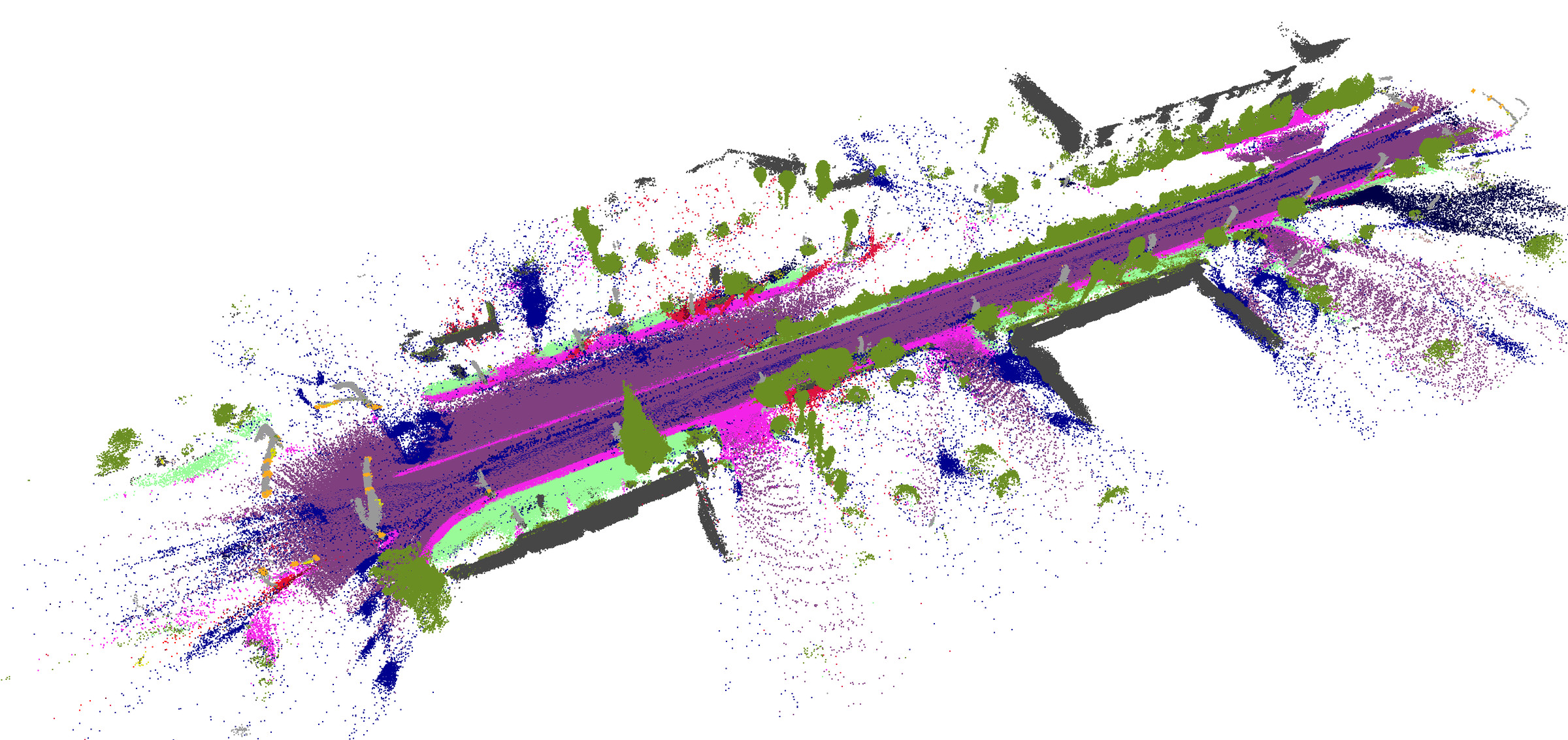}
\paipointcloudrow{(d) Downtown facade}{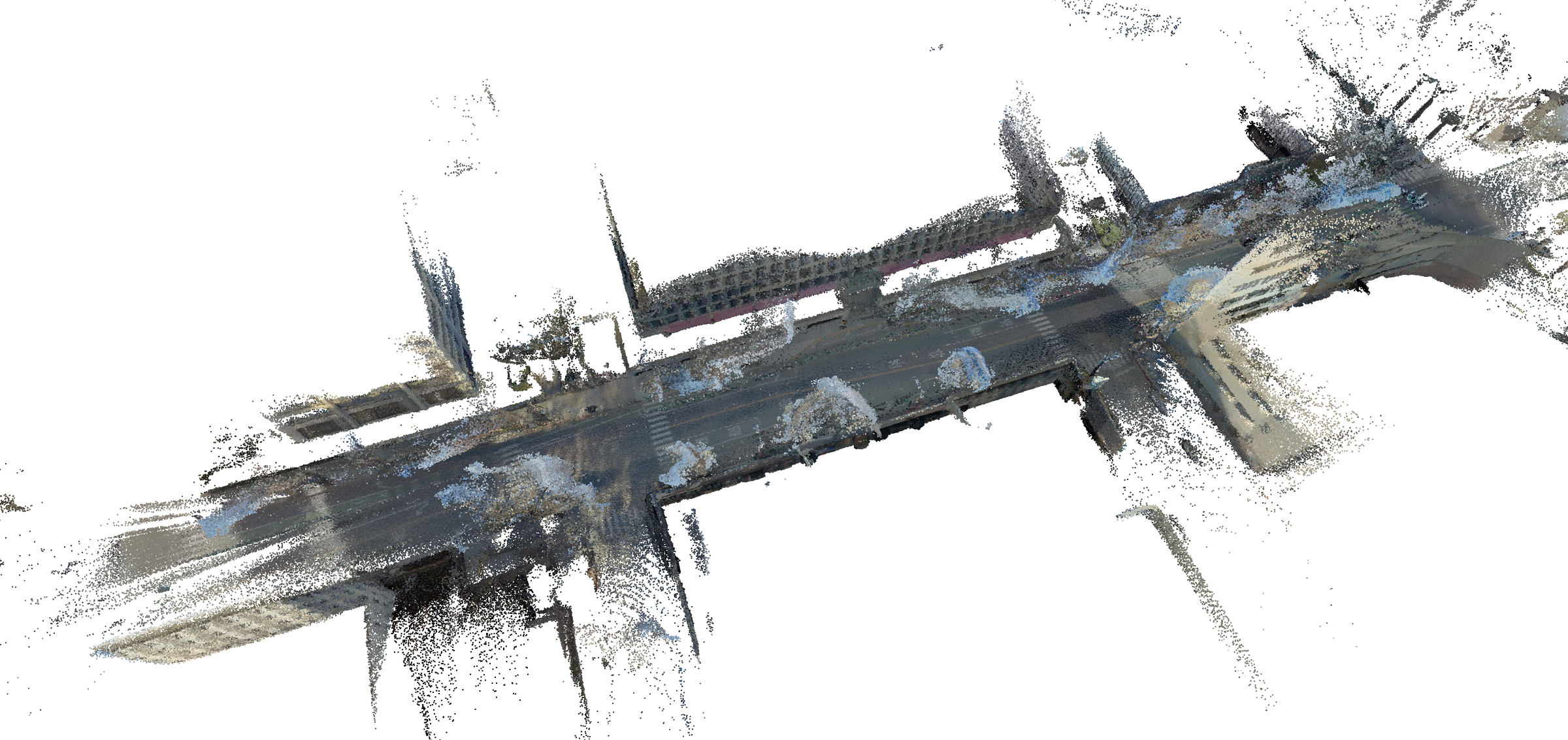}{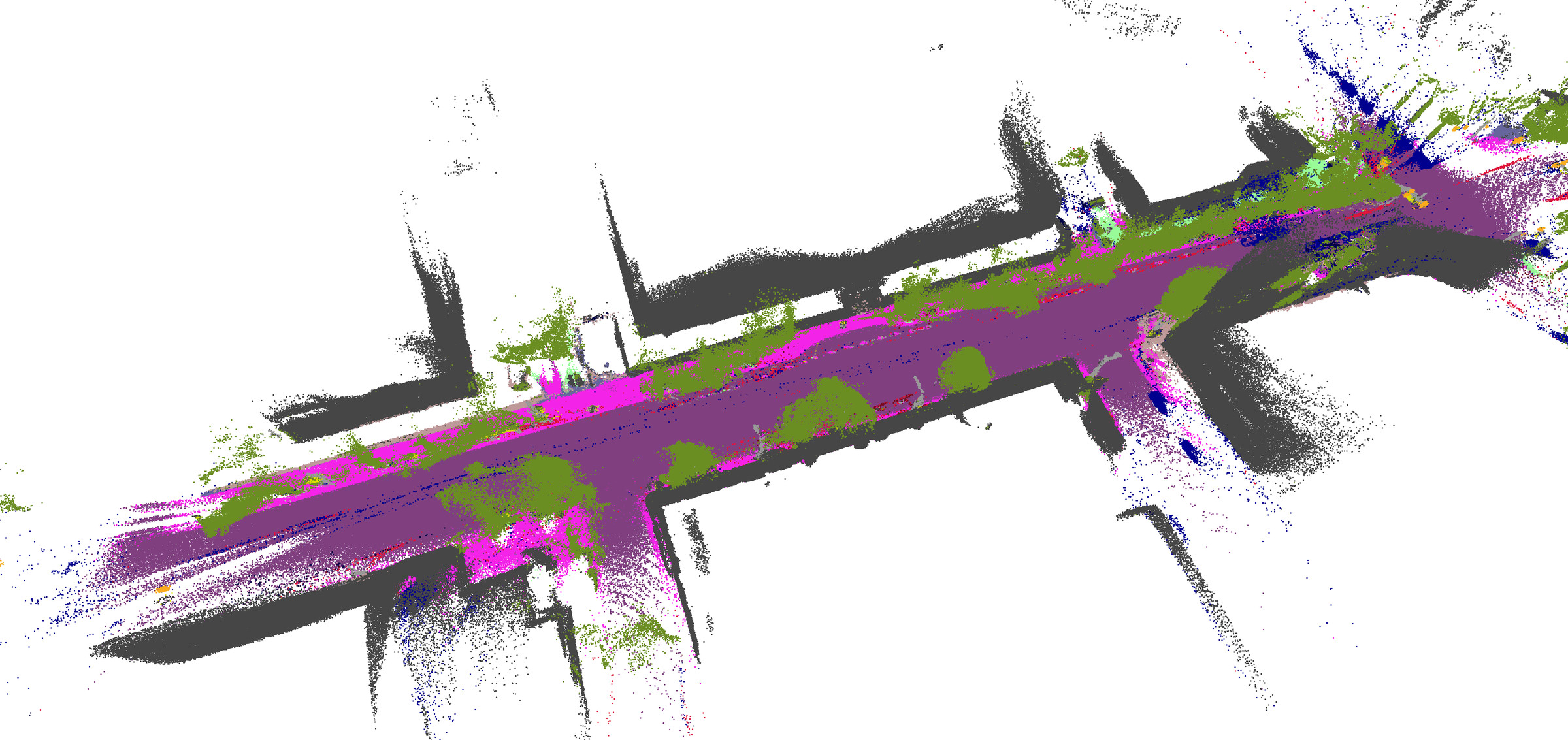}
\end{tabular}
\par\vspace{1pt}
\newcommand{\paipointcloudclass}[2]{%
  \textcolor[RGB]{#1}{\rule[-.1ex]{.8em}{.8em}}\hspace{.3em}#2%
}
{\scriptsize Selected classes:\quad
\paipointcloudclass{128,64,128}{road}\quad
\paipointcloudclass{244,35,232}{sidewalk}\quad
\paipointcloudclass{70,70,70}{building}\quad
\paipointcloudclass{107,142,35}{vegetation}\quad
\paipointcloudclass{152,251,152}{terrain}\quad
\paipointcloudclass{0,0,142}{car}}
\endgroup
\caption{\textbf{Video-level point-cloud reconstructions on PAI using \method with PGO+LM.} Each scene accumulates six-camera predictions over an approximately $20$\,s clip. Before confidence-filtered depth fusion, we remove moving objects using external annotation-assisted masks and suppress sky and ego-vehicle pixels using predicted semantics. \cref{fig:geometry_pai_pointcloud_details} shows a detailed reconstruction of the overpass scene with its RGB input.}
\label{fig:geometry_pai_pointclouds}
\end{figure}

\begin{figure}[!tp]
\centering
\begingroup
\setlength{\tabcolsep}{0pt}
\renewcommand{\arraystretch}{1}
\begin{tabular}{@{}c@{\hspace{.014\linewidth}}c@{\hspace{.014\linewidth}}c@{}}
\small\bfseries RGB input & \small\bfseries RGB point cloud & \small\bfseries Semantic point cloud\\[4pt]
\includegraphics[width=.324\linewidth]{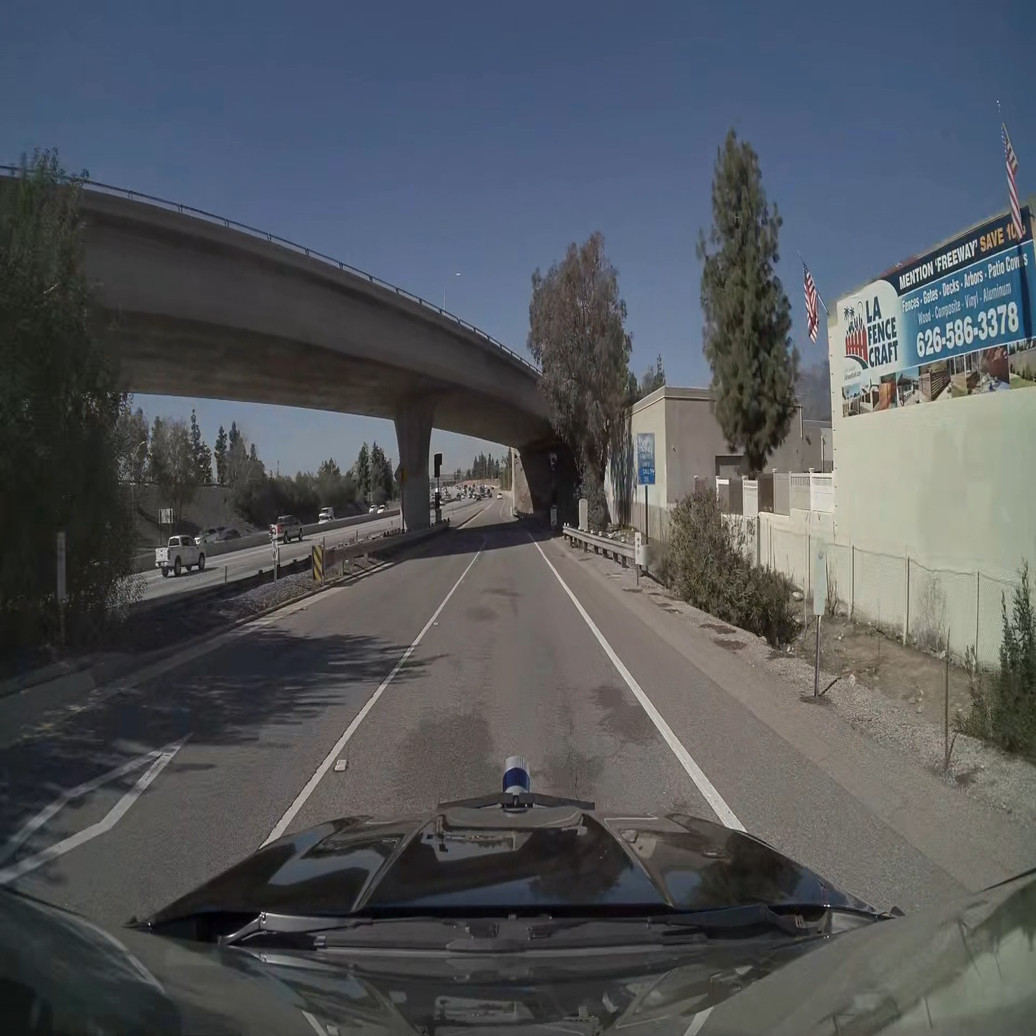} &%
\includegraphics[width=.324\linewidth]{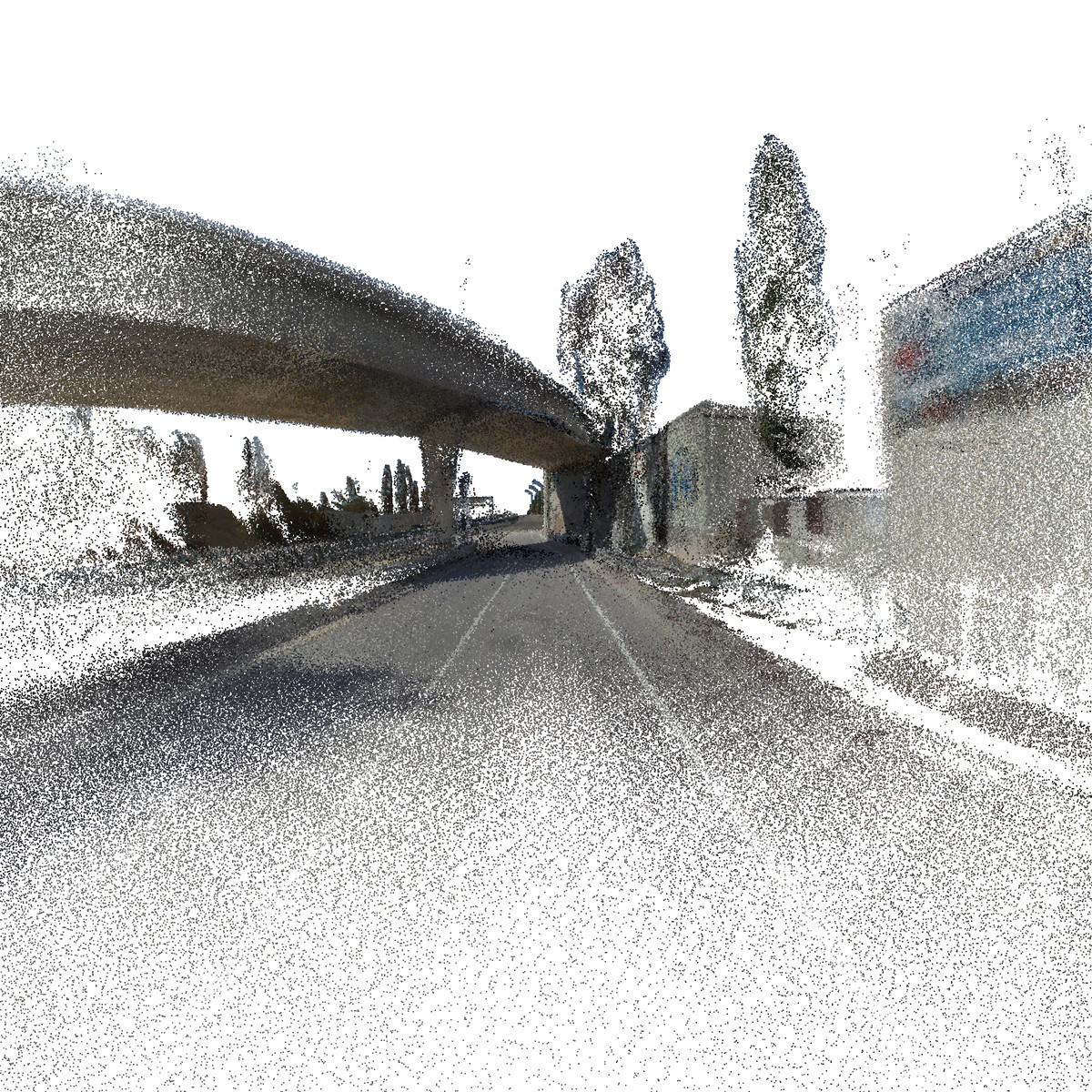} &%
\includegraphics[width=.324\linewidth]{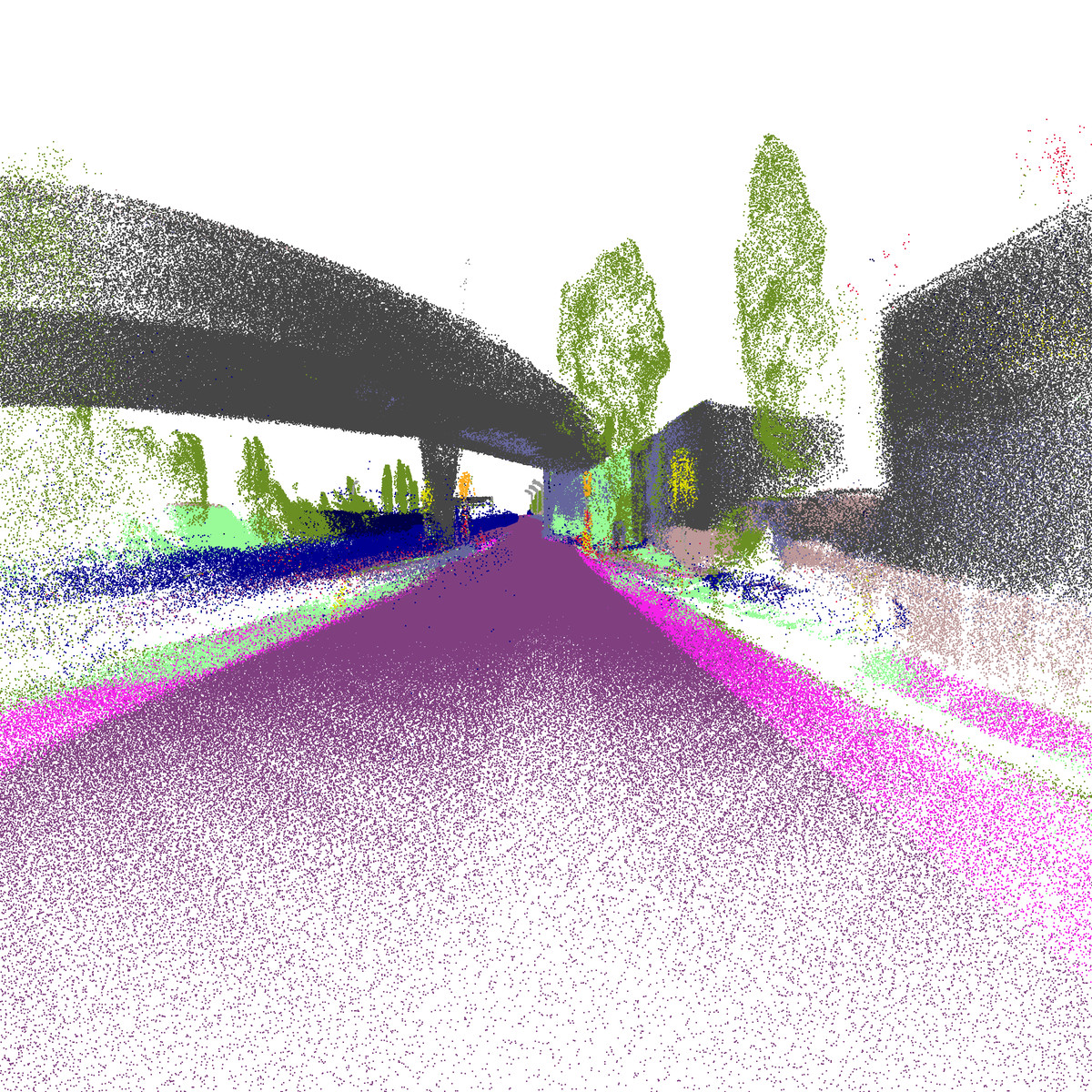}
\end{tabular}
\endgroup
\caption{\textbf{Detailed reconstruction of the overpass scene in \cref{fig:geometry_pai_pointclouds}(b).} The front-wide RGB input at $\tau=8.67$\,s is paired with RGB and semantic point-cloud renderings using the corresponding predicted camera pose and calibrated projection. Both renderings show the same geometry, accumulated from all six cameras across the full clip using depth fusion with dynamic-point restoration. White regions contain no rendered points.}
\label{fig:geometry_pai_pointcloud_details}
\end{figure}

\paragraph{Qualitative results on video-level reconstruction.}
Beyond per-image depth, \cref{fig:geometry_pai_pointclouds} shows four video-level reconstructions whose road layouts, building facades, and vegetation share a common 3D frame. The overpass detail in \cref{fig:geometry_pai_pointcloud_details} pairs one input image with accumulated geometry rendered from the corresponding predicted camera pose, showing the bridge, supporting pier, and lower roadway. The input captures one instant, while the reconstruction integrates observations across the clip.

\FloatBarrier

\subsubsection{Ablations}
\label{sec:geometry_ablations}

We ablate the geometry architecture and supervision to assess their contributions to depth, ego-pose, and rig-calibration accuracy. We also compare window stitching and pose-graph optimization to understand how inference choices affect video-level reconstruction.

\paragraph{Setup.}
\label{sec:geometry_ablation_setup}
All variants share a reduced 16-block aggregator, the same training samples and recipe, and LingBot-Map initialization. Each is trained for 30,000 updates with 12 temporal samples per camera and a uniformly sampled stride up to 5. Starting from a VGGT~\cite{wang2025vggt} baseline adapted to this setup, we cumulatively add factorized global attention, rig-ego factorization with separate rig and ego losses, metric-scale training, relative-pose supervision, and auxiliary semantic supervision. Stitching without semantic masks provides a common inference protocol supported by every variant, enabling a fair model comparison. For inference ablations, we fix variant (f), which achieves the lowest ATE and depth AbsRel, and compare stitching without and with sky masks, PGO, and PGO with a linear-motion prior (LM).

We evaluate on 50 internal clips using four cameras (front-wide, front-tele, cross-left, and cross-right), with RGB and depth at $518\times518$. We sample every second frame from the 15\,FPS anchor timeline and process complete clips in 48-timestamp windows per camera with a 5-timestamp overlap. Metrics are averaged equally across clips. This cohort is separate from the joint-task study in \cref{sec:joint_ablation}.

\paragraph{Impact of different model and supervision designs.}
\label{sec:geometry_model_ablation}

\begin{table}[!tp]
\centering
\footnotesize
\setlength{\tabcolsep}{3pt}
\caption{\textbf{Cumulative geometry ablation on 50 internal clips with stitching and no semantic masks.} Row (c) introduces rig-ego factorization with separate rig and ego losses. ATE uses one multiview Sim(3) alignment per clip, whose scale is also applied to depth.}
\label{tab:geometry_components_ablation}
\begin{tabular*}{\linewidth}{@{\extracolsep{\fill}}p{0.27\linewidth}rrrrrrrr@{}}
\toprule
\multirow{3}{*}{Variant} & \multicolumn{3}{c}{Trajectory} & \multicolumn{3}{c}{Rig} & \multicolumn{2}{c}{Depth} \\
\cmidrule(lr){2-4}\cmidrule(lr){5-7}\cmidrule(lr){8-9}
& ATE$\downarrow$ & RTE$\downarrow$ & RRE$\downarrow$ & RTE$\downarrow$ & RRE$\downarrow$ & mAA@15$\uparrow$ & AbsRel$\downarrow$ & $\delta_{1.25}\uparrow$ \\
& (cm) & (cm) & ($^\circ$) & (cm) & ($^\circ$) & (\%) & & (\%) \\
\midrule
(a) Baseline & 128.48 & 5.04 & 0.086 & 7.75 & \textbf{0.701} & 71.78 & 0.133 & 83.16 \\
(b) + factorized global attention & 149.88 & 4.87 & 0.086 & 7.16 & 0.808 & 74.24 & 0.140 & 82.25 \\
(c) + rig-ego factorization & 141.72 & 4.31 & 0.053 & \textbf{5.00} & 0.796 & \textbf{77.91} & 0.136 & 82.51 \\
(d) + metric-scale training & 70.67 & 3.43 & 0.053 & 5.88 & 0.834 & 76.98 & 0.129 & 83.58 \\
(e) + relative-pose loss & 69.81 & 3.22 & \textbf{0.042} & 5.62 & 0.835 & 76.91 & 0.130 & 83.41 \\
(f) + semantic supervision & \textbf{69.53} & \textbf{2.94} & \textbf{0.042} & 5.77 & 0.802 & 75.69 & \textbf{0.127} & \textbf{83.62} \\
\bottomrule
\end{tabular*}
\end{table}

Adding factorized global attention to the baseline reduces observed end-to-end runtime by $8.5\%$ on one H100 GPU, with a trade-off in video-level and depth accuracy (\cref{tab:geometry_components_ablation}). The rig-ego variant improves trajectory, rig, and depth accuracy, yielding the lowest rig translation error and highest rig mAA@15 among the variants. These gains support explicitly modeling the static camera arrangement alongside temporal motion. Metric-scale training provides the largest gain in video-level reconstruction, approximately halving ATE from $141.72$ to $70.67$\,cm while improving depth. This supports supervising windows in a common unit to improve their assembly. Relative-pose supervision then improves local translation and rotation accuracy, with a small increase in depth error. Auxiliary semantic supervision yields the best depth accuracy and further improves trajectory translation, although rig accuracy does not improve uniformly. Together, the results distinguish complementary contributions to scale consistency, local motion, and dense geometry.

\paragraph{Impact of different inference modes.}
\label{sec:geometry_video_ablation}

\begin{table}[!tp]
\centering
\footnotesize
\setlength{\tabcolsep}{3pt}
\caption{\textbf{Video assembly with stitching and pose graph optimization.} All variants use checkpoint (f) and the protocol in \cref{tab:geometry_components_ablation}, including one multiview Sim(3) evaluation alignment per clip with a shared pose/depth scale. Sky masks filter stitching correspondences; PGO uses capture-time corrections without pixel-alignment masks. LM denotes the linear-motion prior.}
\label{tab:geometry_inference_ablation}
\begin{tabular*}{\linewidth}{@{\extracolsep{\fill}}p{0.27\linewidth}rrrrrrr@{}}
\toprule
\multirow{3}{*}{Variant} & \multicolumn{3}{c}{Trajectory} & \multicolumn{2}{c}{Rig} & \multicolumn{2}{c}{Depth} \\
\cmidrule(lr){2-4}\cmidrule(lr){5-6}\cmidrule(lr){7-8}
& ATE$\downarrow$ & RTE$\downarrow$ & RRE$\downarrow$ & RTE$\downarrow$ & RRE$\downarrow$ & AbsRel$\downarrow$ & $\delta_{1.25}\uparrow$ \\
& (cm) & (cm) & ($^\circ$) & (cm) & ($^\circ$) &  & (\%) \\
\midrule
Stitching, no mask & 69.53 & 2.94 & 0.042 & 5.77 & 0.802 & 0.127 & 83.62 \\
Stitching, sky mask & 65.55 & 2.89 & 0.042 & 5.75 & 0.802 & 0.127 & 83.68 \\
PGO & 33.11 & 2.18 & \textbf{0.041} & 5.64 & \textbf{0.801} & \textbf{0.125} & \textbf{84.02} \\
PGO + LM & \textbf{31.26} & \textbf{1.21} & \textbf{0.041} & \textbf{5.60} & \textbf{0.801} & \textbf{0.125} & 84.01 \\
\bottomrule
\end{tabular*}
\end{table}

\begin{figure}[!tp]
\centering
\includegraphics[width=\linewidth]{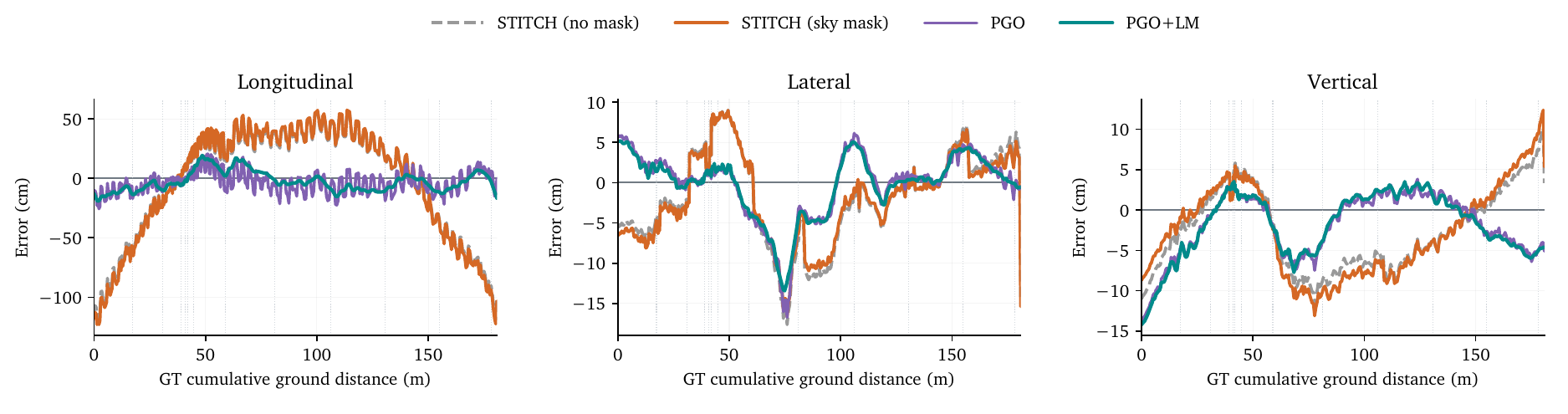}
\caption{\textbf{Camera-position errors under different video assembly methods.} In this example sequence, each method uses one full-clip multiview Sim(3) alignment shared across four cameras. Signed front-wide residuals (cm) are projected onto the ground-truth forward, left, and world-up axes and plotted against cumulative ground distance (m). Dotted lines mark window starts; all frame samples are shown without smoothing.}
\label{fig:geometry_inference_trajectory}
\end{figure}

In \cref{tab:geometry_inference_ablation}, sky masking modestly improves stitching by excluding unreliable finite-depth correspondences from window alignment. PGO then nearly halves ATE relative to sky-masked stitching and improves local motion accuracy. Jointly reconciling overlapping relative-pose constraints allows later windows to revise earlier estimates, reducing error accumulation from sequential alignment. The linear-motion prior further improves translation consistency by discouraging abrupt velocity changes, with a smaller additional gain in ATE. PGO+LM achieves the lowest ATE and relative translation error. On the example sequence in \cref{fig:geometry_inference_trajectory}, PGO reduces the large longitudinal deviations of stitching, while LM suppresses short-interval oscillations. PGO retains the predicted depth; differences in its depth scores reflect the pose-derived scale used for evaluation.

\subsection{3D object detection and tracking}
\label{sec:detection_experiments}

We evaluate 3D object detection and the trajectories assembled from its predictions. We study adaptation to nuScenes, performance on the full PAI test split, and the effects of detector design and offline inference through ablations.

\subsubsection{Public benchmarks}
\label{sec:detection_public_benchmarks}

We evaluate \method on the public nuScenes~\cite{caesar2020nuscenes} 3D detection benchmark and compare it with state-of-the-art camera-only 3D detectors under the official protocol.

\paragraph{Setup.}
We initialize from the \method detector pre-trained on our internal NRE, PlanNet, and MLMCF data without nuScenes, re-initialize its classifier for the 10 nuScenes classes, and fine-tune all weights. For a fair comparison with prior methods, we follow the standard nuScenes training and evaluation protocol, using 2\,Hz annotated keyframes from all six cameras across 700 training scenes and 150 validation scenes (evaluation only).

We use the image and BEV augmentations and GridMask of StreamPETR~\cite{wang2023exploring} and GeoUP~\cite{xu2026geoup}, with the same $\pm51.2$\,m detection range.
We train for 10,000 iterations using AdamW with a batch size of 32 windows and a base learning rate of $4.5\times10^{-4}$, applying learning-rate multipliers of 0.1 and 0.03 to the aggregator and image encoder, respectively.
Following prior work, we evaluate predictions on all 150 validation scenes using the official nuScenes devkit and report mAP and the nuScenes detection score (NDS), ensuring a consistent evaluation protocol for the comparisons in \cref{tab:detection_pub_nuscenes}. At test time, \method processes consecutive eight-keyframe windows, so its predictions also use future frames within each window.

\paragraph{Baselines.}
We compare with PETRv2~\cite{liu2023petrv2}, SparseBEV~\cite{liu2023sparsebev}, Stream3DPPE~\cite{shu20233dppe}, RoPETR~\cite{ji2025ropetr}, StreamPETR~\cite{wang2023exploring}, and RayDN~\cite{liu2024ray} using their published results. The most related method, GeoUP~\citep{xu2026geoup}, combines a VGGT-initialized~\cite{wang2025vggt} ViT-L backbone with a RayDN-style detector and four-frame inputs. We evaluate the released GeoUP-nuScenes checkpoint, trained exclusively on nuScenes, for which the paper reports 57.9 mAP and 64.4 NDS.
We also include the published results of GeoUP-Joint\footnote{The available implementation details are currently insufficient to reproduce GeoUP-Joint's results.}, trained jointly on nuScenes, Argoverse~2, Waymo, KITTI, and DDAD with detection, depth, and occupancy supervision.
For RayDN, we report GeoUP's reproduced results using a DINOv3~\cite{simeoni2025dinov3} ViT-L backbone.

\paragraph{Results.}
\method achieves the best results, 60.9 mAP and 66.5 NDS (\cref{tab:detection_pub_nuscenes}). It improves over the second-best method GeoUP-Joint by +1.7 mAP and +1.2 NDS. This suggests that the \method checkpoint is a strong starting point for new camera rigs and label sets. A larger generic backbone alone does not explain the gains: with a DINOv3 ViT-L, RayDN reaches only 49.9 mAP, below its EVA02-L version, whereas both geometry-pretrained ViT-L models exceed 56 mAP.

\begin{table}[!tp]
\centering
\small
\setlength{\tabcolsep}{4pt}
\renewcommand{\arraystretch}{1.13}
\caption{
\textbf{Camera-only 3D object detection performance on the nuScenes validation set (150 scenes).} All models are trained using supervision on the 700-scene nuScenes training set.
}
\label{tab:detection_pub_nuscenes}
\begin{tabular*}{\linewidth}{@{\extracolsep{\fill}}lllrr@{}}
\toprule

\multirow{2}{*}{Model} 
& \multicolumn{2}{c}{Backbone}
& \multicolumn{2}{c}{Detection} \\

\cmidrule(lr){2-3} \cmidrule(lr){4-5}

 & \multicolumn{1}{c}{Architecture} & \multicolumn{1}{c}{Backbone Init.} & mAP $\uparrow$ & NDS $\uparrow$\\
 
\midrule
	
PETRv2 & ResNet101 & nuImages & 42.1 & 52.4 \\

SparseBEV & ResNet101 & nuImages & 50.1 & 59.2 \\

Stream3DPPE & VoV-99 & FCOS3D & 50.0 & 58.5 \\

RoPETR & VoV-99 & FCOS3D & 52.9 & 61.4 \\

StreamPETR & EVA02-L & Cascade R-CNN & 52.1 & 60.8 \\

RayDN & EVA02-L & Cascade R-CNN & 54.1 & 62.4 \\

\midrule

RayDN & ViT-L & DINOv3 & 49.9 & 55.3 \\

GeoUP-nuScenes & ViT-L & VGGT & 56.5 & 63.0 \\

GeoUP-Joint & ViT-L & VGGT & 59.2 & 65.3 \\

\midrule

\textbf{\method} & ViT-L & \method & \textbf{60.9} & \textbf{66.5} \\

\bottomrule
\end{tabular*}
\end{table}

\subsubsection{PAI benchmark}
\label{sec:detection_pai_benchmark}

\paragraph{Setup.}
We evaluate the unified \method checkpoint on all 1,253 PAI test clips, covering 750,552 timestamps, using the checkpoint obtained after Gaussian-reconstruction fine-tuning.
We use six physical cameras at $1036\times1036$ resolution and 32-timestamp windows with spacing of two source frames during inference. Interleaved sampling covers every source timestamp. Predictions are evaluated against the officially released PAI cuboid annotations~\cite{nvidia2025physicalai-av} (see the classes listed in \cref{sec:detection_decoder}), restricted to objects visible in the selected cameras.
Detection uses BEV center-distance matching at thresholds of $0.5$, $1$, $2$, and $4$\,m. Tracking uses the full offline pipeline and center-distance HOTA~\cite{luiten2021hota}, with similarity decreasing to zero at $2$\,m. To focus on the most representative and common classes, we report the five-class macro average over automobile, person, rider, bus, and heavy truck, together with the all-twelve-class result. Tracking statistics are combined over clips within each class before taking the unweighted class mean.

\paragraph{Results and distance breakdown.}

\begin{figure}[!tp]
\centering
\includegraphics[width=\linewidth]{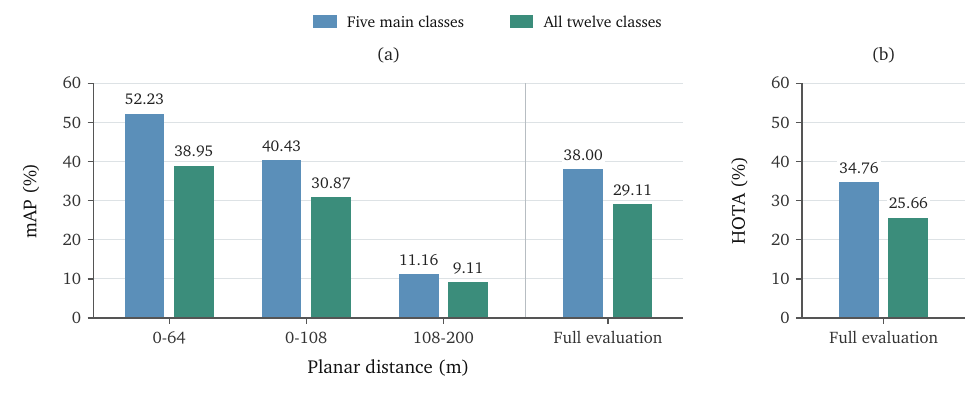}
\caption{\textbf{\method detection and tracking on the PAI test split.} (a) Detection mAP by planar distance from the ego vehicle, with full-evaluation scores shown separately. Each distance interval independently filters predicted and ground-truth box centers. (b) Tracking HOTA over the full evaluation, using center-distance matching. Colors distinguish averages over the five main classes and all twelve classes.}
\label{fig:detection_pai_ranges}
\end{figure}

\method achieves $38.00\%$ mAP and $34.76\%$ HOTA over the five main classes, and $29.11\%$ mAP and $25.66\%$ HOTA over all twelve classes in the full evaluation. \cref{fig:detection_pai_ranges} reports detection mAP by planar distance from the ego vehicle and tracking HOTA over the full evaluation. For the five main classes, mAP is $52.23\%$ within $64\,\mathrm{m}$, $40.43\%$ within $108\,\mathrm{m}$, and $11.16\%$ in the $108$--$200\,\mathrm{m}$ interval. The corresponding scores over all twelve classes are $38.95\%$, $30.87\%$, and $9.11\%$, respectively. The two near-range evaluations overlap and are scored independently. These results show the remaining difficulty of long-range detection. Since the evaluation regions and class definitions differ from nuScenes, the reported mAP scores should not be compared directly with its headline results.

\begin{figure*}[t]
\centering
\includegraphics[width=\linewidth]{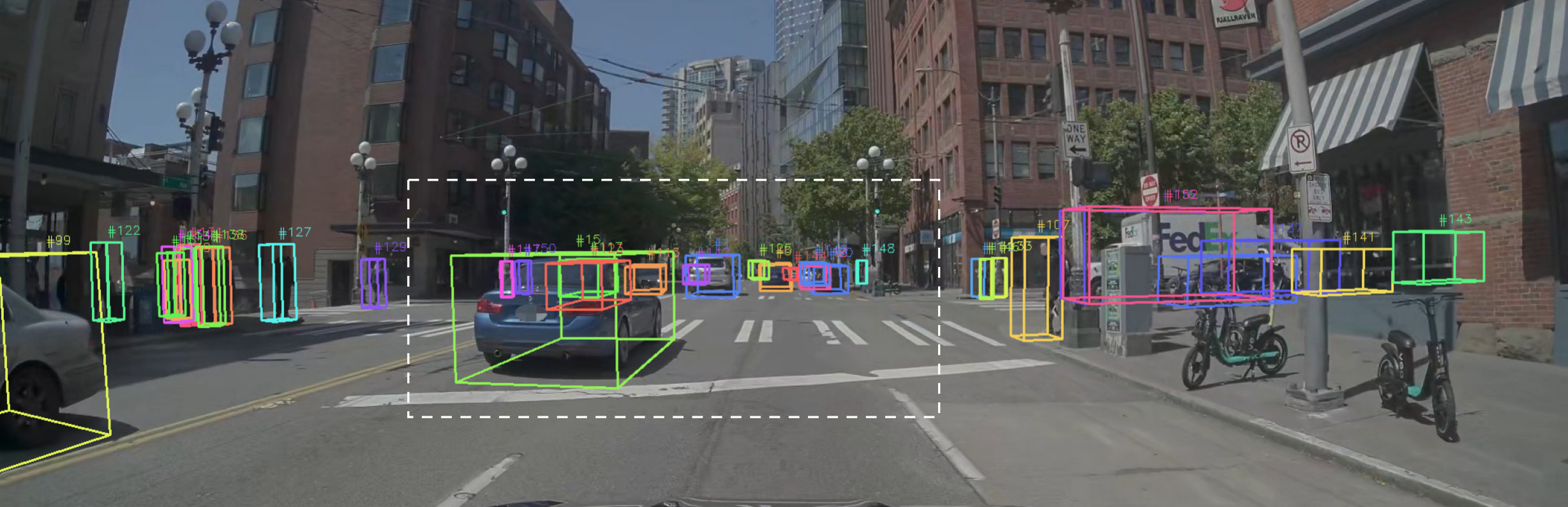}
\par\vspace{3pt}
\noindent
\begin{minipage}[t]{0.325\linewidth}
\centering
\includegraphics[width=\linewidth]{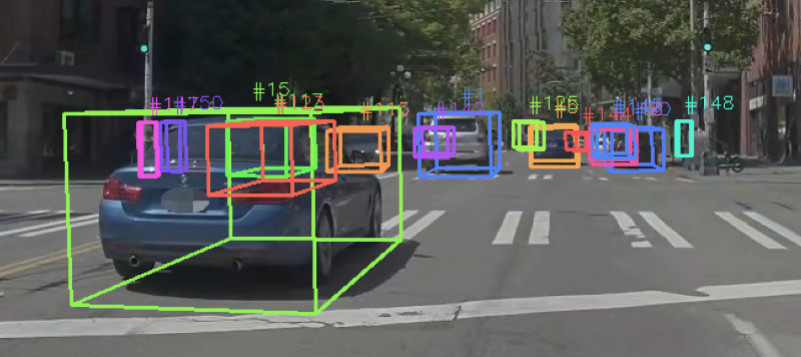}
\par\vspace{1pt}
{\small $8\,\mathrm{s}$}
\end{minipage}\hfill%
\begin{minipage}[t]{0.325\linewidth}
\centering
\includegraphics[width=\linewidth]{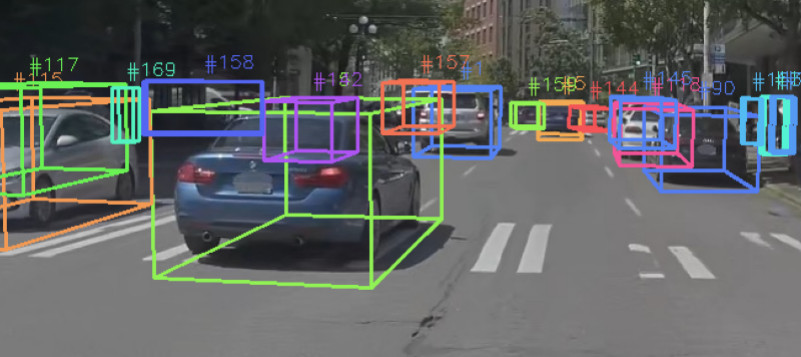}
\par\vspace{1pt}
{\small $10\,\mathrm{s}$}
\end{minipage}\hfill%
\begin{minipage}[t]{0.325\linewidth}
\centering
\includegraphics[width=\linewidth]{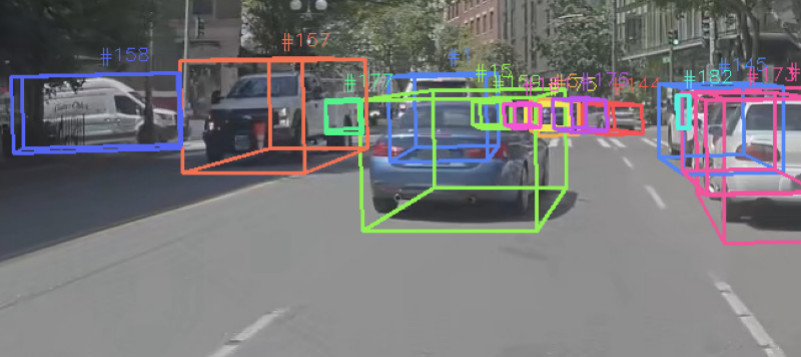}
\par\vspace{1pt}
{\small $12\,\mathrm{s}$}
\end{minipage}
\caption{\textbf{\method detection and tracking on the PAI dataset.} \method offline tracks are projected into the front-wide camera. Top: the overview at $8\,\mathrm{s}$; the dashed rectangle marks the region shown in the first temporal crop. Bottom: cropped views at $8$, $10$, and $12\,\mathrm{s}$. Colors and numeric labels denote track identities. We display the five main classes with confidence at least $0.4$.}
\label{fig:detection_pai_rgb_tracks}
\end{figure*}

\paragraph{Qualitative results.}
\cref{fig:detection_pai_rgb_tracks} shows how the offline pipeline assembles detections into 3D tracks in a PAI street scene, including vehicles and pedestrians at different distances.

\subsubsection{Ablations}
\label{sec:detection_ablation}
\label{sec:detection_ablations}

\paragraph{Setup.}
We train the component variants for 80,000 updates on the internal driving datasets PlanNet and MLMCF using four fixed front cameras and detection-only supervision. Training windows contain 8 temporal samples with randomized spacing up to 8 source frames. We evaluate the final checkpoints on the 300-clip PlanNet validation split, reporting mAP and center-distance HOTA averaged over automobile, person, rider, bus, and heavy truck. Ground truth is restricted to objects visible in the four cameras. \cref{tab:detection_components} also includes a model trained with a scaled schedule and multiple tasks.

\paragraph{Impact of model and supervision choices.}
\cref{tab:detection_components} compares dense-attention PETR detectors with VoVNet, DINOv2, and VGGT features. Starting from VGGT, we increase image resolution, add FPN features with deformable cross-attention, and introduce query denoising with auxiliary 2D supervision.
Higher image resolution improves both detection and tracking while keeping the VGGT aggregator grid unchanged. FPN features and deformable attention further improve mAP, with a decrease in HOTA. Query denoising and auxiliary 2D supervision improve both metrics, reaching $34.7\%$ mAP and $32.5\%$ HOTA. The scaled multi-task reference reaches $47.0\%$ mAP and $43.6\%$ HOTA under the different evaluation settings specified in the caption.

\begin{table}[!tp]
\centering
\small
\setlength{\tabcolsep}{4pt}
\caption{\textbf{Ablation study for 3D object detection and tracking.} All models are trained on four cameras, and all except the last configuration use a reduced $80k$ steps training schedule. We indicate in parenthesis next to each backbone its image resolution.}
\label{tab:detection_components}
\begin{tabular*}{\linewidth}{@{\extracolsep{\fill}}lrr@{}}
\toprule
Configuration & mAP $\uparrow$ & HOTA $\uparrow$ \\
\midrule
VoVNet ($320\!\times\!800$) & 21.4 & 21.5 \\
DINOv2 ($532\!\times\!952$) & 30.1 & 28.8 \\
VGGT ($518\!\times\!518$) & 27.5 & 26.6 \\
\quad + resolution $1036\!\times\!1036$ & 31.7 & 30.2 \\
\quad + deformable attention & 32.1 & 29.3 \\
\quad + QueryDN / auxiliary YOLOX & 34.7 & 32.5 \\
\midrule
Scaled schedule and multi-task training & \textbf{47.0} & \textbf{43.6} \\
\bottomrule
\end{tabular*}
\end{table}

\paragraph{Temporal context at inference.}
We vary the number of timestamps processed jointly at inference time using a fixed checkpoint of the scaled multi-task model. Intuitively, increasing the window size enables enhancing per-view image features with further spatio-temporal context via the multi-frame aggregator transformer.
Increasing the window from 2 to 32 timestamps improves mAP by $10.07$ points and HOTA by $6.89$ points without retraining. Extending to 48 timestamps adds $0.59$ mAP points but reduces HOTA by $0.32$ points. Detection therefore continues to improve over the tested range, while tracking reaches its highest score at 32 timestamps.

\begin{table}[!tp]
\centering
\small
\setlength{\tabcolsep}{4pt}
\caption{\textbf{Temporal-context ablation with a fixed checkpoint.} Metrics are five-class averages in percentages; HOTA uses center-distance matching with four-camera visibility-filtered ground truth. Temporal span is $(T-1)\,2/30$ seconds for $T$ timestamps at stride two in 30\,Hz video.}
\label{tab:detection_temporal_context}
\begin{tabular*}{\linewidth}{@{\extracolsep{\fill}}lrrr@{}}
\toprule
Timestamps & Span (s) & mAP $\uparrow$ & HOTA $\uparrow$ \\
\midrule
2 & 0.07 & 37.0 & 36.7 \\
4 & 0.20 & 39.5 & 39.1 \\
8 & 0.47 & 42.1 & 40.5 \\
16 & 1.00 & 44.9 & 42.0 \\
32 & 2.07 & 47.0 & \textbf{43.6} \\
48 & 3.13 & \textbf{47.6} & 43.3 \\
\bottomrule
\end{tabular*}
\end{table}

\paragraph{Tracking motion estimation and smoothing.}
We hold the detector outputs, association schedule, confidence filtering, class mapping, and ground truth fixed, using the same scaled multi-task checkpoint with 32-timestamp inference windows. The baseline uses detector velocities for motion prediction, Hungarian association, gap interpolation, and Savitzky--Golay smoothing. All three configurations use the same two association passes, with maximum gaps of 24 and 50 native 30\,Hz frame indices. We first introduce Kalman prediction and updates, then replace planar trajectory smoothing with a Rauch--Tung--Striebel (RTS) backward pass. RTS is applied to all classes after the final association pass and to person tracks after both passes.
As shown in \cref{tab:detection_tracking_ablation}, Kalman filtering improves HOTA from $41.4$ to $42.1$, a gain of $0.7$ points, and RTS smoothing adds $1.5$ points, reaching $43.6$ for a total gain of $2.2$ points over the velocity baseline.
\begin{table}[!tp]
\centering
\small
\setlength{\tabcolsep}{4pt}

\begin{minipage}[t]{0.42\textwidth}
\vspace{0pt}
\centering
\caption{\textbf{Tracking inference ablation.}}
\label{tab:detection_tracking_ablation}
\vspace{4pt}
\begin{tabular*}{\linewidth}{@{\extracolsep{\fill}}lr@{}}
\toprule
Configuration & HOTA $\uparrow$ \\
\midrule
Velocity baseline & 41.4 \\
\quad + Kalman filtering & 42.1 \\
\quad + RTS smoothing & \textbf{43.6} \\
\bottomrule
\end{tabular*}
\end{minipage}
\hfill
\begin{minipage}[t]{0.56\textwidth}
\vspace{0pt}
\centering
\caption{\textbf{Impact of predicted ego-motion on tracking.}}
\label{tab:detection_egomotion_ablation}
\begin{tabular*}{\linewidth}{@{\extracolsep{\fill}}lrr@{}}
\toprule
& \multicolumn{2}{c}{Ego-motion} \\
\cmidrule(lr){2-3}
Configuration & Ground truth & Predicted \\
\midrule
GT boxes and identities (oracle) & 100.0 & 83.8 \\
Predicted boxes and tracker      & 43.6  & 40.5 \\
\bottomrule
\end{tabular*}
\end{minipage}

\end{table}

\paragraph{Impact of predicted ego-motion.}

As described in \cref{sec:mot}, detections are transformed into a common reference frame for multi-object tracking using ego-poses. We evaluate the impact of replacing ground-truth ego-poses with predicted poses obtained through sequential window alignment. 
In \cref{tab:detection_egomotion_ablation}, we first isolate pose-induced localization error using ground-truth boxes and identities, without running the tracker. Replacing ground-truth ego-poses with predicted poses reduces HOTA from $100.0$ to $83.8$, indicating residual localization error even with perfect detections and identities. With predicted detections and a fixed tracking configuration, the impact is more modest: HOTA decreases from $43.6$ to $40.5$, a drop of $3.1$ points ($7.1\%$ relative). Thus, while pose accuracy affects localization, the full tracking pipeline retains most of its performance when using predicted ego-motion.

\subsection{Occupancy}

Our occupancy branch predicts occupied space and assigns semantic labels on a high-resolution voxel grid from multi-camera video. We evaluate this capability along two main directions: (i) binary occupancy and semantic prediction accuracy on public and PAI benchmarks; and (ii) the contribution of each design choice through component ablations.

\subsubsection{Public benchmark results}

\paragraph{Setup.}
We fine-tune \method on OpenOccupancy~\cite{wang2023openoccupancy}, a surround-view semantic occupancy benchmark that extends nuScenes~\cite{caesar2020nuscenes} with dense voxel annotations. The benchmark evaluates 16 semantic classes at a voxel resolution of $0.2\,\mathrm{m}$ over a region of interest of $[-51.2,51.2]\times[-51.2,51.2]\times[-5,3]\,\mathrm{m}$. We train for $30{,}000$ updates with an initial learning rate of $3\times10^{-4}$ and cosine decay, a batch size of $8$, temporal stride $1$, $1036\times1036$ input images, and random color augmentation. Compared with pre-training, this benchmark changes the semantic classes and spatial extent while retaining the same voxel resolution.

\paragraph{Baselines.}

We compare \method with camera-only baselines spanning plane-based, dense volumetric, and sparse 3D representations. TPVFormer~\cite{huang2023tpvformer} aggregates multi-camera features into three orthogonal feature planes. Dense volumetric approaches include MonoScene~\cite{cao2022monoscene}, OccGen~\cite{wang2024occgen}, and OccMamba~\cite{li2025occmamba}, while CONet~\cite{wang2023openoccupancy} uses coarse-to-fine volumetric refinement. Sparse approaches include SparseOcc~\cite{tang2024sparseocc} and SPSC~\cite{guo2026spsc}. We also compare against SPSC-temporal, which incorporates historical frames and achieves the strongest baseline results in our comparison. For CONet, OccGen, OccMamba, and both SPSC variants, we use the camera-only OpenOccupancy scores reported in~\cite{guo2026spsc}.

\paragraph{Results.}
\method achieves the highest overall IoU and mIoU among the methods in \cref{tab:occupancy_openoccupancy}: $27.4$ IoU and $18.1$ mIoU, compared with $25.5$ and $16.2$ for SPSC-temporal~\cite{guo2026spsc}. It also improves on SPSC-temporal in most semantic classes. The largest remaining gaps to this baseline concern \textit{barrier} and \textit{traffic cone}, which may reflect their absence from the pre-training label set. These results highlight the value of large-scale joint pre-training across multiple tasks: after adaptation to OpenOccupancy, our pre-trained occupancy network outperforms the TPV-based, dense volumetric, and sparse 3D baselines included in this comparison, despite the simple design of its BEV occupancy decoder.

\begin{figure}[!tp]
\centering
\captionsetup{font=small,skip=6pt}
\begingroup
\setlength{\tabcolsep}{0pt}
\renewcommand{\arraystretch}{1}

\newcommand{\openoccupancyrow}[8]{%
  \raisebox{-0.5\height}{%
    \includegraphics[
      width=0.30\linewidth,
      trim={0bp #6bp #2bp #5bp},clip
    ]{assets/openoccupancy/#1}%
  } &%
  \raisebox{-0.5\height}{%
    \includegraphics[
      width=0.34\linewidth,
      trim={788bp #8bp #3bp #7bp},clip
    ]{assets/openoccupancy/#1}%
  } &%
  \raisebox{-0.5\height}{%
    \includegraphics[
      width=0.34\linewidth,
      trim={#4bp #8bp 0bp #7bp},clip
    ]{assets/openoccupancy/#1}%
  } \\[0pt]%
}

\begin{tabular}{
  @{}c@{\hspace{0.01\linewidth}}
  c@{\hspace{0.01\linewidth}}c@{}
}
\small Input camera frames
& \small \method prediction
& \small GT \\[5pt]

\openoccupancyrow{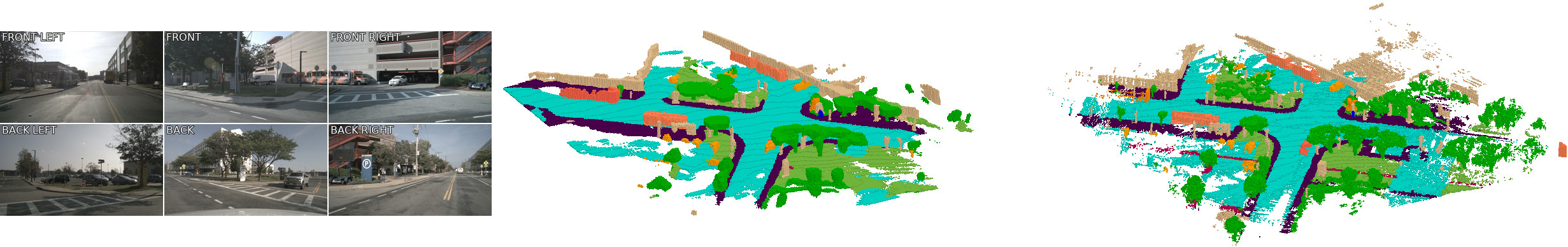}{1690}{841}{1629}{49}{49}{3}{1}
\openoccupancyrow{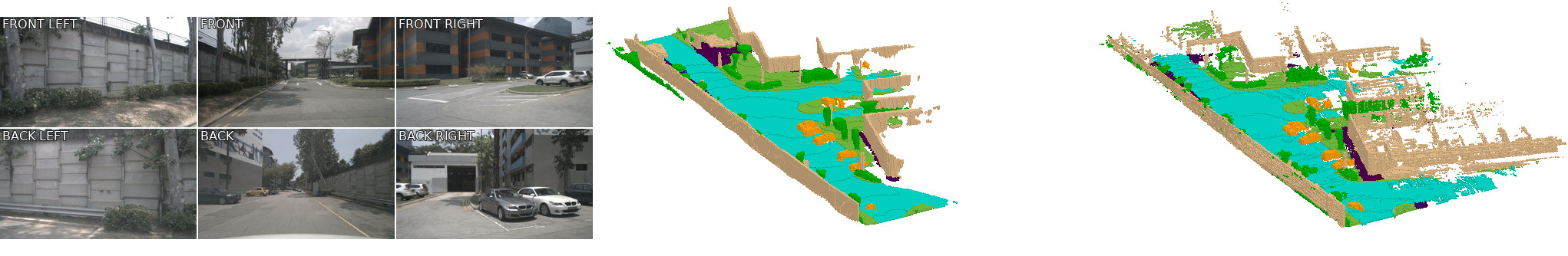}{1270}{631}{1419}{22}{22}{1}{36}
\openoccupancyrow{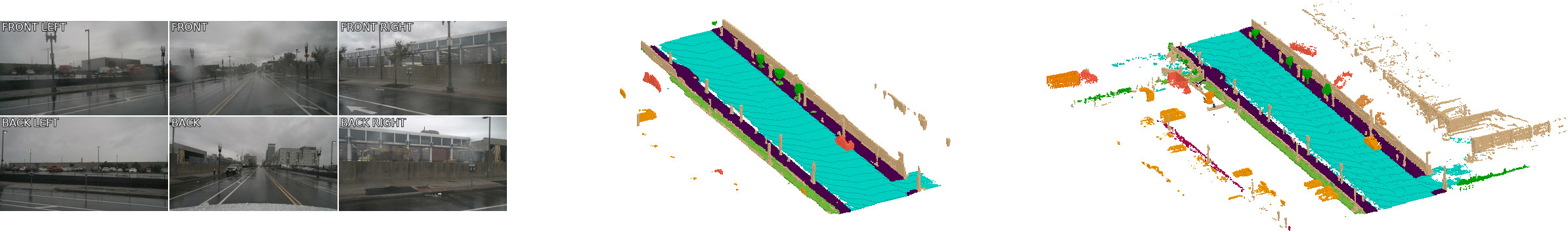}{1616}{804}{1592}{32}{32}{1}{1}
\openoccupancyrow{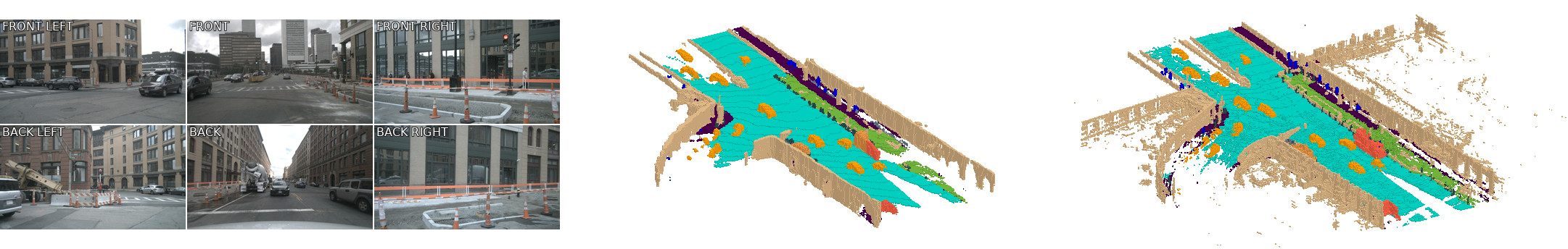}{1390}{691}{1479}{27}{27}{19}{1}
\openoccupancyrow{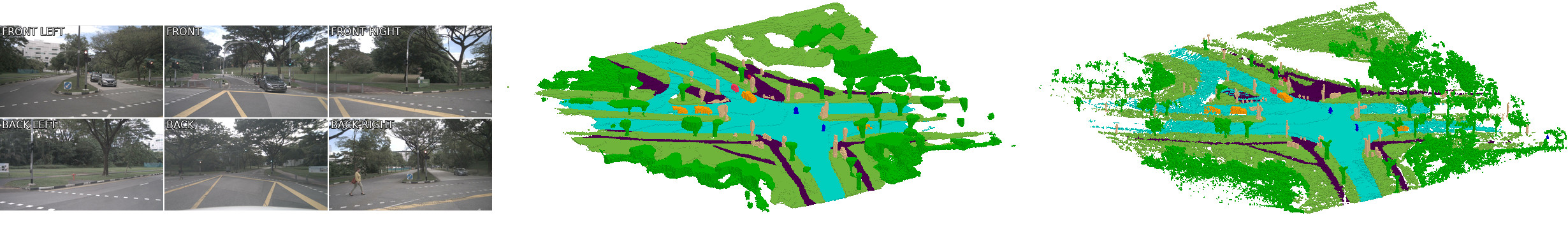}{1688}{840}{1628}{40}{40}{1}{34}
\openoccupancyrow{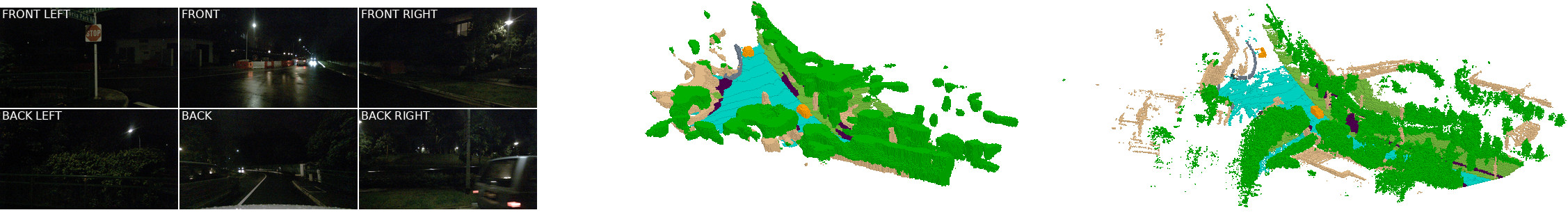}{1482}{737}{1525}{11}{11}{1}{13}
\openoccupancyrow{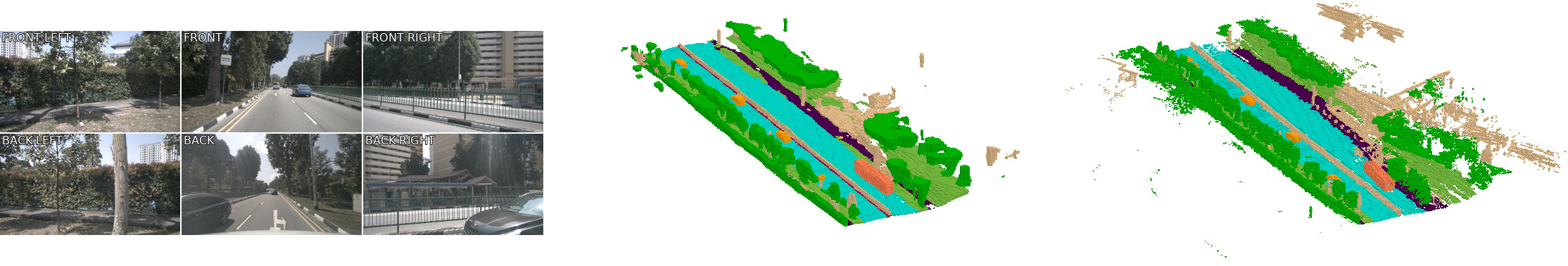}{1456}{724}{1512}{44}{44}{1}{11}

\end{tabular}
\par\vspace{4pt}

\newcommand{\openoccupancylegenditem}[2]{%
  \textcolor[RGB]{#1}{%
    \rule[-0.1ex]{0.7em}{0.7em}%
  }\hspace{0.25em}#2%
}

\resizebox{0.98\linewidth}{!}{%
  \rmfamily\fontsize{7}{8}\selectfont
  \openoccupancylegenditem{112,128,144}{barrier}\hspace{0.4em}%
  \openoccupancylegenditem{220,20,60}{bicycle}\hspace{0.4em}%
  \openoccupancylegenditem{255,127,80}{bus}\hspace{0.4em}%
  \openoccupancylegenditem{255,158,0}{car}\hspace{0.4em}%
  \openoccupancylegenditem{233,150,70}{constr. veh.}\hspace{0.4em}%
  \openoccupancylegenditem{255,61,99}{moto.}\hspace{0.4em}%
  \openoccupancylegenditem{0,0,230}{ped.}\hspace{0.4em}%
  \openoccupancylegenditem{47,79,79}{traf. cone}\hspace{0.4em}%
  \openoccupancylegenditem{255,140,0}{trailer}\hspace{0.4em}%
  \openoccupancylegenditem{255,99,71}{truck}\hspace{0.4em}%
  \openoccupancylegenditem{0,207,191}{drive. surf.}\hspace{0.4em}%
  \openoccupancylegenditem{175,0,75}{other flat}\hspace{0.4em}%
  \openoccupancylegenditem{75,0,75}{sidewalk}\hspace{0.4em}%
  \openoccupancylegenditem{112,180,60}{terrain}\hspace{0.4em}%
  \openoccupancylegenditem{222,184,135}{manmade}\hspace{0.4em}%
  \openoccupancylegenditem{0,175,0}{veg.}%
}

\endgroup
\caption{\textbf{Semantic occupancy predictions on OpenOccupancy.} Each example pairs six camera views with the predicted occupancy and ground truth.}
\label{fig:occupancy_openoccupancy}
\end{figure}

The qualitative examples in \cref{fig:occupancy_openoccupancy} cover urban scenes, including nighttime conditions. \method reconstructs extended regions of \textit{driveable surface} and \textit{sidewalk}, together with surrounding \textit{vegetation} and \textit{manmade} structures. It also localizes individual \textit{car}, \textit{bicycle}, and \textit{pedestrian} instances. The predictions recover both the overall scene layout and local geometric detail, including thin roadside structures.

\begin{table*}[!tp]
\centering
\small
\caption{\textbf{Semantic occupancy results on OpenOccupancy~\cite{wang2023openoccupancy}.} Scores are percentages.}
\label{tab:occupancy_openoccupancy}
\resizebox{\linewidth}{!}{%
\begin{tabular}{l|cc|*{16}{c}}
\toprule
\multirow{2}{*}{Method} & \multirow{2}{*}{IoU} & \multirow{2}{*}{mIoU}
& \multicolumn{16}{c}{Per-class IoU} \\
\cmidrule(lr){4-19}
& & & barrier & bicycle & bus & car
& const. v. & motor. & ped. & t. cone
& trailer & truck & drive. surf. & other flat & sidewalk
& terrain & manmade & veg. \\
\midrule
MonoScene~\cite{cao2022monoscene}
& 18.4 & 6.9 & 7.1 & 3.9 & 9.3 & 7.2 & 5.6 & 3 & 5.9
& 4.4 & 4.9 & 4.2 & 14.9 & 6.3 & 7.9 & 7.4 & 10 & 7.6 \\
TPVFormer~\cite{huang2023tpvformer}
& 15.3 & 7.8 & 9.3 & 4.1 & 11.3 & 10.1 & 5.2 & 4.3 & 5.9
& 5.3 & 6.8 & 6.5 & 13.6 & 9 & 8.3 & 8 & 9.2 & 8.2 \\
CONet~\cite{wang2023openoccupancy}
& 20.1 & 12.8 & 13.2 & 8.1 & 15.4 & 17.2 & 6.3 & 11.2 & 10
& 8.3 & 4.7 & 12.1 & 31.4 & 18.8 & 18.7 & 16.3 & 4.8 & 8.2 \\
OccMamba~\cite{li2025occmamba}
& 21.7 & 13.2 & 13.2 & 7.4 & 15.3 & 17.6 & 6.1 & 10.3 & 10.3
& 6.5 & 6.2 & 13.3 & 32.9 & 20.5 & 19.8 & 17.4 & 4.8 & 8.7 \\
SparseOcc~\cite{tang2024sparseocc}
& 21.8 & 14.1 & 16.1 & 9.3 & 15.1 & 18.6 & 7.3 & 9.4 & 11.2
& 9.4 & 7.2 & 13 & 31.8 & 21.7 & 20.7 & 18.8 & 6.1 & 10.6 \\
OccGen~\cite{wang2024occgen}
& 23.4 & 14.5 & 15.5 & 9.1 & 15.3 & 19.2 & 7.3 & 11.3 & 11.8
& 8.9 & 5.9 & 13.7 & 34.8 & 22 & 21.8 & 19.5 & 6 & 9.9 \\
SPSC~\cite{guo2026spsc}
& 24.3 & 15.2 & 18.9 & 9.6 & \textbf{19.4} & \textbf{21.3} & 8.5 & 11.0 & \textbf{13.9}
& 9.4 & 6.6 & 14 & 34.8 & 19.2 & 22.4 & 18.8 & 5.3 & 9.6 \\
\midrule
SPSC-temporal~\cite{guo2026spsc} & 25.5 & 16.2 & \textbf{19.9} & 10.1 & 14.3 & 19.1 & \textbf{10.8} & 10.6 & 7.1 & \textbf{11.3} & 8.2 & 14.8 & \textbf{40.5} & 23.4 & 25.9 & 21.6 & 8.0 & 13.3 \\
\method & \textbf{27.4} & \textbf{18.1} & 18.0 & \textbf{11.2} & 17.3 & 21.1 & 10.1 & \textbf{14.3}& 11.3 & 8.1 & \textbf{9.7} & \textbf{16.5} & 39.9 & \textbf{28.5} & \textbf{28.0} & \textbf{24.2} & \textbf{13.4} & \textbf{17.7} \\
\bottomrule
\end{tabular}%
}
\end{table*}

\subsubsection{PAI benchmark}

\paragraph{Setup.}
We fine-tune the temporal aggregator and semantic occupancy modules of \method on $1{,}000$ clips curated from our PlanNet dataset and annotated using the pipeline described in \cref{sec:pai_benchmark}. PlanNet shares the same camera setup as PAI. We train for $45{,}000$ iterations using six cameras, $1036\times1036$ input images, an initial learning rate of $3\times10^{-3}$ with cosine decay, batch size of $8$ and a temporal stride of $2$. We retain the pre-trained classifier weights for shared classes and randomly initialize the output weights for new PAI classes.
The PAI occupancy annotations have a resolution of $0.1\,\mathrm{m}$, which we re-voxelize to $0.2\,\mathrm{m}$ to match our output resolution. We use frequency-based semantic label assignment to preserve class coverage and limit the erosion of thin structures and class boundaries during re-voxelization. During evaluation, each forward pass uses $6$ cameras at eight timestamps sampled with temporal stride $2$, corresponding to $48$ input images. We apply a binary occupancy confidence threshold of $0.5$. \cref{tab:occupancy_pai_benchmark} reports binary IoU, semantic mIoU, and per-class scores over the full ROI, $[-20,100]\times[-30,30]\times[-3,5]\,\mathrm{m}$, and separate longitudinal regions. Following OpenOccupancy~\cite{wang2023openoccupancy}, we evaluate over the full ROI without visibility masking, including both visible and occluded regions.
Due to limited annotations for \textit{motorcycle}, \textit{bicycle}, and \textit{train}, we remap \textit{motorcycle} and \textit{bicycle} to the existing \textit{rider} class. We exclude \textit{train} from semantic evaluation while retaining its voxels as occupied in binary evaluation.

\paragraph{Results.}
\method achieves $31.5$ IoU and $14.1$ mIoU over the full ROI. IoU/mIoU is higher in the nearby regions, at $38.4/18.4$ for $[-20,0]\,\mathrm{m}$ and $35.0/16.0$ for $[0,50]\,\mathrm{m}$, and falls to $22.9/6.9$ in the $[50,100]\,\mathrm{m}$ region. The decline is particularly pronounced for thin structures, such as poles and traffic lights/signs, and small objects, such as people and riders.
The daytime and nighttime examples in \cref{fig:occupancy_pai} show road layouts, vehicles, and roadside structures. Distant geometry is less complete, consistent with the lower scores at longer ranges.

\begin{table*}[!tp]
\centering
\scriptsize
\setlength{\tabcolsep}{3pt}
\caption{\textbf{\method semantic occupancy results on the PAI test set.} Scores are percentages. All rows use $y\in[-30,30]$\,m and $z\in[-3,5]$\,m. In the class headers, build. denotes building, traf.l. denotes traffic light, traf.s. denotes traffic sign, and veg. denotes vegetation.}
\label{tab:occupancy_pai_benchmark}
\resizebox{\textwidth}{!}{%
\begin{tabular}{l|c|c|*{15}{c}}
\toprule
\multirow{2}{*}{ROI $x$ (m)} & \multirow{2}{*}{IoU} & \multirow{2}{*}{mIoU}
& \multicolumn{15}{c}{Per-class IoU} \\
\cmidrule(lr){4-18}
& & & road & sidewalk & build. & wall & fence & pole & traf.l. & traf.s. & veg. & terrain & person & rider & car & truck & bus \\
\midrule
$[-20,0]$ & 38.4 & 18.4 & 60.6 & 34.2 & 8.2 & 13.7 & 8.0 & 10.0 & 7.9 & 5.6 & 13.5 & 30.1 & 9.6 & 14.0 & 26.9 & 18.4 & 15.2 \\
$[0,50]$ & 35.0 & 16.0 & 56.1 & 30.0 & 6.9 & 12.6 & 6.2 & 8.3 & 7.1 & 5.0 & 11.5 & 25.2 & 8.3 & 11.9 & 21.8 & 14.9 & 14.3 \\
$[50,100]$ & 22.9 & 6.9 & 35.8 & 16.1 & 2.4 & 8.3 & 2.7 & 1.1 & 1.0 & 0.4 & 4.2 & 14.4 & 0.8 & 1.1 & 7.0 & 4.1 & 4.3 \\
\midrule
$[-20,100]$ & 31.5 & 14.1 & 49.2 & 25.8 & 5.9 & 11.3 & 5.5 & 6.8 & 5.9 & 3.9 & 9.7 & 22.1 & 7.9 & 11.3 & 19.8 & 13.7 & 12.8 \\
\bottomrule
\end{tabular}
}
\end{table*}

\begin{figure}[!tp]
\centering
\captionsetup{font=small,skip=6pt}
\begingroup
\setlength{\tabcolsep}{0pt}
\renewcommand{\arraystretch}{1}

\newcommand{\pairow}[8]{%
  \raisebox{-0.5\height}{\includegraphics[width=0.30\linewidth,trim={0bp #6bp #2bp #5bp},clip]{assets/pai/#1}} &%
  \raisebox{-0.5\height}{\includegraphics[width=0.34\linewidth,trim={788bp #8bp #3bp #7bp},clip]{assets/pai/#1}} &%
  \raisebox{-0.5\height}{\includegraphics[width=0.34\linewidth,trim={#4bp #8bp 0bp #7bp},clip]{assets/pai/#1}} \\[0pt]%
}

\begin{tabular}{@{}c@{\hspace{0.01\linewidth}}c@{\hspace{0.01\linewidth}}c@{}}
\small Input camera frames & \small \method prediction & \small GT \\[5pt]

\pairow{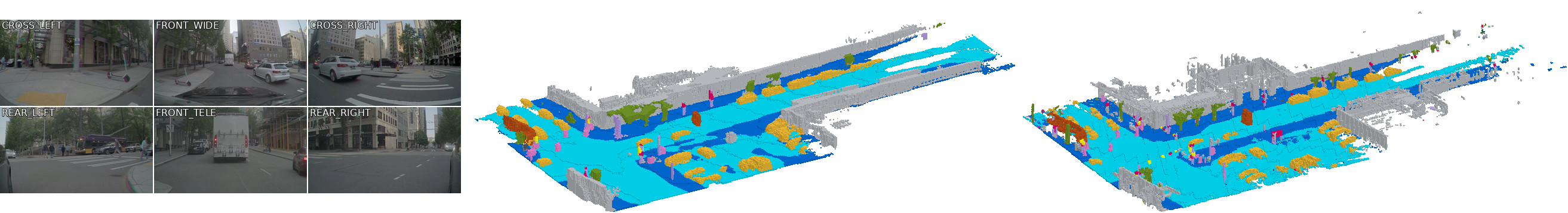}{1854}{923}{1711}{33}{33}{18}{1}
\pairow{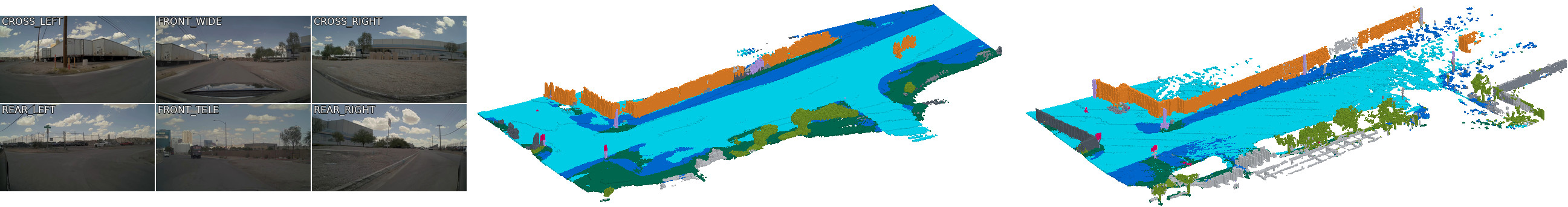}{1822}{907}{1695}{26}{26}{6}{1}
\pairow{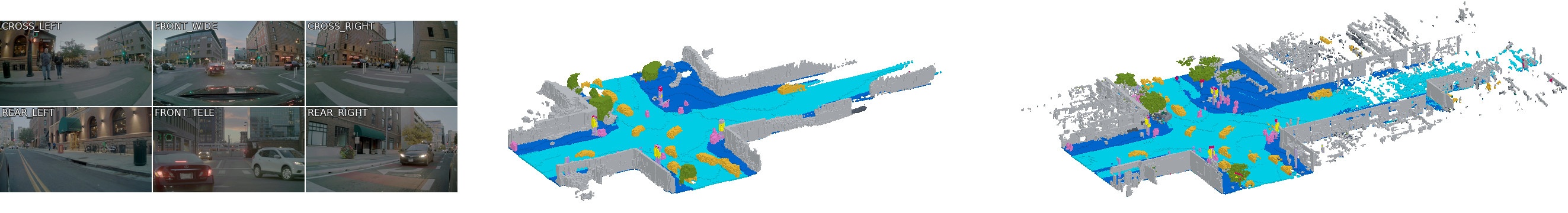}{1874}{933}{1721}{35}{35}{1}{19}
\pairow{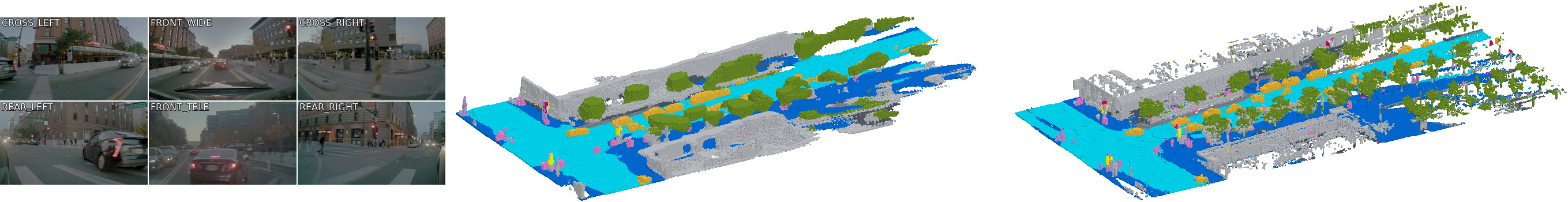}{1946}{969}{1757}{30}{30}{3}{1}
\pairow{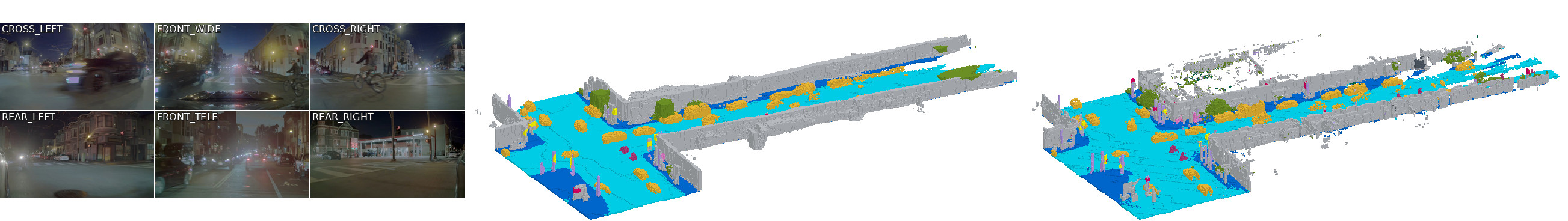}{1834}{913}{1701}{39}{39}{55}{1}
\pairow{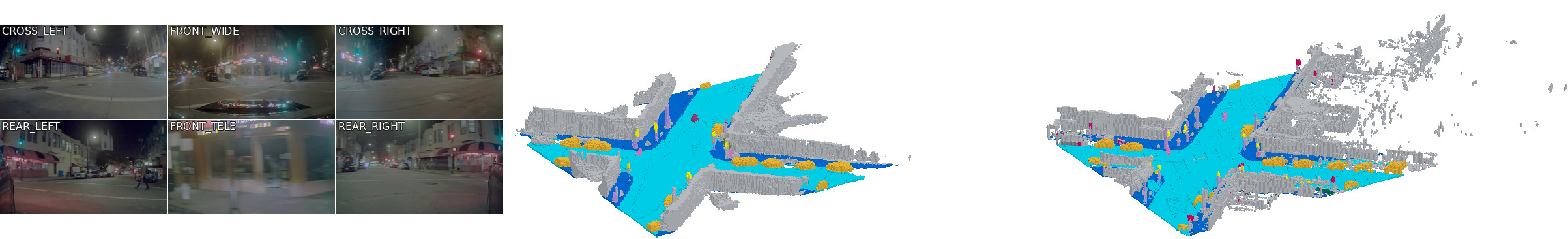}{1634}{813}{1601}{38}{38}{20}{1}
\pairow{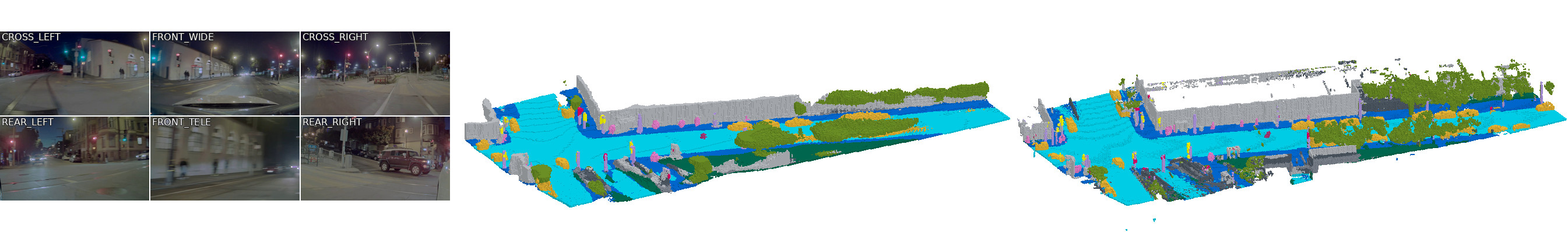}{1918}{955}{1743}{54}{54}{99}{1}

\end{tabular}
\par\vspace{4pt}

\newcommand{\pailegenditem}[2]{%
  \textcolor[RGB]{#1}{\rule[-0.1ex]{0.7em}{0.7em}}\hspace{0.25em}#2%
}

\resizebox{0.98\linewidth}{!}{%
  \rmfamily\fontsize{7}{8}\selectfont
  \pailegenditem{0,204,230}{road}\hspace{0.4em}%
  \pailegenditem{0,102,204}{sidewalk}\hspace{0.4em}%
  \pailegenditem{181,184,189}{building}\hspace{0.4em}%
  \pailegenditem{128,135,145}{wall}\hspace{0.4em}%
  \pailegenditem{80,91,103}{fence}\hspace{0.4em}%
  \pailegenditem{181,154,232}{pole}\hspace{0.4em}%
  \pailegenditem{255,255,0}{traf. light}\hspace{0.4em}%
  \pailegenditem{230,0,102}{traf. sign}\hspace{0.4em}%
  \pailegenditem{107,143,36}{veg.}\hspace{0.4em}%
  \pailegenditem{0,102,77}{terrain}\hspace{0.4em}%
  \pailegenditem{230,128,204}{person}\hspace{0.4em}%
  \pailegenditem{175,57,123}{rider}\hspace{0.4em}%
  \pailegenditem{230,179,51}{car}\hspace{0.4em}%
  \pailegenditem{230,129,37}{truck}\hspace{0.4em}%
  \pailegenditem{185,78,22}{bus}%
}

\endgroup
\caption{\textbf{Semantic occupancy predictions on PAI.} Each example pairs six camera views with the predicted occupancy and ground truth.}
\label{fig:occupancy_pai}
\end{figure}

\subsubsection{Component ablation}

\paragraph{Setup.}
We study the occupancy architecture and training recipe of \method through seven configurations and report the results in \cref{tab:occupancy_components_ablation}. Each variant is trained on NRE, PlanNet, or their combination, and evaluated on the same $50$ validation clips from NRE, covering different driving scenarios. These experiments use $4$ input cameras at $512\times512$ resolution and an ROI of $[0,100]\times[-25,25]\times[-3,5]\,\mathrm{m}$. Training focuses only on the semantic occupancy task, with auxiliary BEV segmentation supervision where indicated. The studied variants vary the backbone, BEV projection, auxiliary supervision, training data, training iterations, and global batch size.

\paragraph{Results.}
Replacing LSS~\cite{philion2020lss} with our BEV deformable attention module improves all reported metrics, increasing full-ROI IoU and mIoU from $24.03$ and $10.37$ to $27.08$ and $11.44$, respectively. Replacing DINOv2 with VGGT further increases performance to $27.64$ IoU and $11.94$ mIoU, with improvements in all regions.
The auxiliary BEV segmentation task increases full-ROI mIoU from $11.94$ to $12.28$, and adding Dice supervision further raises it to $12.51$. 
Adding PlanNet to the training data further improves both geometric and semantic prediction, increasing full-ROI IoU and mIoU to $28.64$ and $13.33$. Extending training and increasing the global batch size through additional GPUs produces the highest scores across all reported metrics and regions: $32.35$ IoU and $15.84$ mIoU over the full ROI, $35.28$ and $18.51$ within $50\,\mathrm{m}$, and $28.61$ and $11.99$ beyond $50\,\mathrm{m}$.

\begin{table}[!tp]
\centering
\small
\caption{\textbf{Architecture and training ablations for semantic occupancy.} All configurations use $4$ input cameras at $512\times512$ resolution and an ROI of $[0,100]\times[-25,25]\times[-3,5]\,\mathrm{m}$.}
\label{tab:occupancy_components_ablation}
\begin{tabular*}{\linewidth}{@{\extracolsep{\fill}}lcccccc@{}}
\toprule
\multirow{2}{*}{Version}
& \multicolumn{2}{c}{0--100 m}
& \multicolumn{2}{c}{0--50 m}
& \multicolumn{2}{c}{50--100 m} \\
\cmidrule(lr){2-3}
\cmidrule(lr){4-5}
\cmidrule(lr){6-7}
& IoU & mIoU & IoU & mIoU & IoU & mIoU \\
\midrule
DINOv2 + LSS + occ. head      & 24.03 & 10.37 & 27.32 & 12.51 & 19.68 & 7.19 \\
\midrule
DINOv2 + BEV attn. + occ. head & 27.08 & 11.44 & 29.77 & 13.57 & 23.44 & 8.45 \\
VGGT + BEV attn. + occ. head   & 27.64 & 11.94 & 30.33 & 14.01 & 24.01 & 8.96 \\
\quad + BEV segmentation & 27.46 & 12.28 & 30.16 & 14.54 & 23.80 & 8.94 \\
\quad + Dice loss           & 27.46 & 12.51 & 30.34 & 14.62 & 23.61 & 9.36 \\
\quad + additional training data       & 28.64 & 13.33 & 31.42 & 15.68 & 24.89 & 9.85 \\
\midrule
\quad + longer training and larger batch  & \textbf{32.35} & \textbf{15.84}
                            & \textbf{35.28} & \textbf{18.51}
                            & \textbf{28.61} & \textbf{11.99} \\
\bottomrule
\end{tabular*}
\end{table}

\subsection{Gaussian reconstruction}
\begin{figure}[!tp]
\centering
\begingroup
\setlength{\tabcolsep}{0pt}
\renewcommand{\arraystretch}{1}
\begin{tabular}{@{}c@{\hspace{.008\linewidth}}c@{\hspace{.008\linewidth}}c@{\hspace{.008\linewidth}}c@{\hspace{.008\linewidth}}c@{\hspace{.008\linewidth}}c@{}}

\includegraphics[width=.16\linewidth]{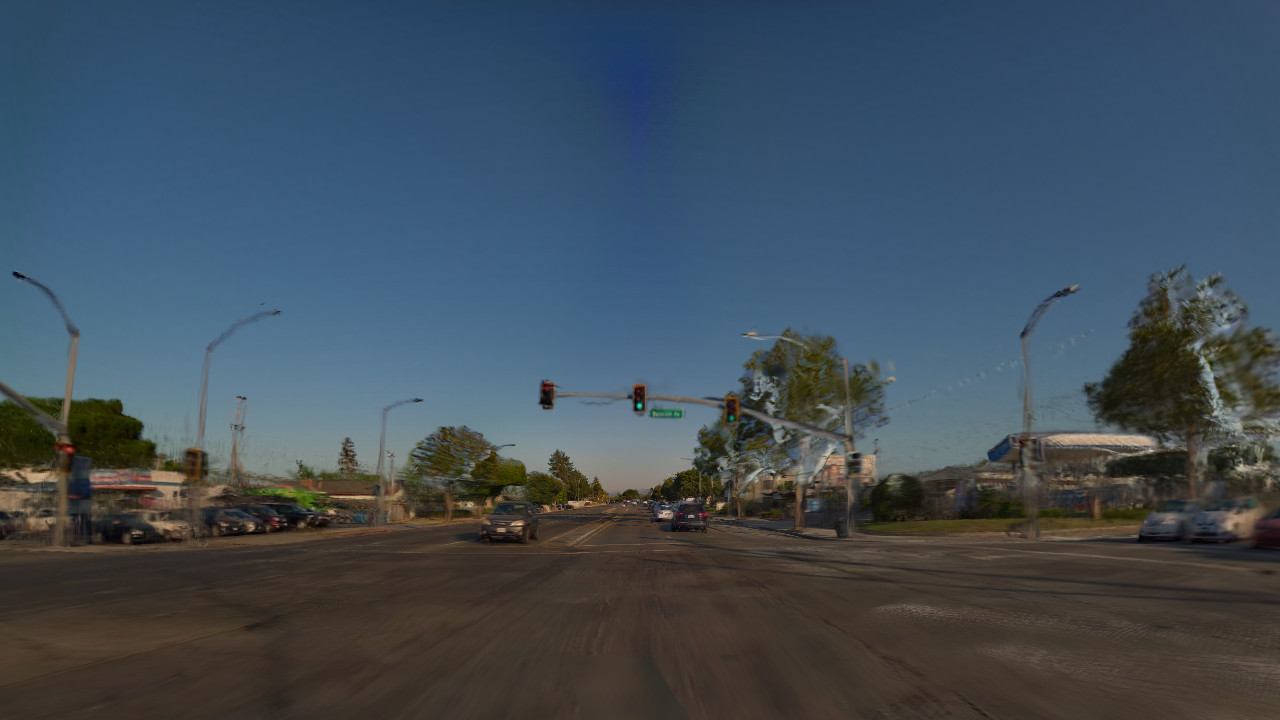} & \includegraphics[width=.16\linewidth]{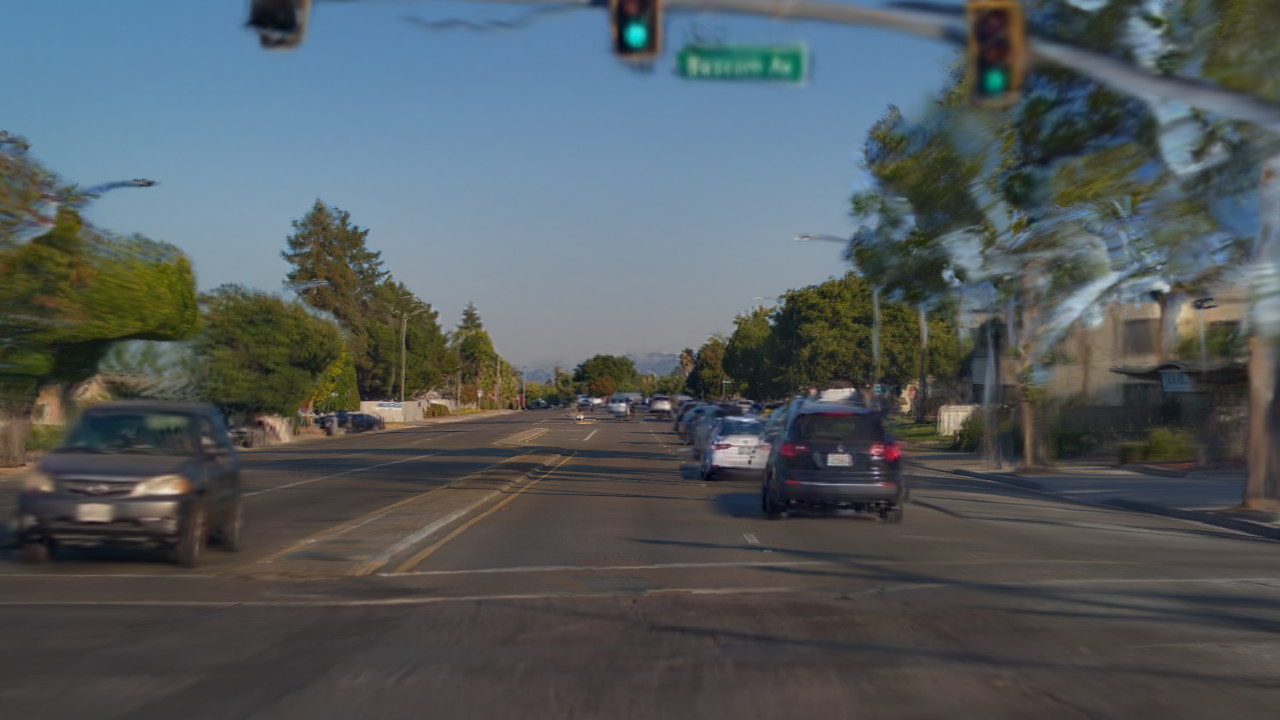} & \includegraphics[width=.16\linewidth]{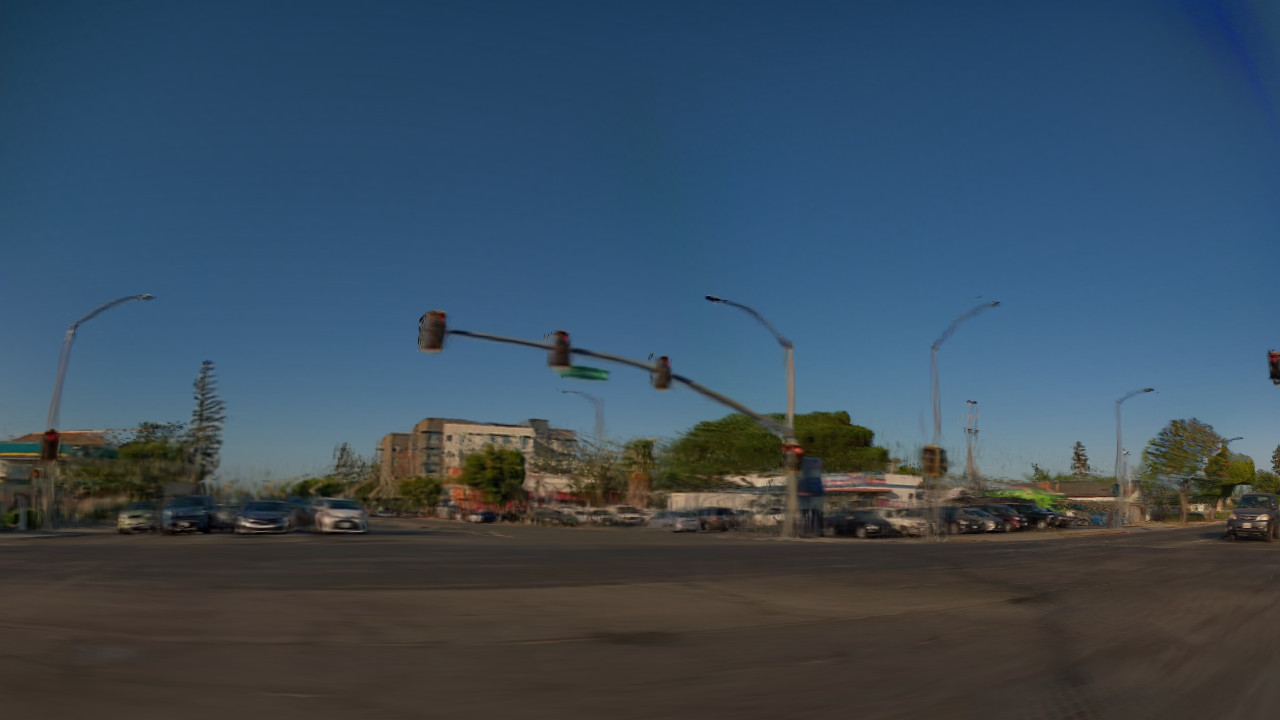} & \includegraphics[width=.16\linewidth]{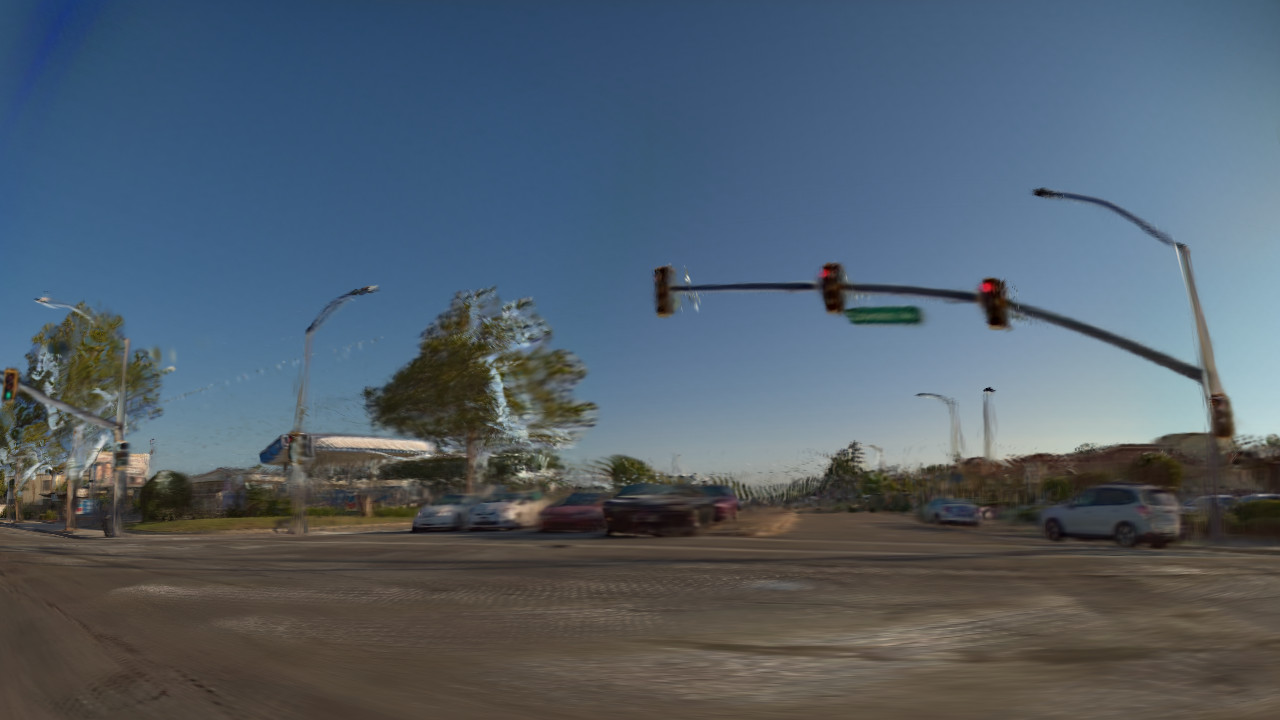} & \includegraphics[width=.16\linewidth]{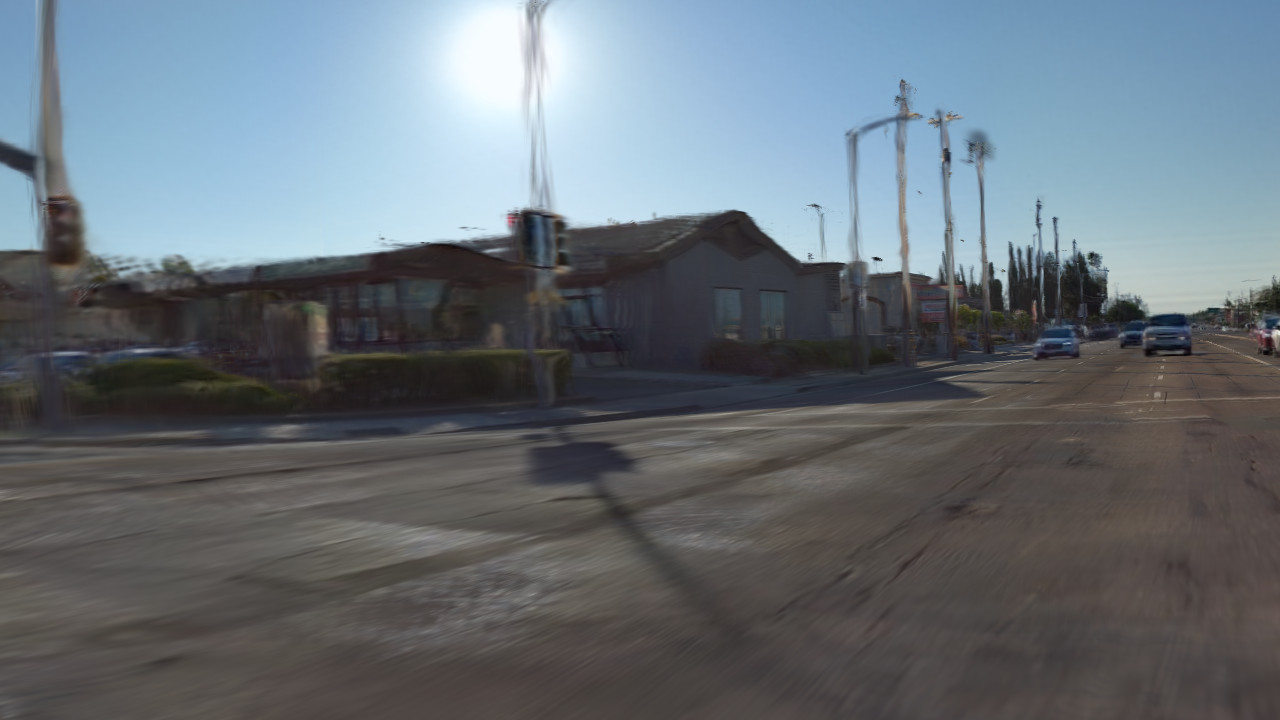} & \includegraphics[width=.16\linewidth]{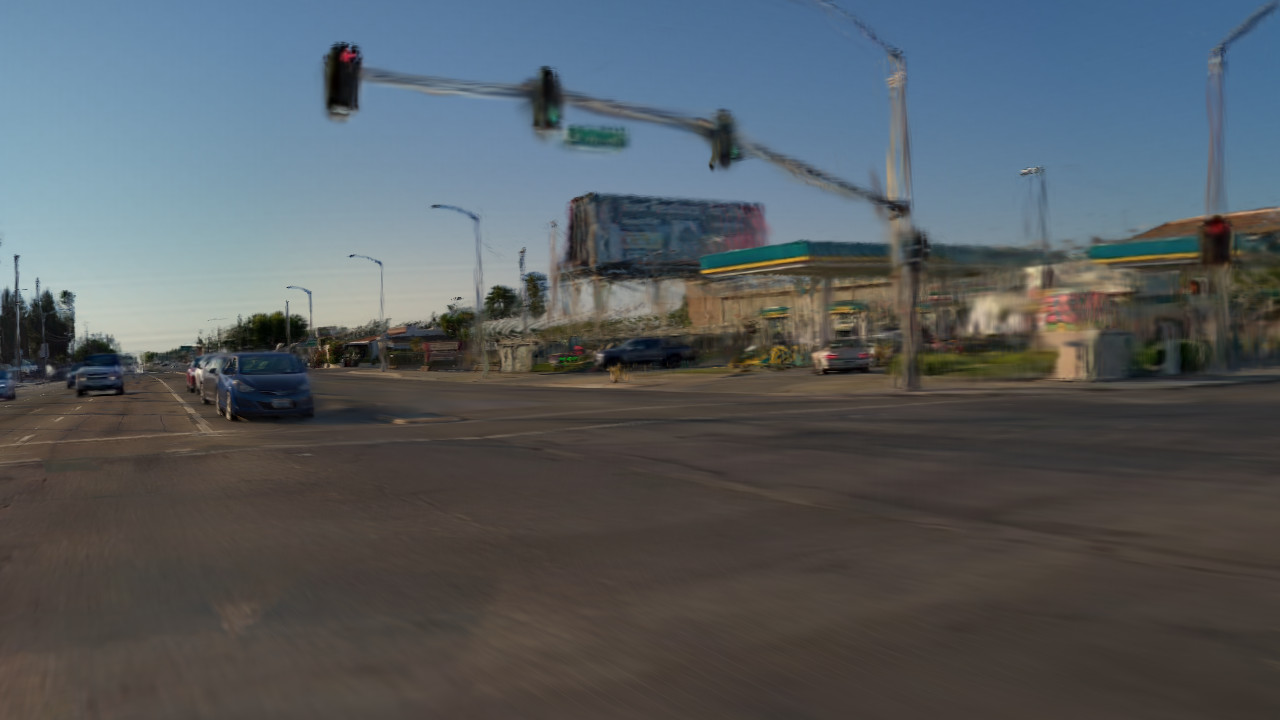}\\[6pt]
\includegraphics[width=.16\linewidth]{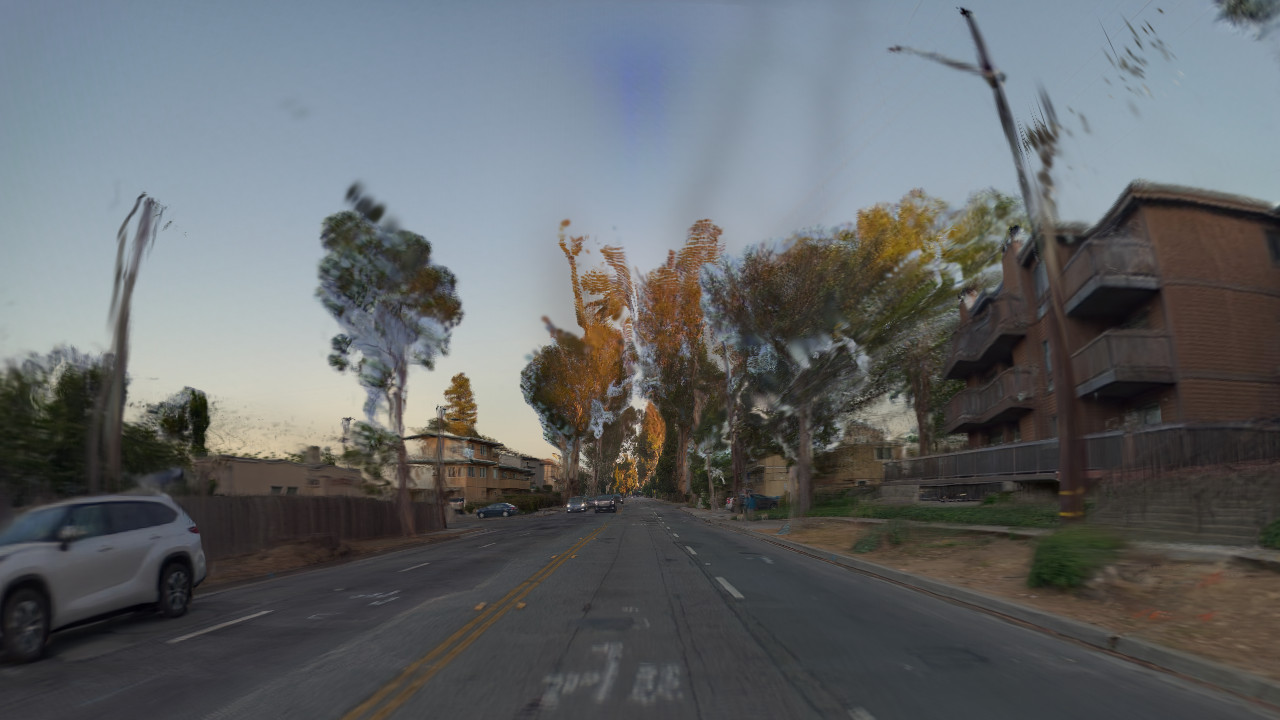} & \includegraphics[width=.16\linewidth]{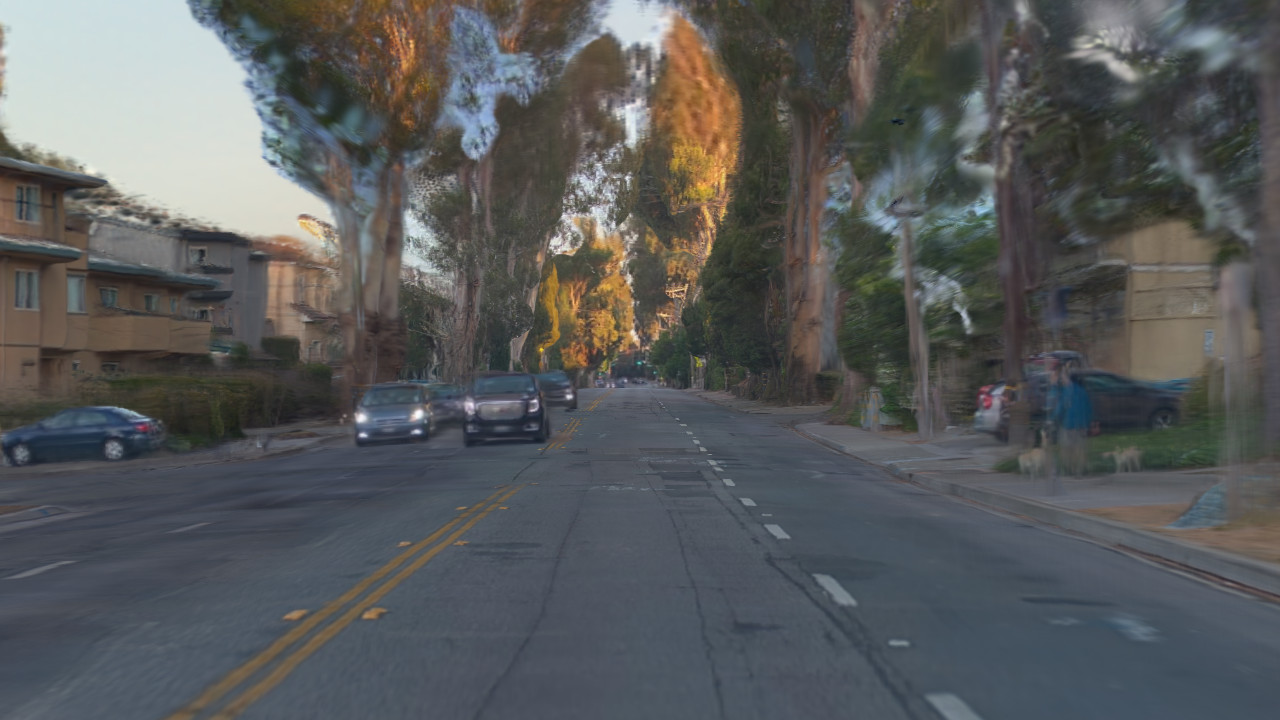} & \includegraphics[width=.16\linewidth]{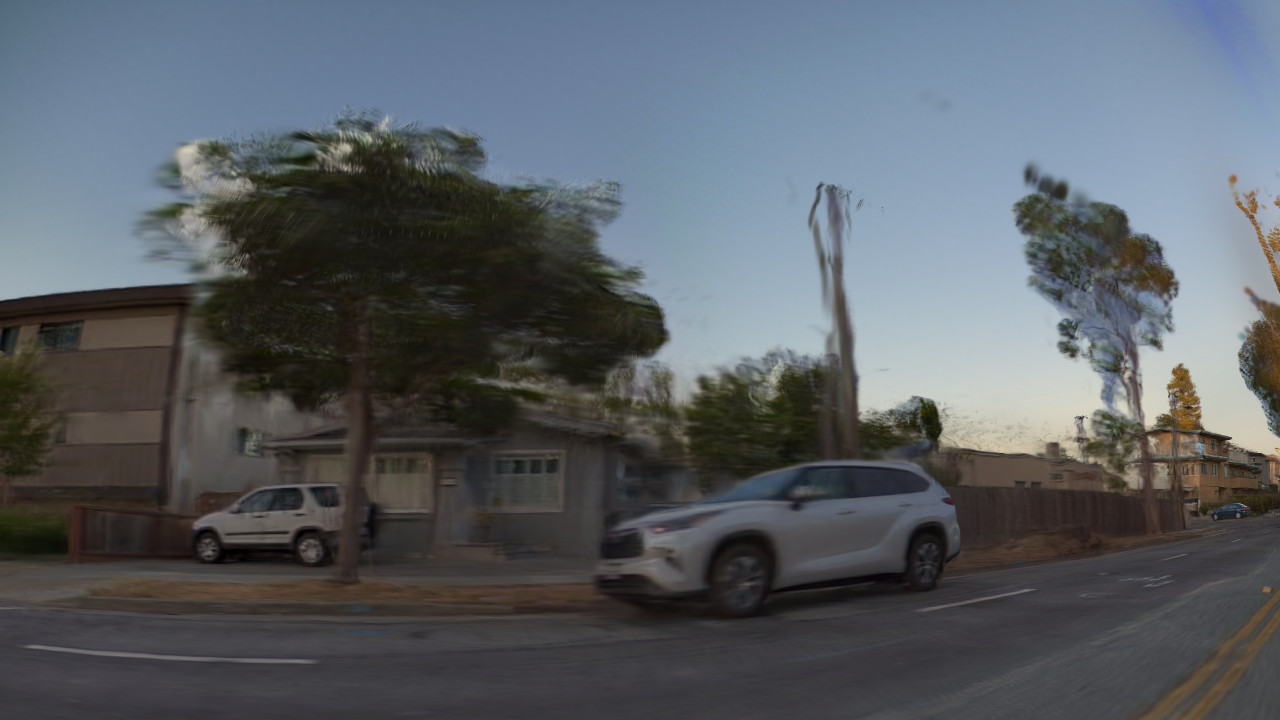} & \includegraphics[width=.16\linewidth]{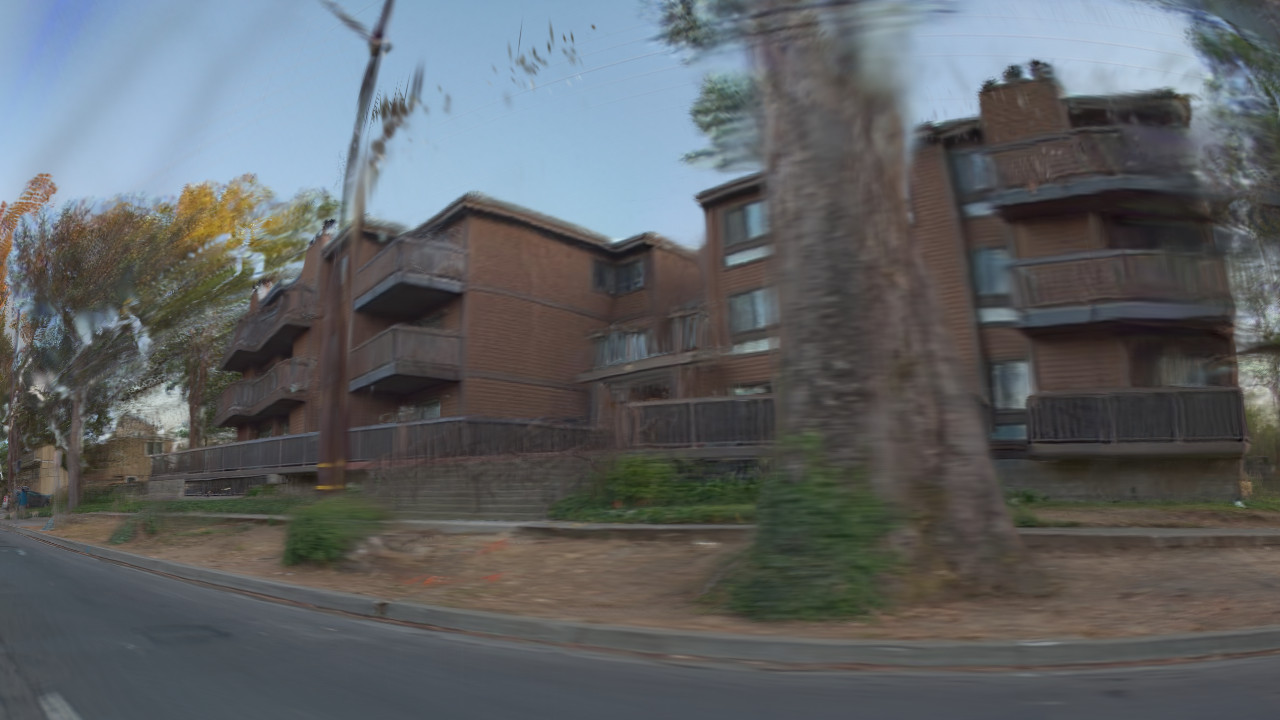} & \includegraphics[width=.16\linewidth]{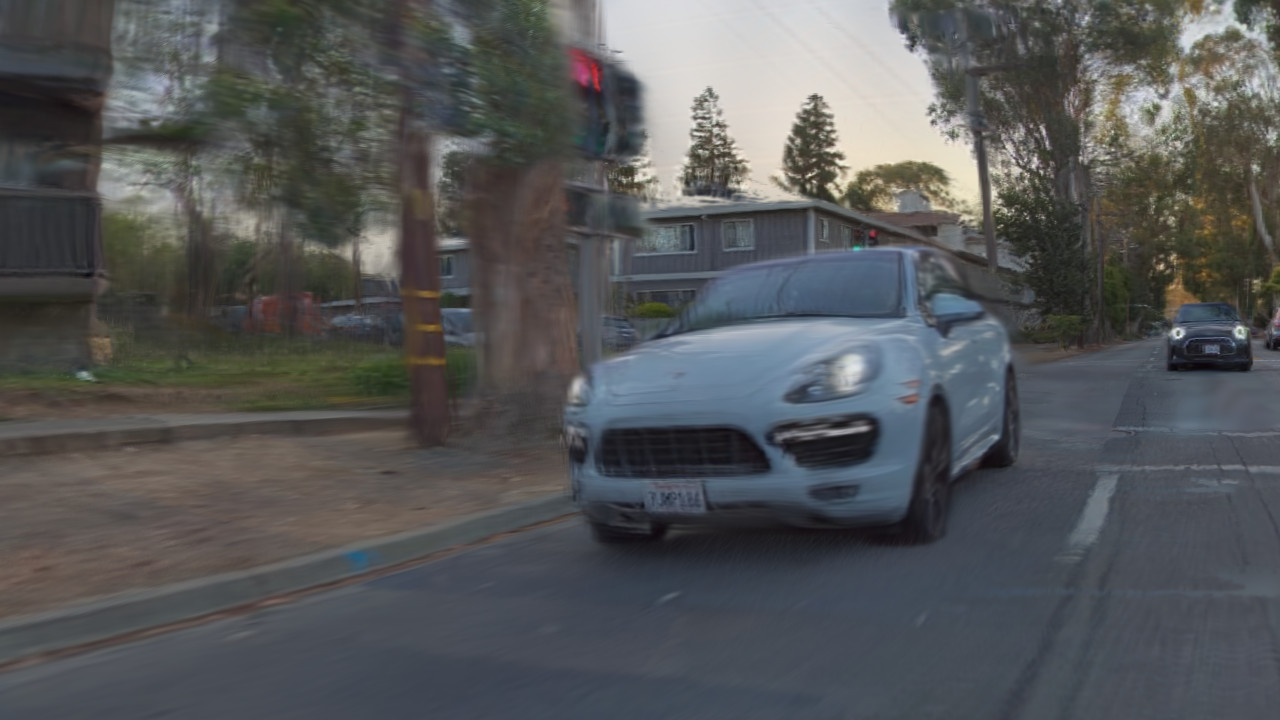} & \includegraphics[width=.16\linewidth]{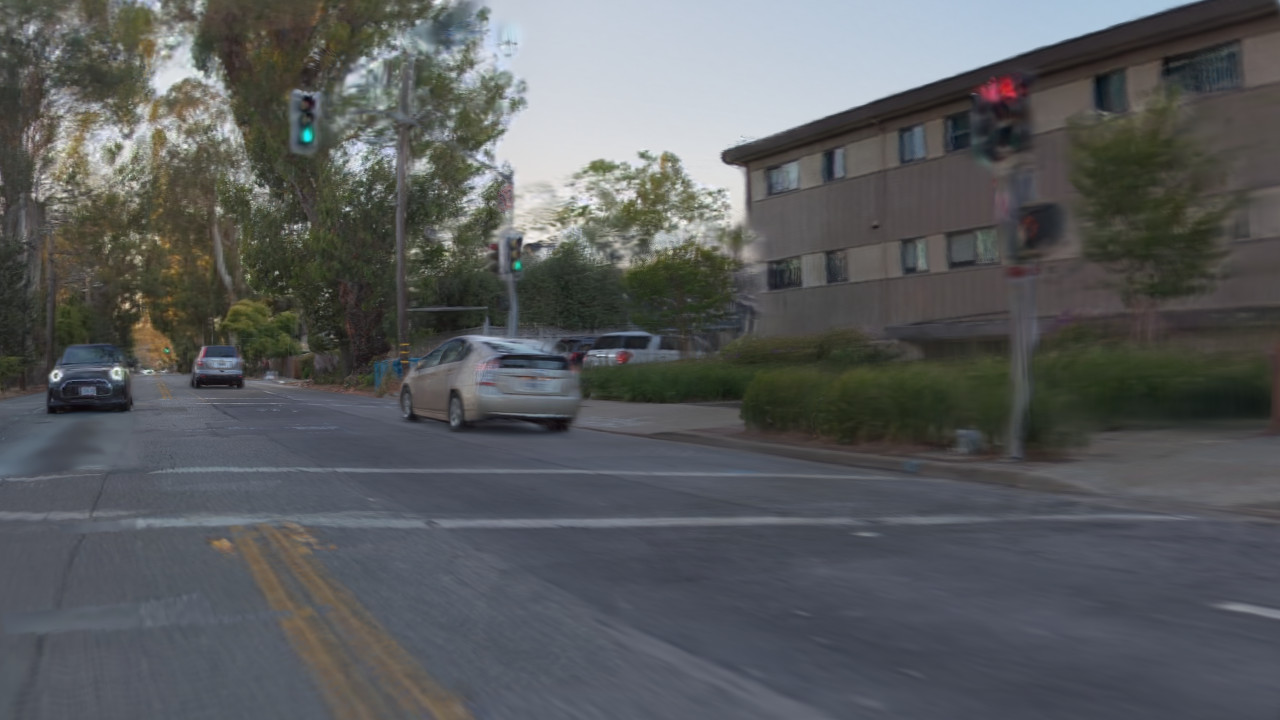}\\[6pt]
\includegraphics[width=.16\linewidth]{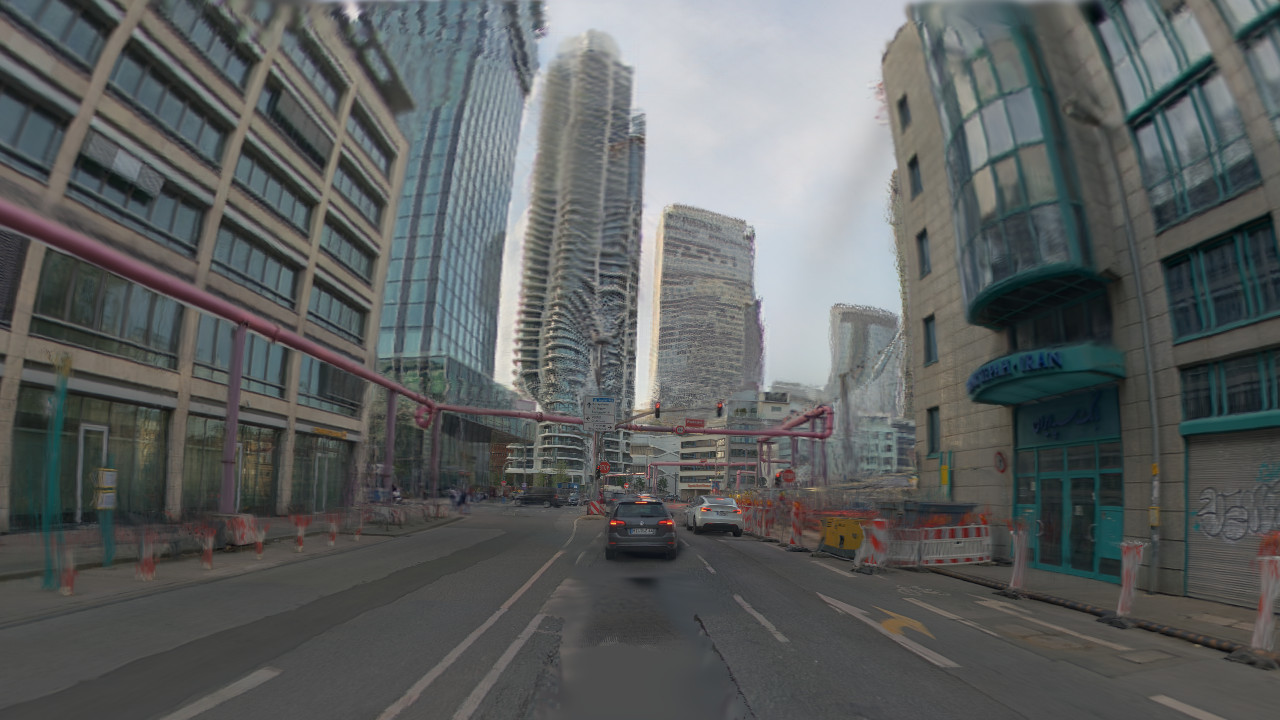} & \includegraphics[width=.16\linewidth]{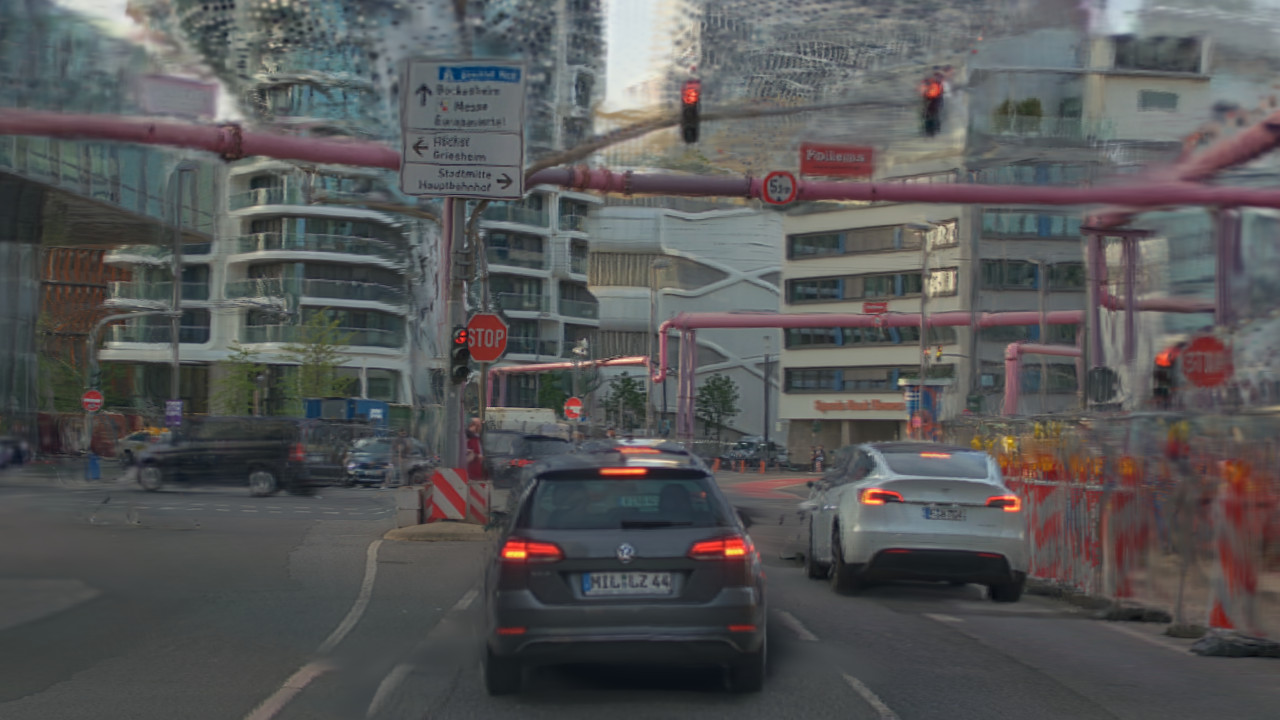} & \includegraphics[width=.16\linewidth]{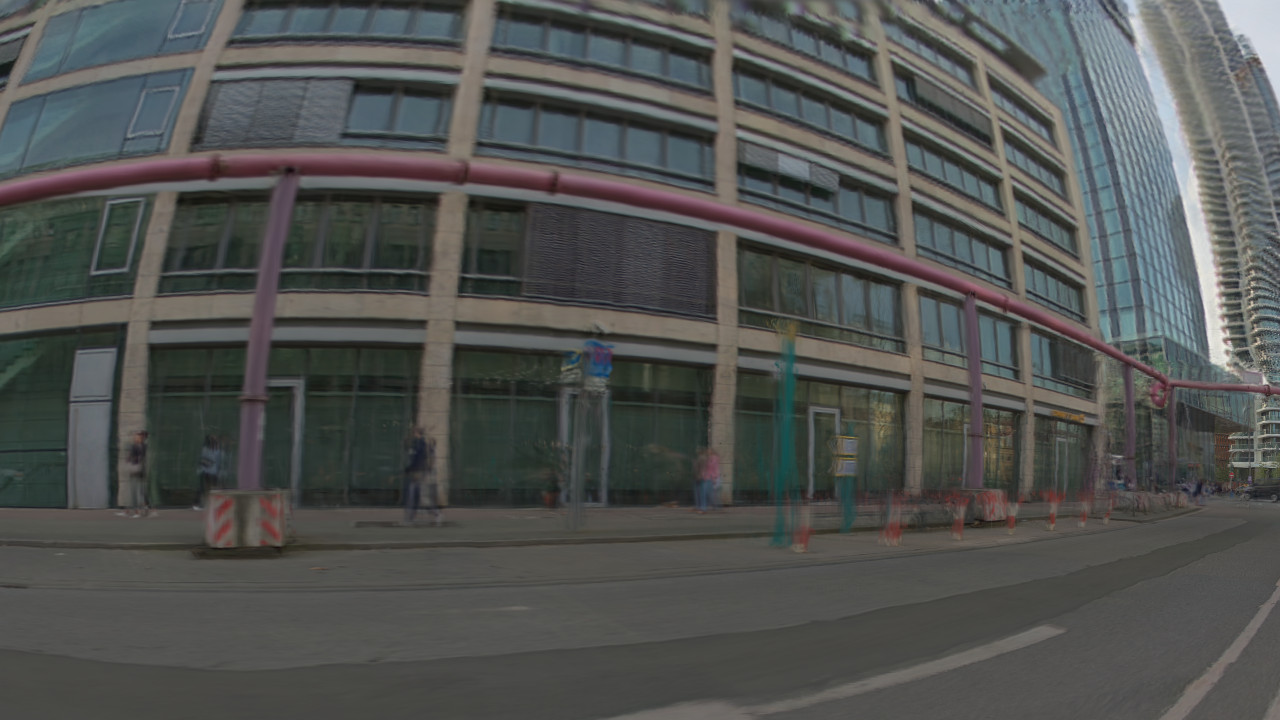} & \includegraphics[width=.16\linewidth]{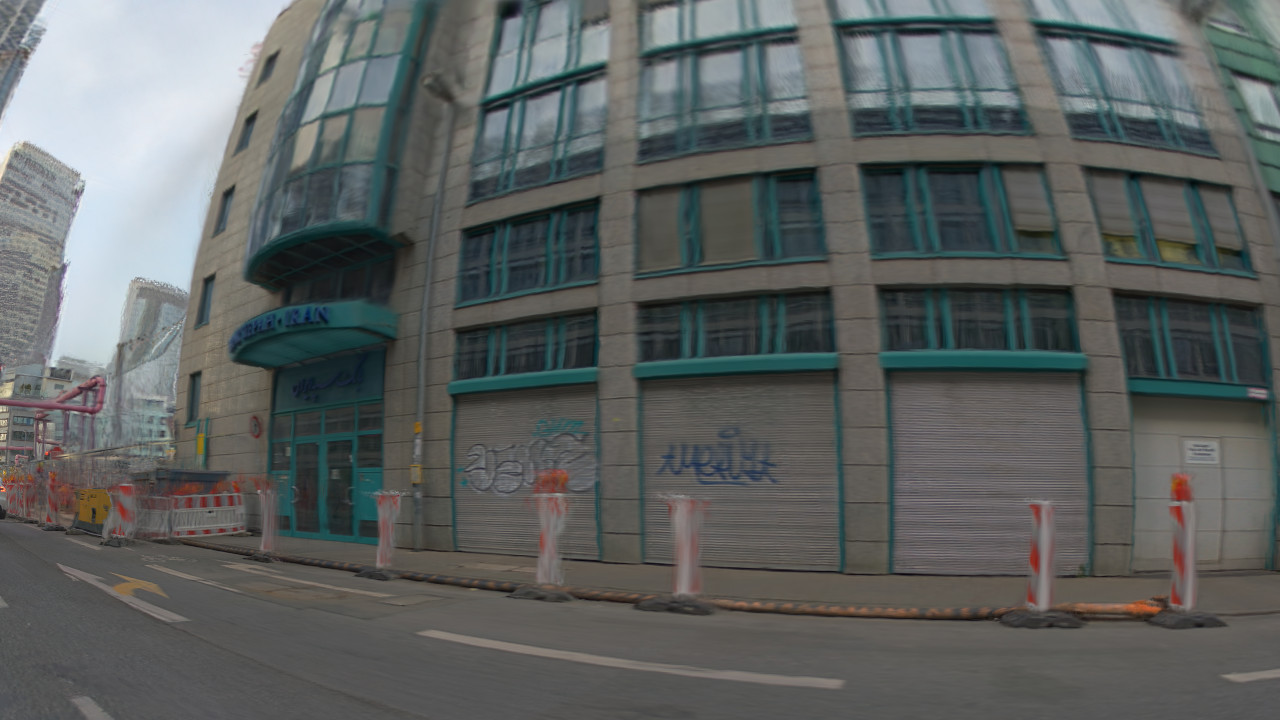} & \includegraphics[width=.16\linewidth]{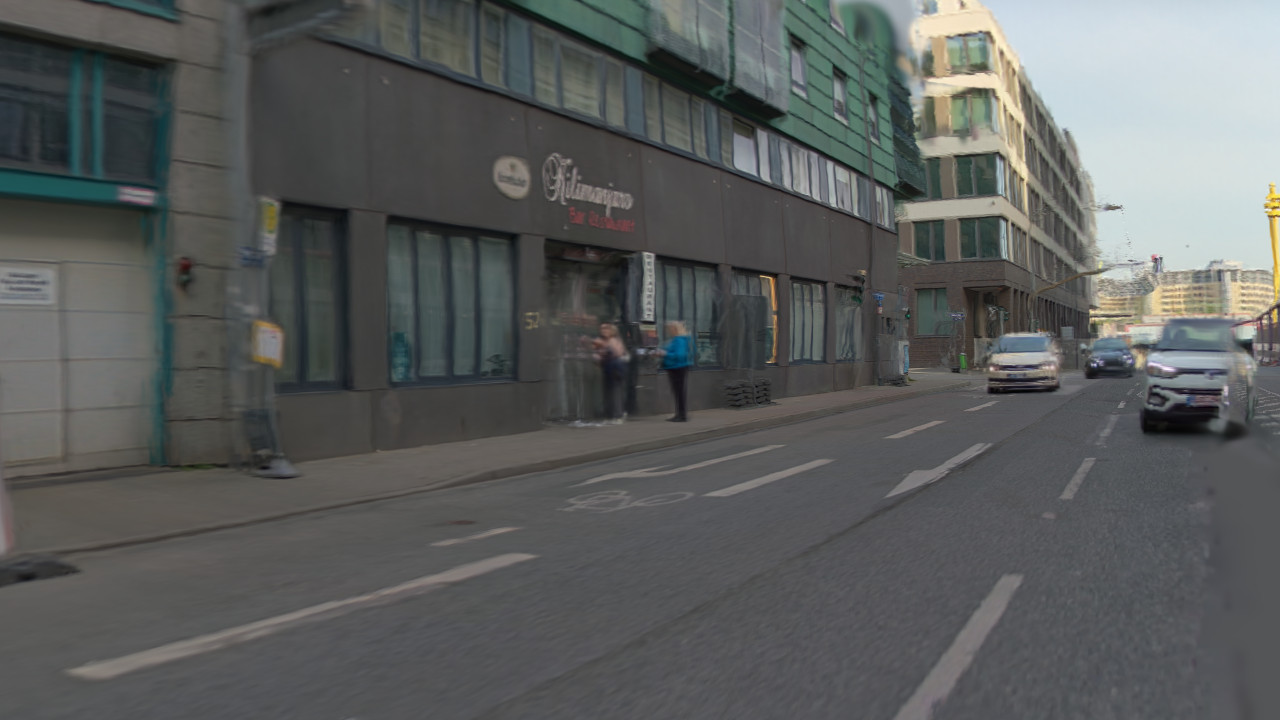} & \includegraphics[width=.16\linewidth]{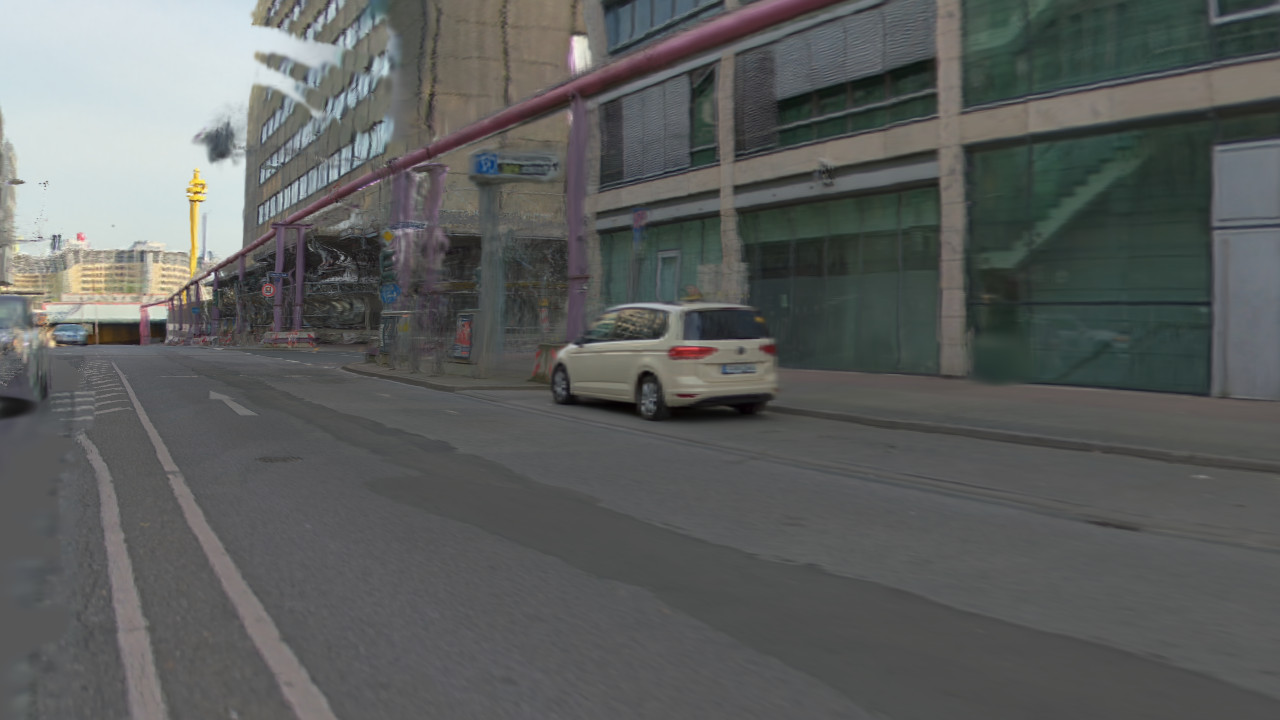}\\[6pt]
\includegraphics[width=.16\linewidth]{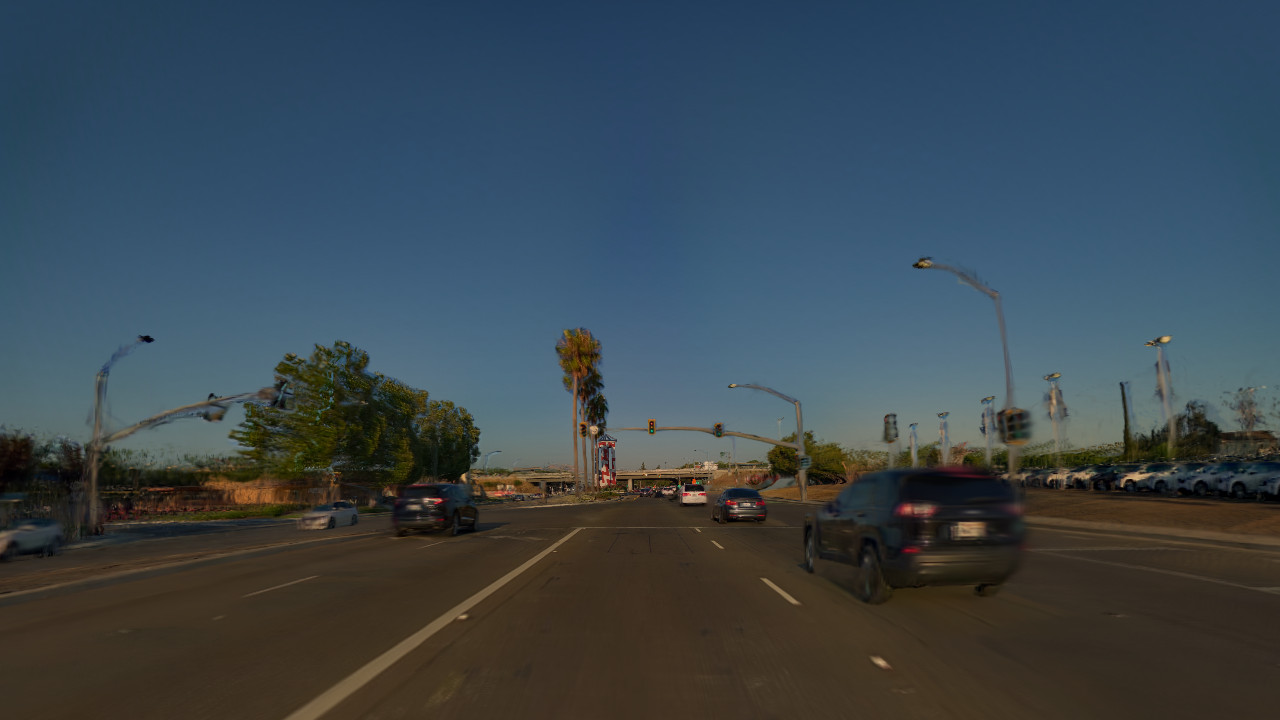} & \includegraphics[width=.16\linewidth]{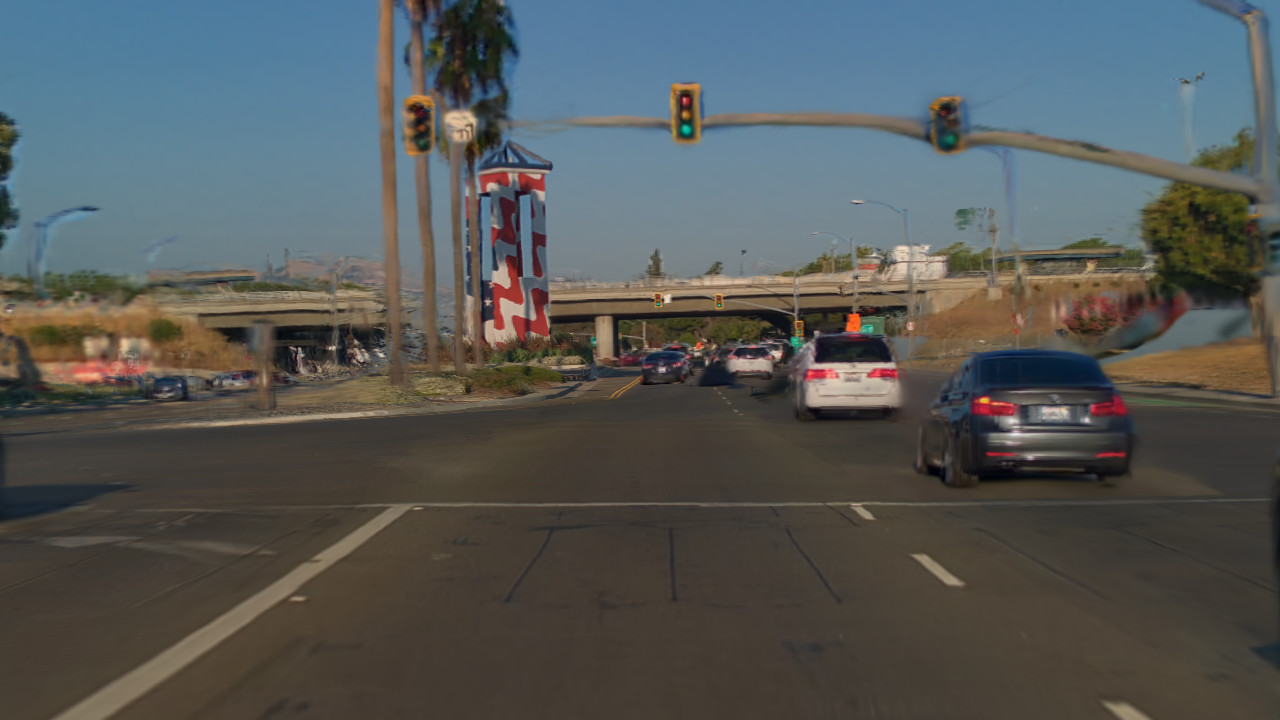} & \includegraphics[width=.16\linewidth]{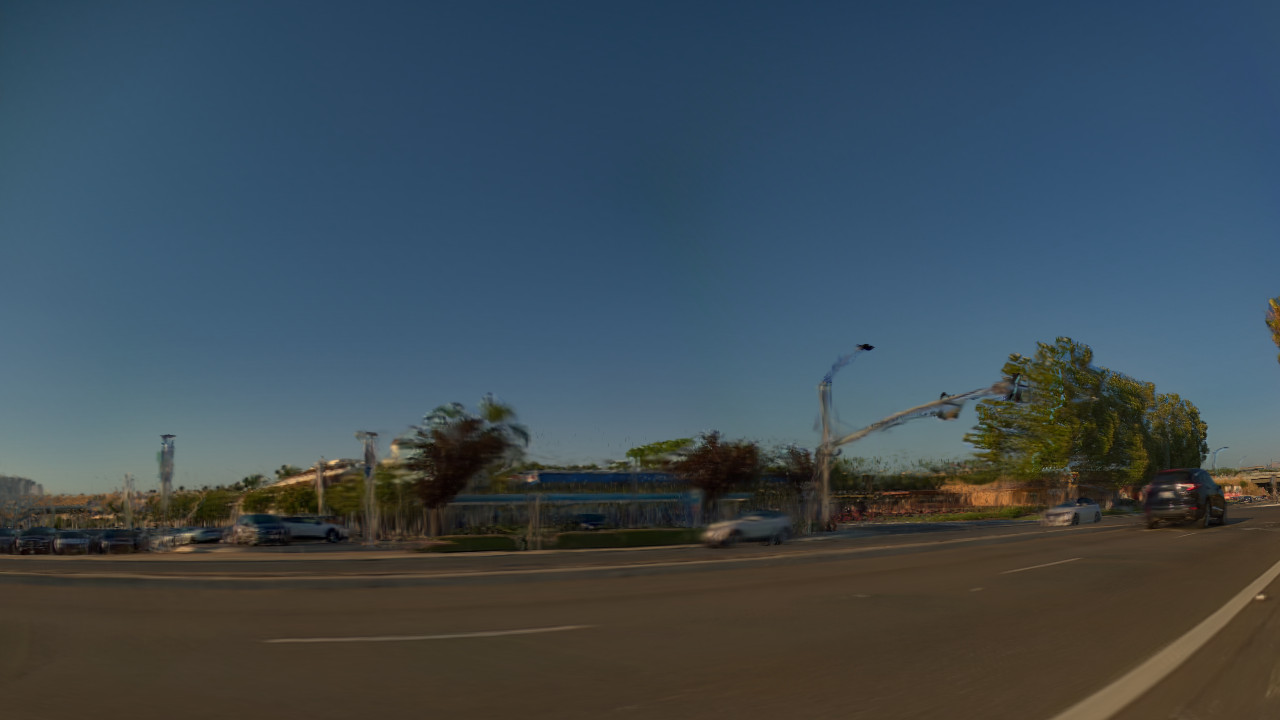} & \includegraphics[width=.16\linewidth]{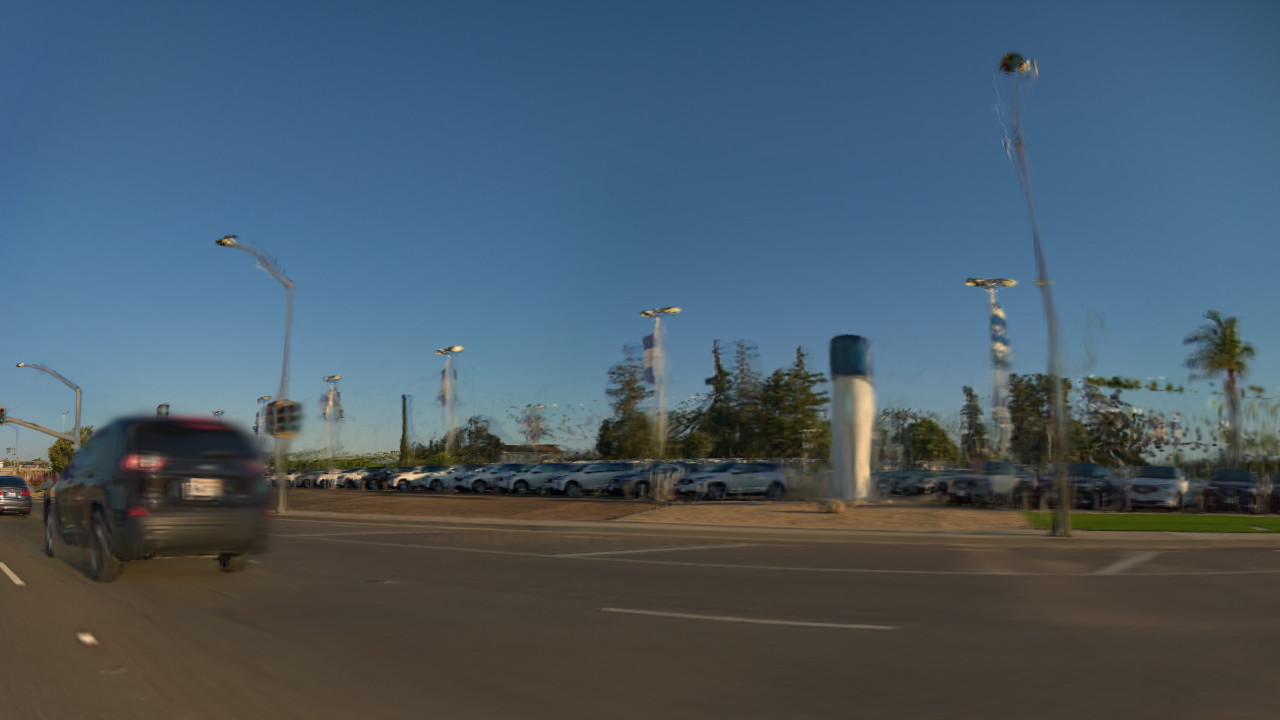} & \includegraphics[width=.16\linewidth]{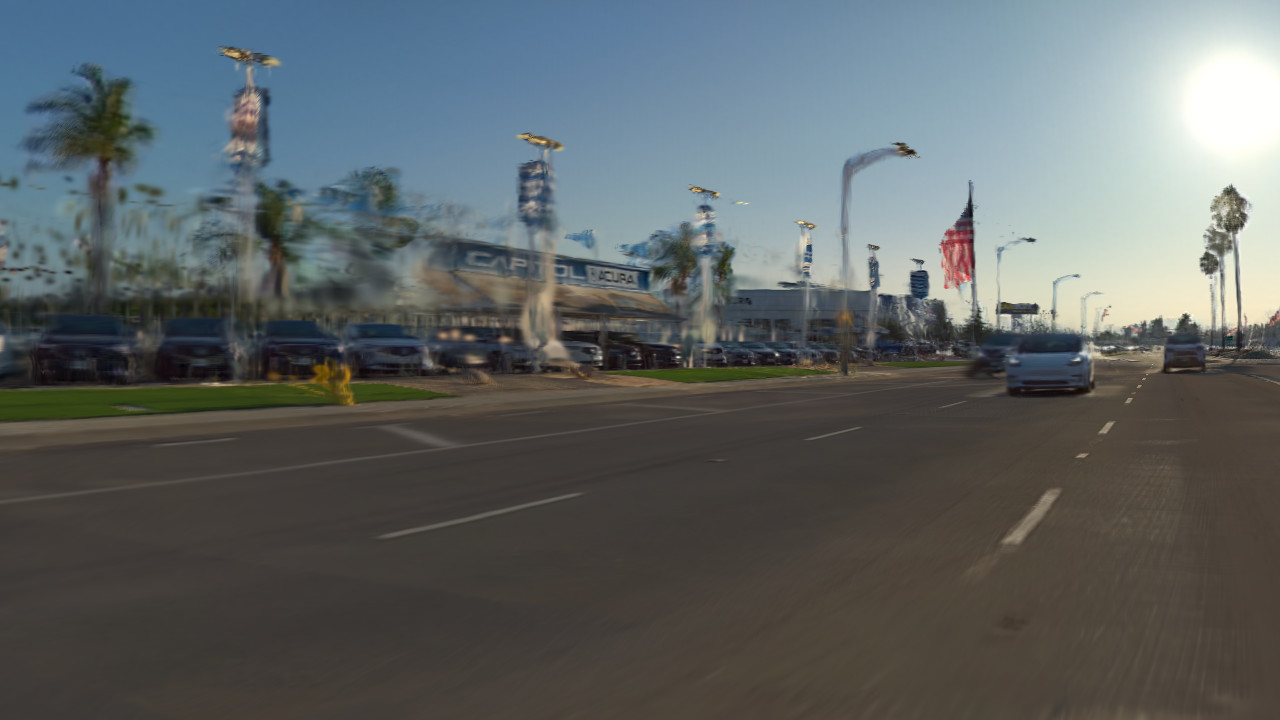} & \includegraphics[width=.16\linewidth]{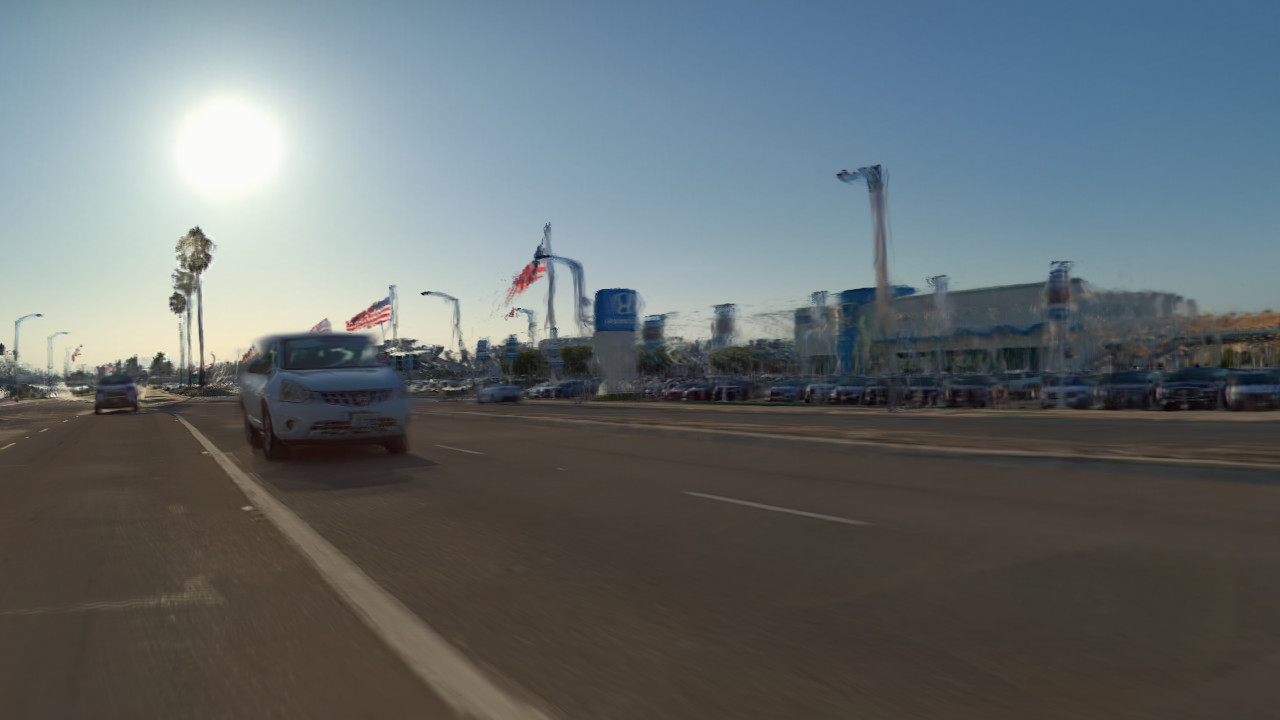}\\[6pt]
{\scriptsize Front-wide} & {\scriptsize Front-tele} & {\scriptsize Cross-left} & {\scriptsize Cross-right} & {\scriptsize Rear-right} & {\scriptsize Rear-left}\\[4pt]
\end{tabular}
\endgroup
\caption{\textbf{Six-camera Gaussian reconstruction with \method.} Each row shows approximately synchronized native-camera renderings of one example scene from the benchmark dataset. Gaussians are predicted feed-forward without per-scene optimization.}
\label{fig:gaussian_six_camera}
\end{figure}

\begin{figure}[!tp]
\centering
\begingroup
\setlength{\tabcolsep}{0pt}
\renewcommand{\arraystretch}{1}
\newcommand{\gaussiancomparisonimage}[2]{%
  \begin{tikzpicture}[baseline=(render.south)]
    \node[inner sep=0pt,outer sep=0pt] (render)
      {\includegraphics[width=.24\linewidth]{assets/gaussian/#1}};
    \node[overlay,anchor=north east,xshift=-1pt,yshift=-1pt,
      fill=black,fill opacity=.65,text=white,text opacity=1,
      inner sep=1.5pt,font=\sffamily\scriptsize] at (render.north east) {#2};
  \end{tikzpicture}%
}
\begin{tabular}{@{}c@{\hspace{.008\linewidth}}c@{\hspace{.024\linewidth}}c@{\hspace{.008\linewidth}}c@{}}
\multicolumn{2}{c}{\small Front-wide} & \multicolumn{2}{c}{\small Cross-left}\\[2pt]
\scriptsize Instant NuRec & \scriptsize Foundation4D & \scriptsize Instant NuRec & \scriptsize Foundation4D\\[4pt]
\multicolumn{4}{l}{\scriptsize Sunny interchange and elevated road}\\[2pt]
\gaussiancomparisonimage{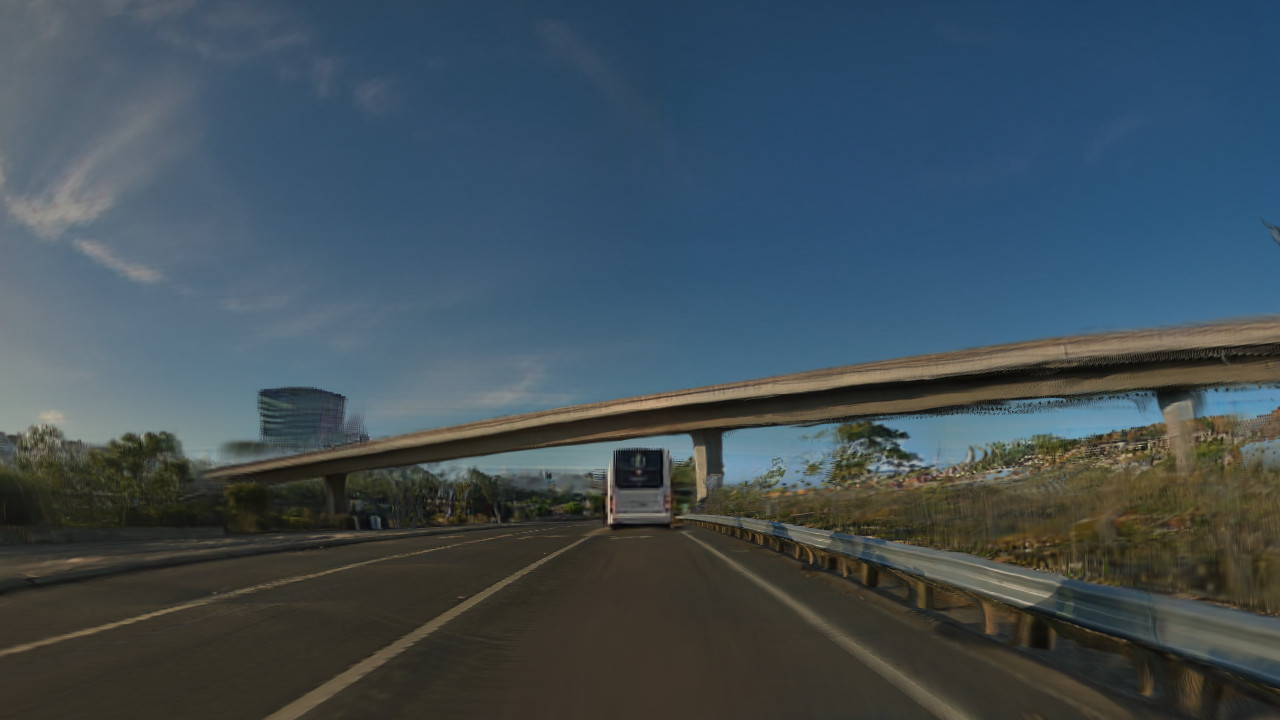}{21.25} &
\gaussiancomparisonimage{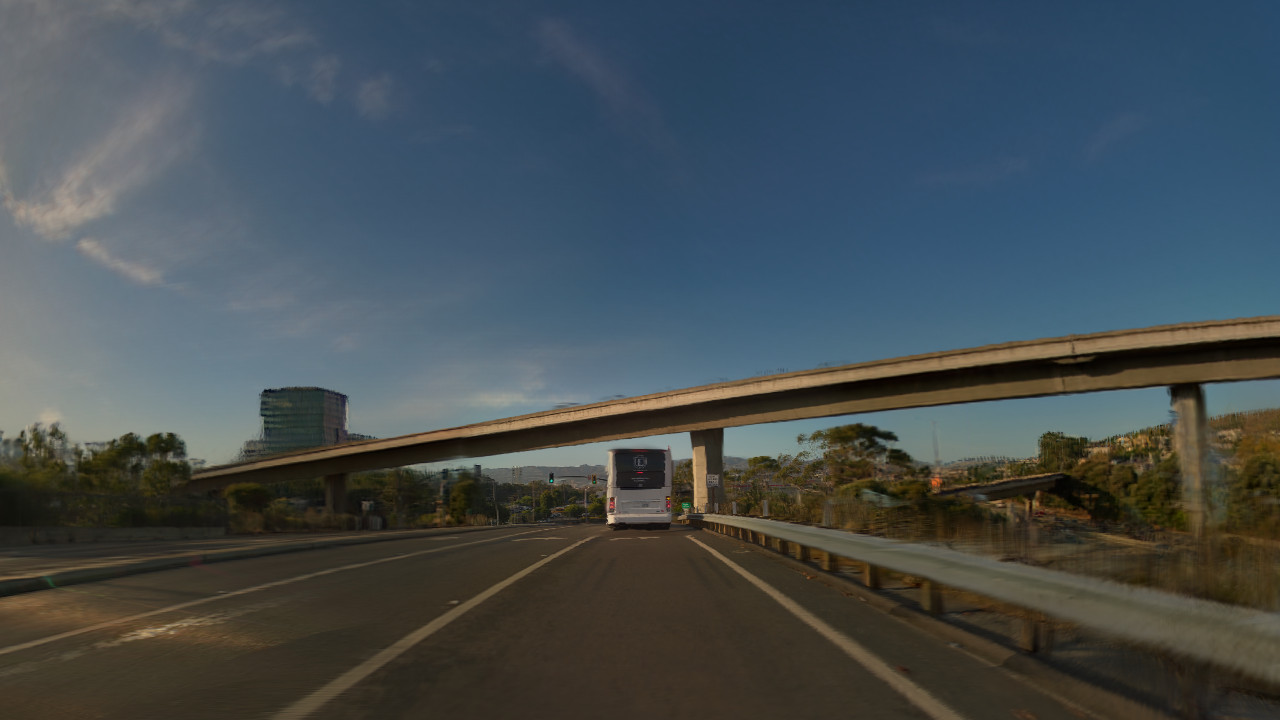}{25.60} &
\gaussiancomparisonimage{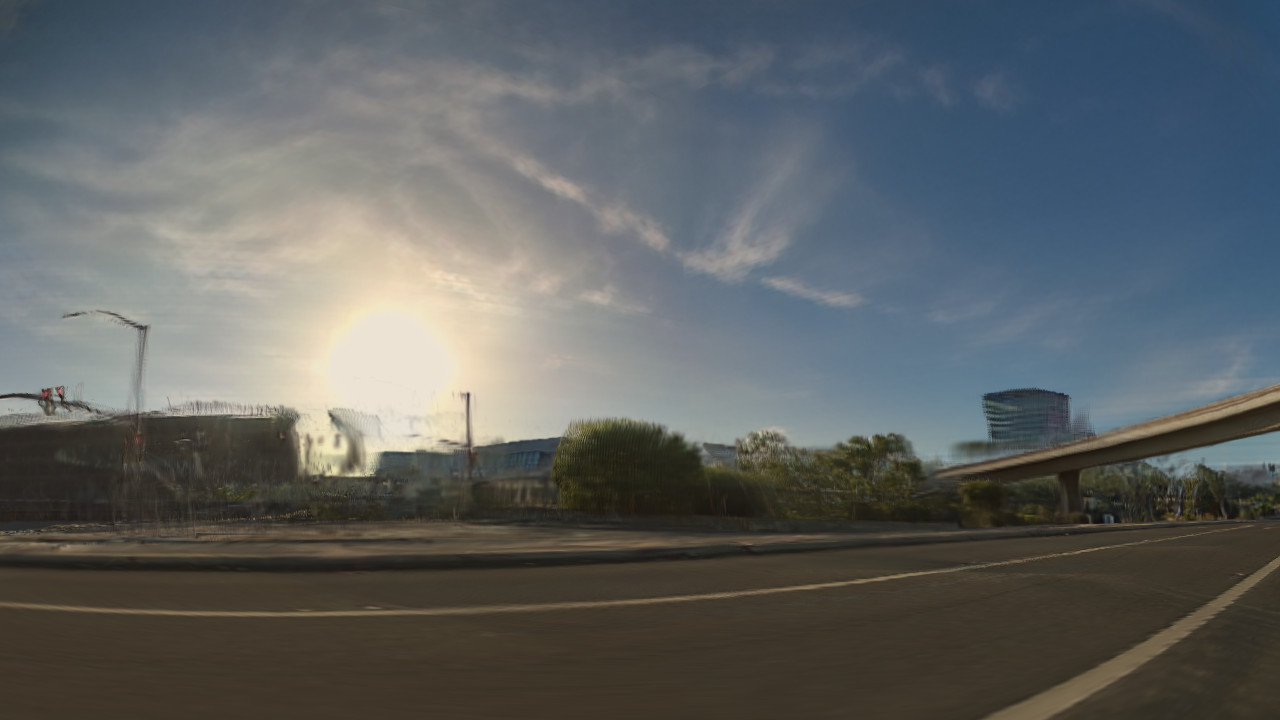}{22.07} &
\gaussiancomparisonimage{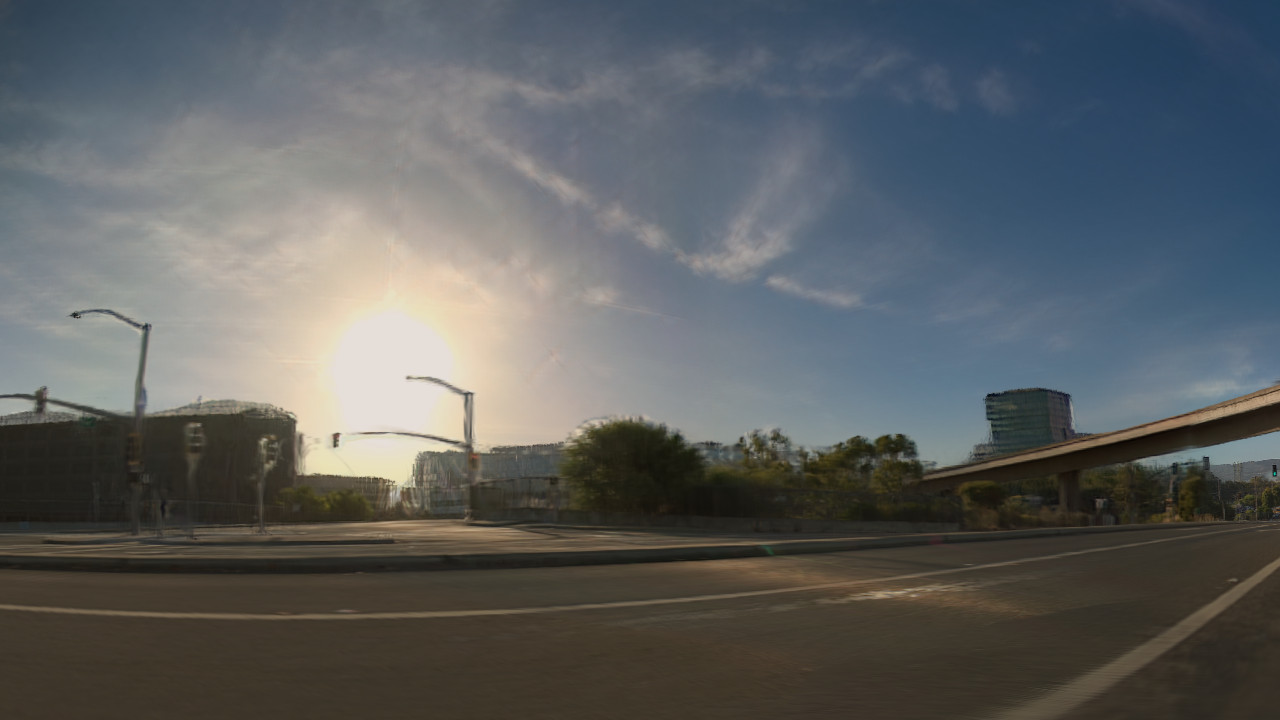}{24.05}\\[5pt]
\multicolumn{4}{l}{\scriptsize Overcast highway}\\[2pt]
\gaussiancomparisonimage{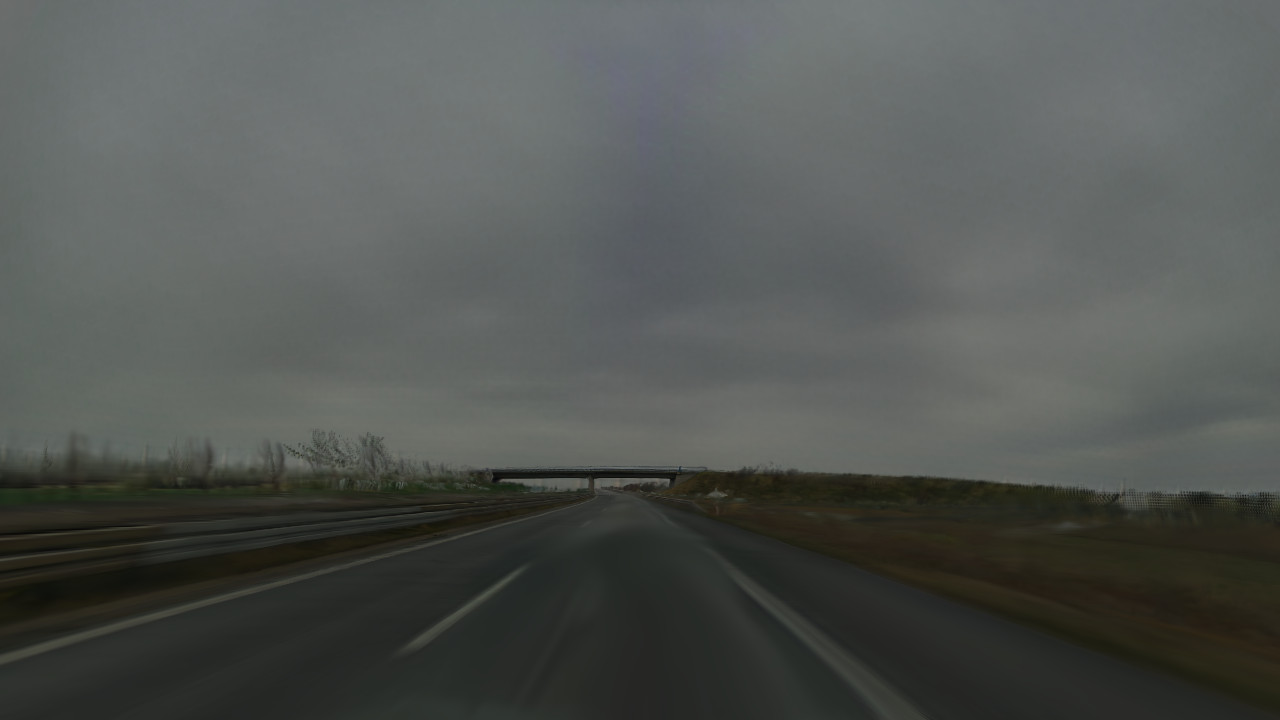}{29.68} &
\gaussiancomparisonimage{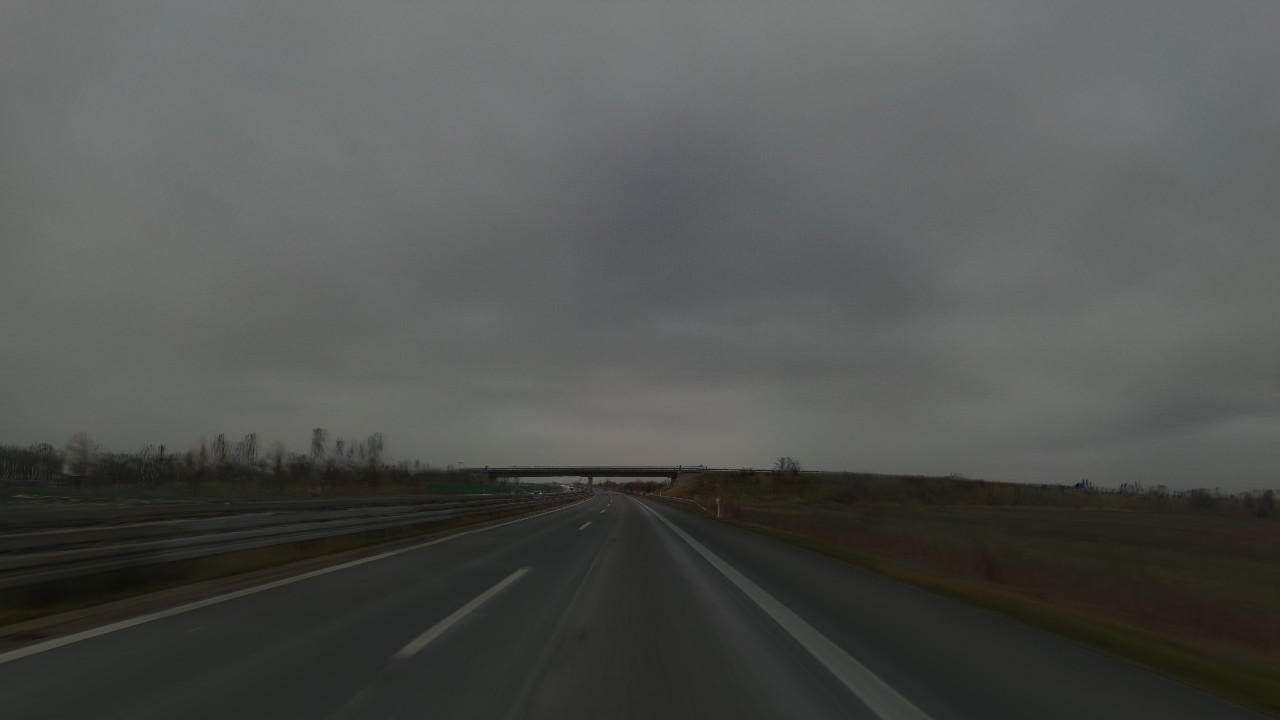}{32.17} &
\gaussiancomparisonimage{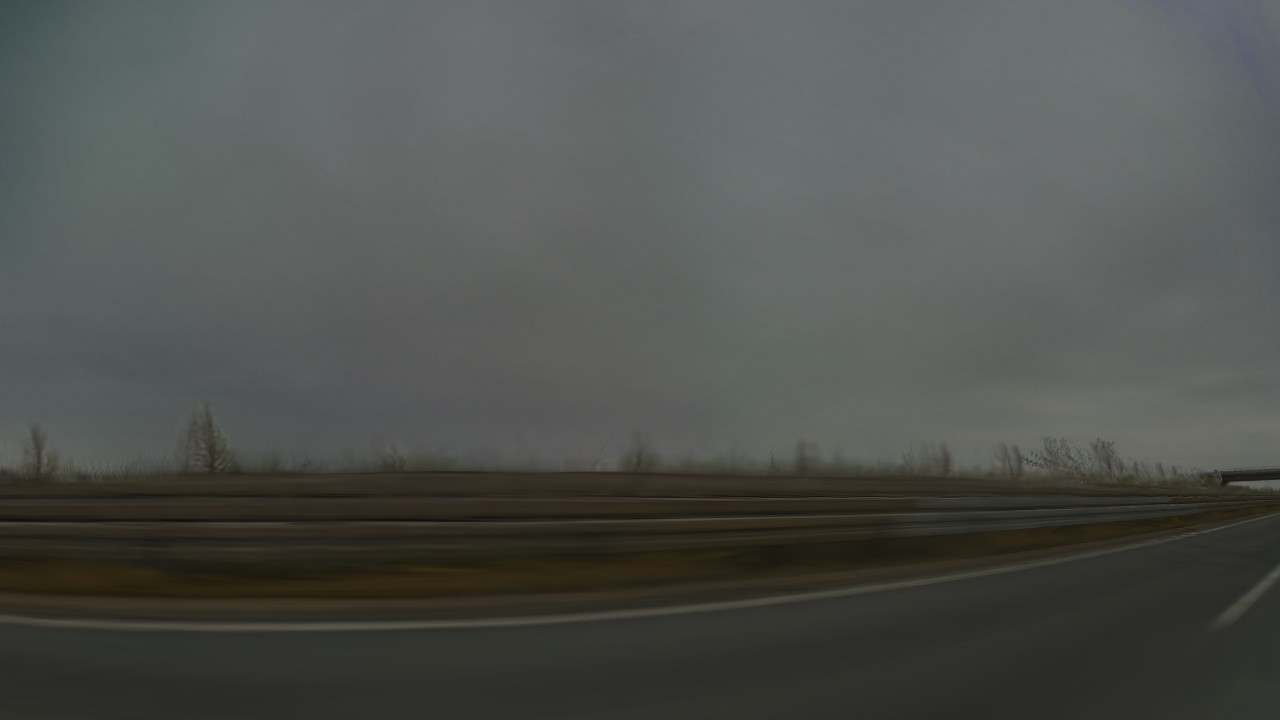}{29.67} &
\gaussiancomparisonimage{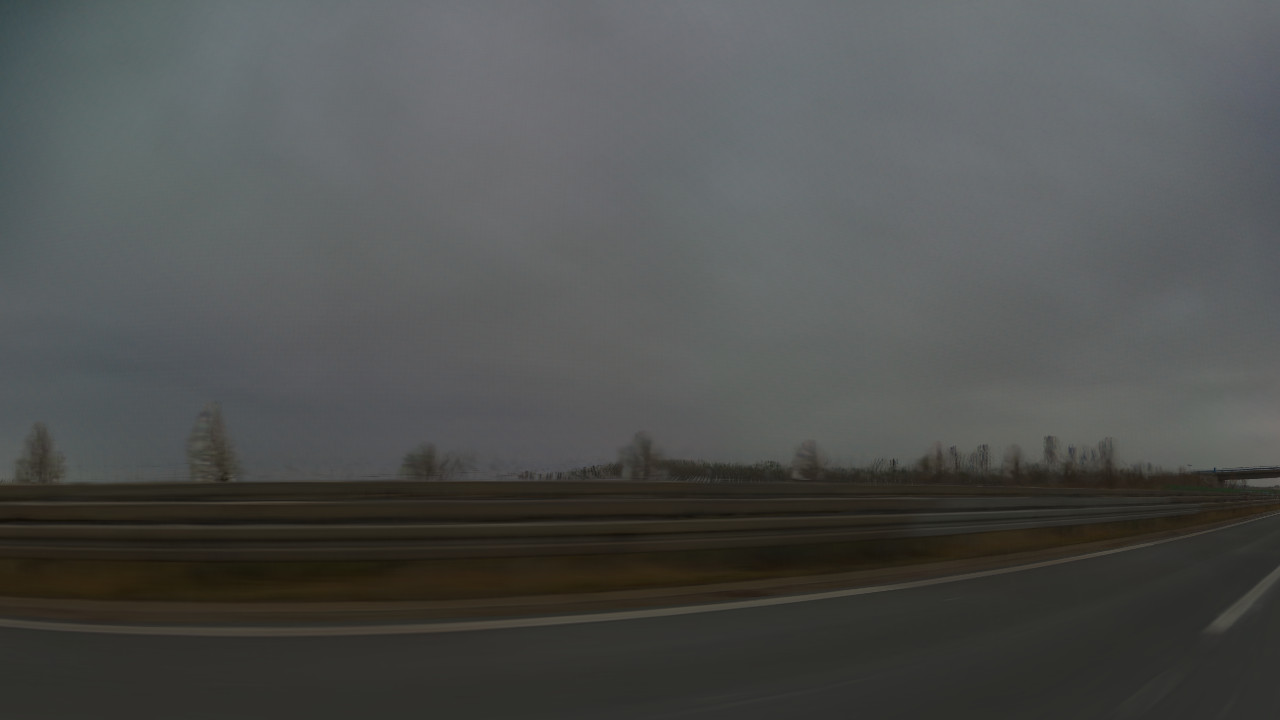}{31.80}\\[5pt]
\multicolumn{4}{l}{\scriptsize Sunny highway with road signs}\\[2pt]
\gaussiancomparisonimage{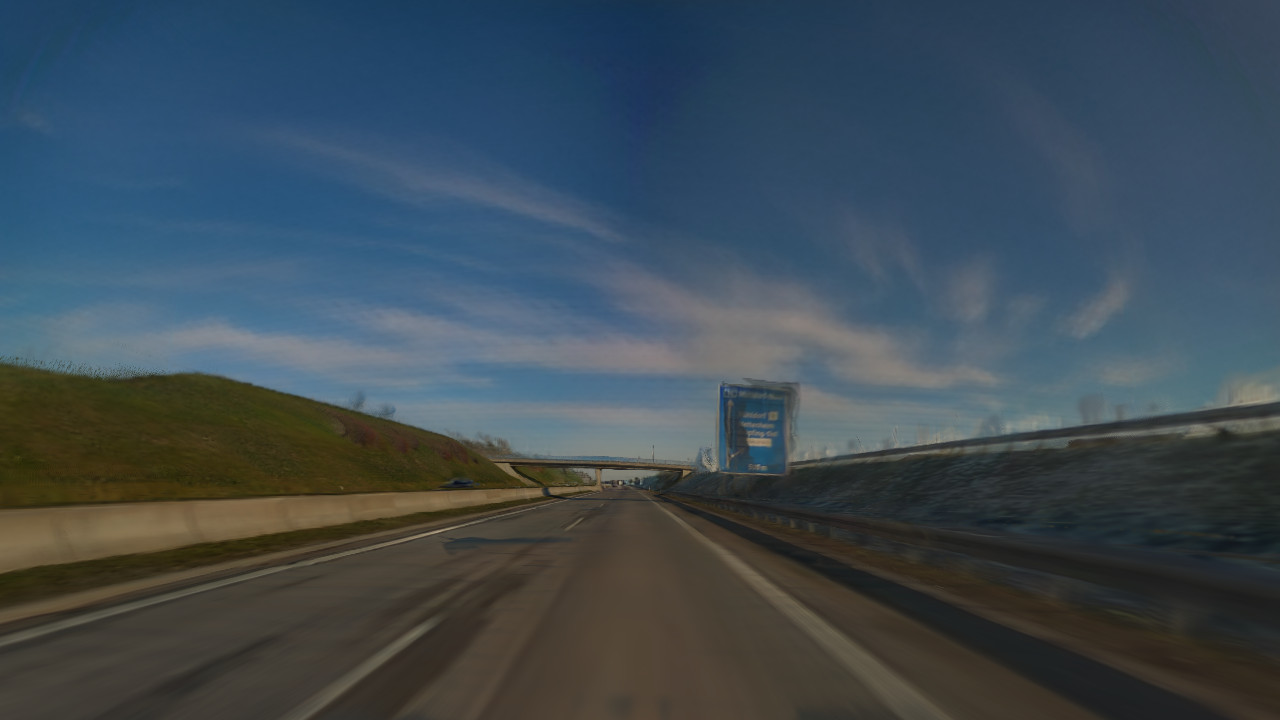}{26.47} &
\gaussiancomparisonimage{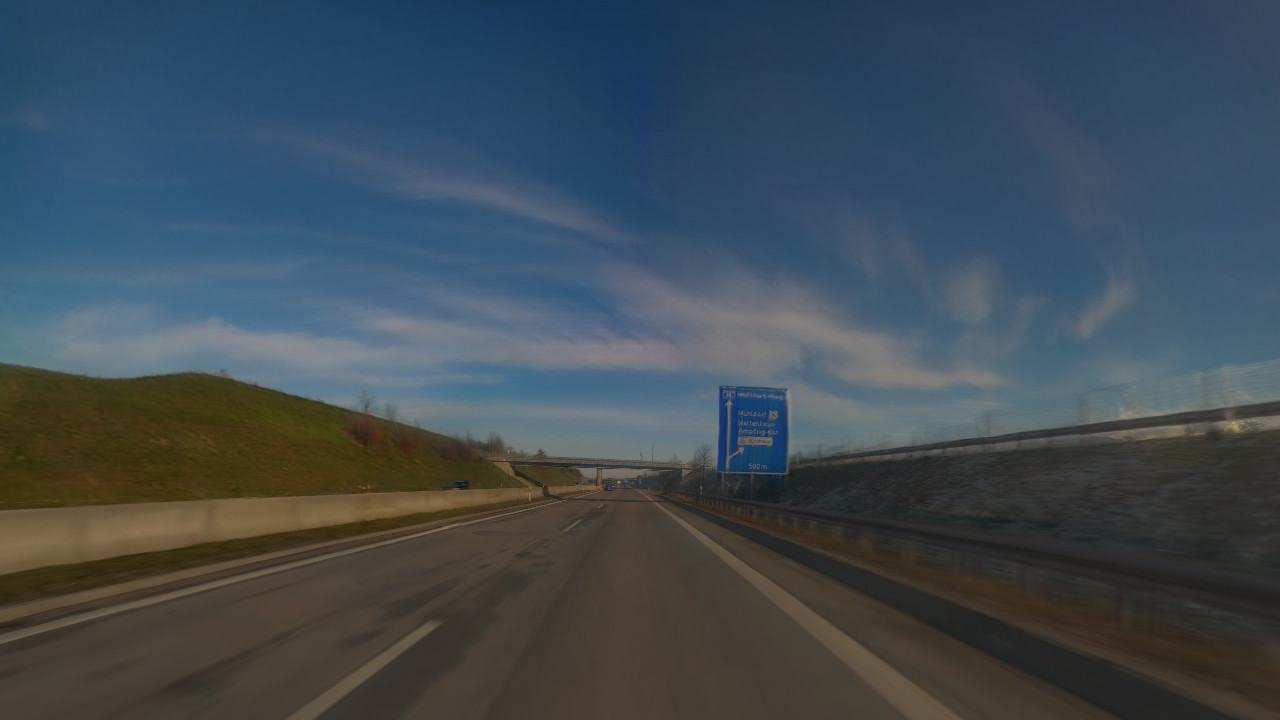}{27.09} &
\gaussiancomparisonimage{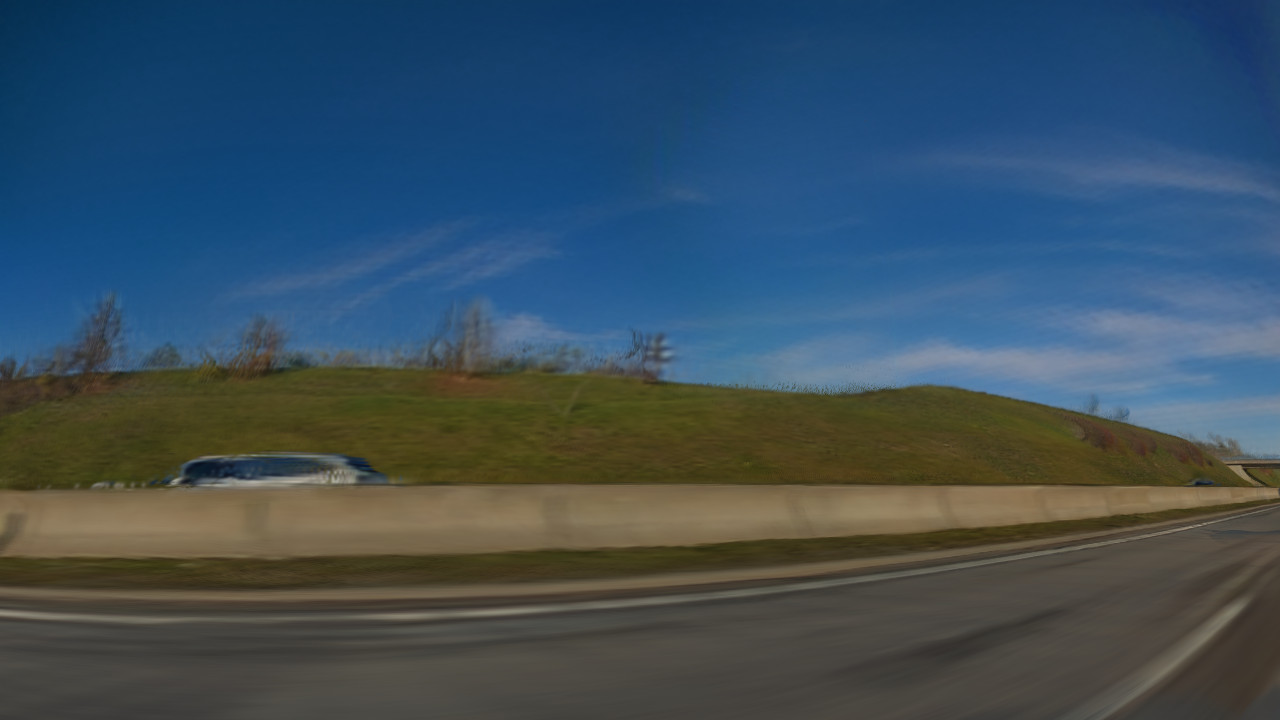}{25.26} &
\gaussiancomparisonimage{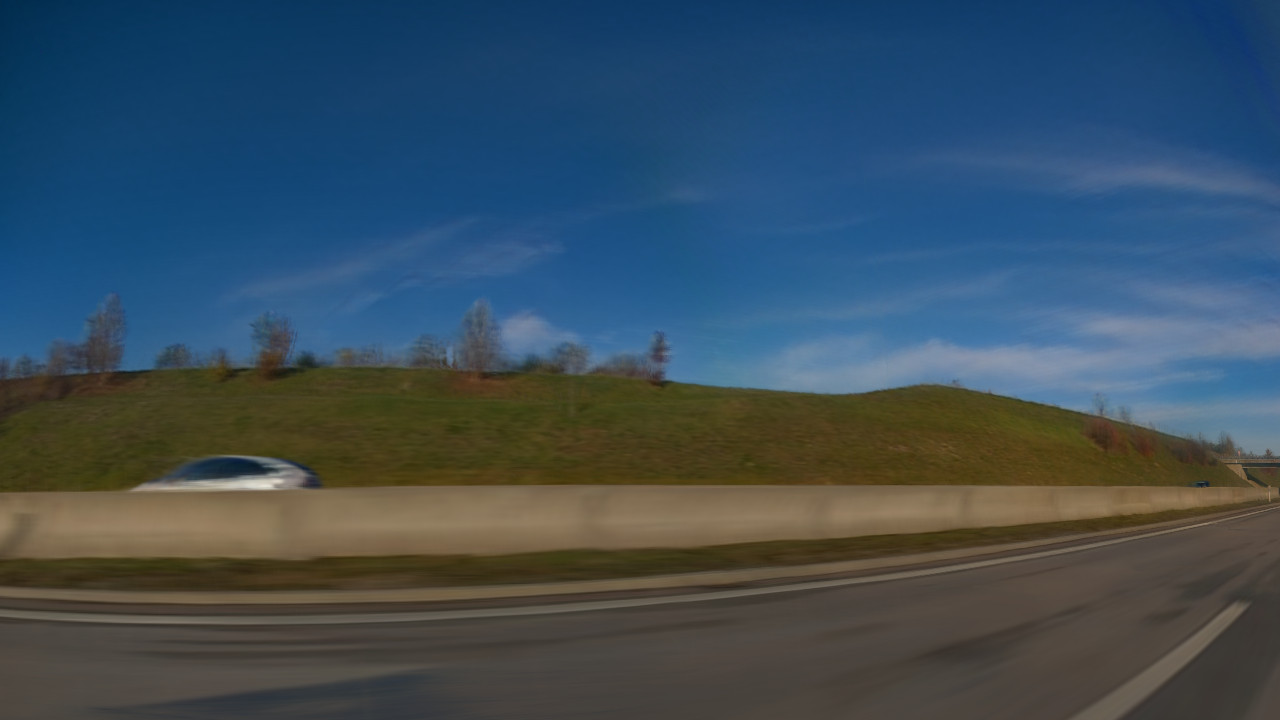}{25.99}\\[5pt]
\multicolumn{4}{l}{\scriptsize Sunny rural highway}\\[2pt]
\gaussiancomparisonimage{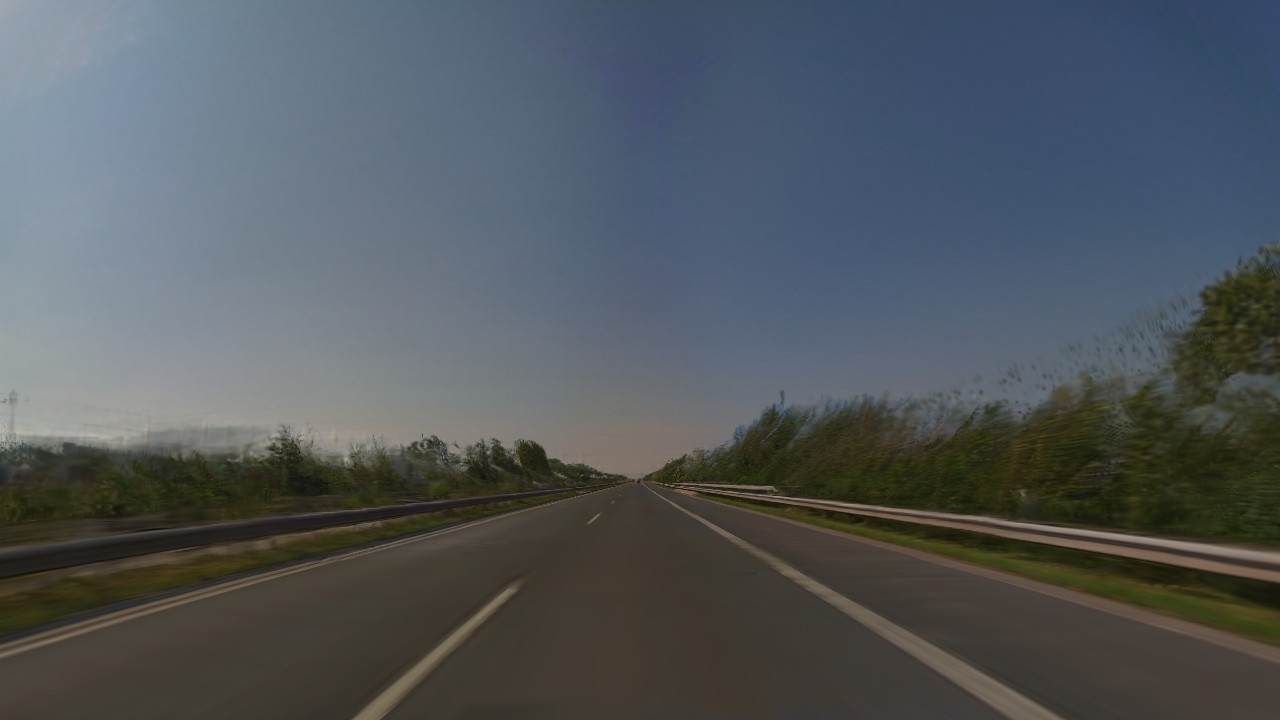}{25.69} &
\gaussiancomparisonimage{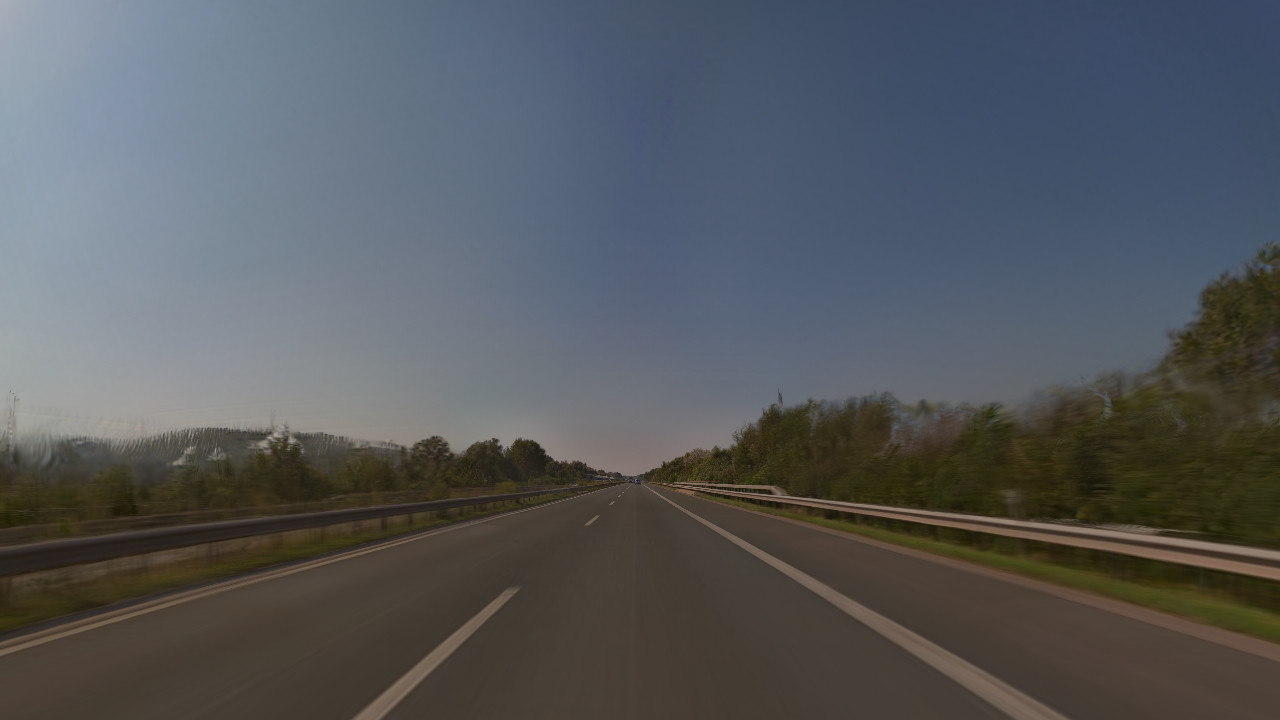}{27.45} &
\gaussiancomparisonimage{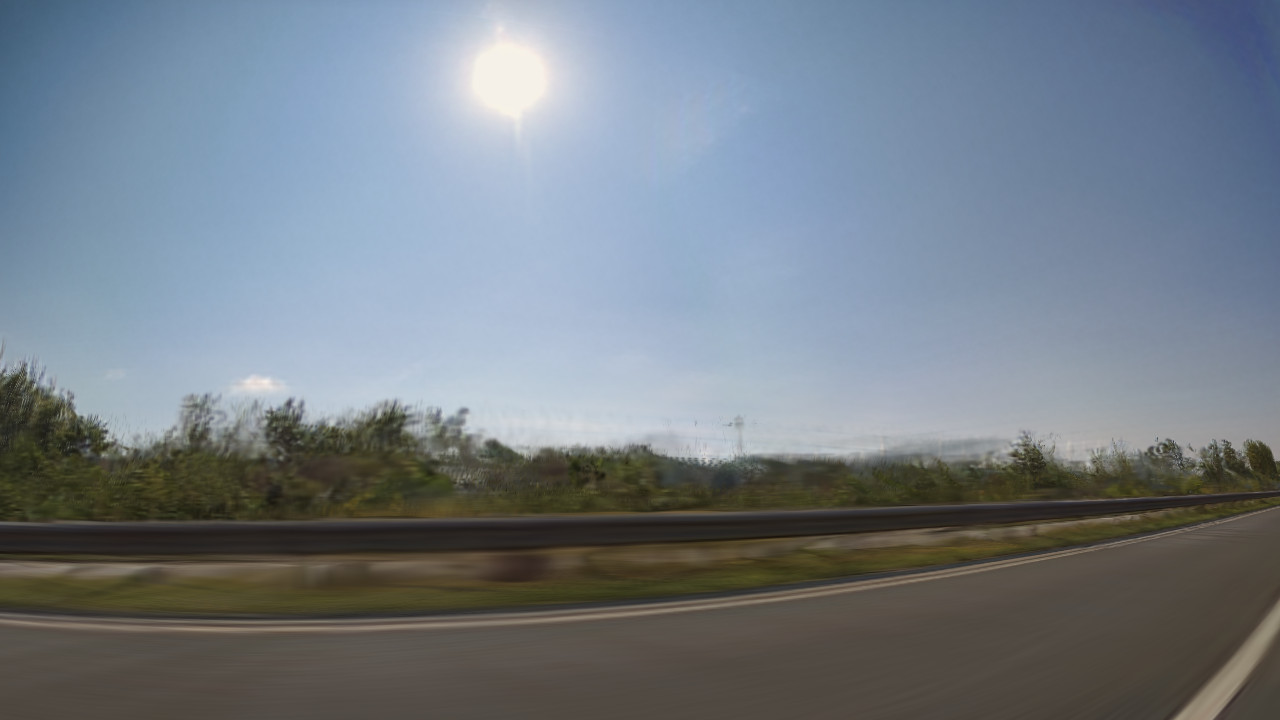}{25.28} &
\gaussiancomparisonimage{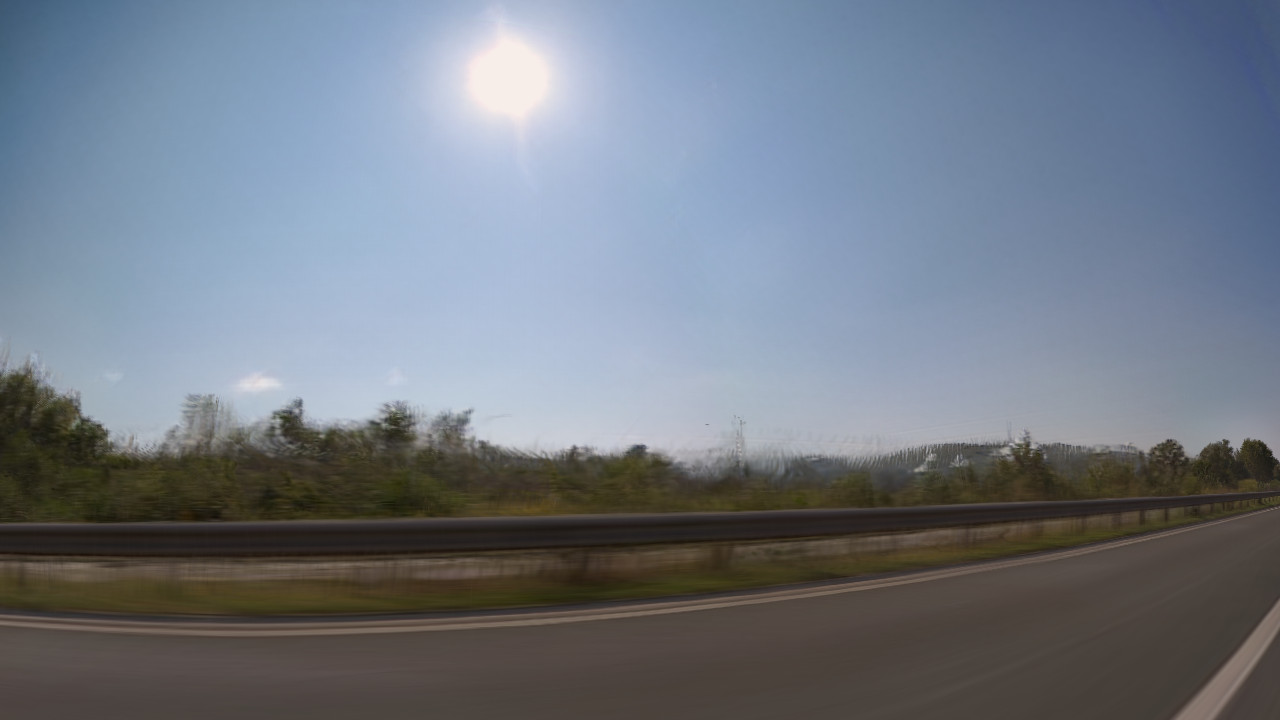}{26.02}\\
\end{tabular}
\endgroup
\caption{\textbf{Gaussian renders from four example scenes.} Each row compares Instant NuRec and Foundation4D at the same target exposure timestamp in the front-wide and cross-left cameras. Labels at the top right give per-image PSNR in dB.}
\label{fig:gaussian_kelvin_comparison_2}
\end{figure}

\begin{table}[!tp]
\centering
\small
\caption{\textbf{Gaussian reconstruction quality.}
Both Foundation4D variants use predicted actor tracks, constructed using
their respective ego poses. The GT-ego variant uses GT source and rendering
poses; the predicted-ego variant uses predicted source and rendering poses. Instant NuRec takes GT ego poses as input. Scores average five common cameras
equally.}
\label{tab:gaussian_rgb}
\begingroup
\setlength{\tabcolsep}{4pt}
\begin{tabular*}{\linewidth}{@{\extracolsep{\fill}}llrrrrrr@{}}
\toprule
Target set & Region
& \multicolumn{2}{c}{Instant NuRec}
& \multicolumn{2}{c}{\shortstack{Foundation4D\\with GT ego pose}}
& \multicolumn{2}{c}{\shortstack{Foundation4D\\with predicted ego pose}}\\
\cmidrule(lr){3-4}\cmidrule(lr){5-6}\cmidrule(lr){7-8}
& & PSNR (dB) $\uparrow$ & SSIM $\uparrow$
& PSNR (dB) $\uparrow$ & SSIM $\uparrow$
& PSNR (dB) $\uparrow$ & SSIM $\uparrow$\\
\midrule
Overall & All
& 22.66 & 0.78 & \textbf{24.02} & \textbf{0.80} & 23.94 & 0.80\\
& Static
& 22.83 & 0.78 & \textbf{24.34} & \textbf{0.81} & 24.27 & 0.81\\
& Dynamic
& 20.01 & 0.72 & \textbf{20.03} & \textbf{0.73} & 19.98 & 0.73\\
\midrule
Interpolation & All
& 22.86 & 0.78 & \textbf{23.98} & \textbf{0.80} & 23.91 & 0.80\\
& Static
& 23.04 & 0.78 & \textbf{24.30} & \textbf{0.81} & 24.22 & 0.80\\
& Dynamic
& \textbf{20.20} & 0.72 & 20.05 & \textbf{0.73} & 20.03 & 0.73\\
\midrule
Extrapolation & All
& 20.34 & 0.75 & \textbf{20.77} & \textbf{0.76} & 20.65 & 0.76\\
& Static
& 20.51 & 0.76 & \textbf{21.20} & \textbf{0.77} & 21.09 & 0.76\\
& Dynamic
& \textbf{17.71} & \textbf{0.66} & 16.00 & 0.63 & 15.84 & 0.62\\
\bottomrule
\end{tabular*}
\endgroup
\end{table}

We evaluate whether the Gaussian head of \method produces useful multi-camera 3D Gaussian reconstructions after joint fine-tuning. We compare the \method Gaussian reconstruction against Instant NuRec~\cite{nvidia2026instantnurec}, using its released multi-view 1.1.0 checkpoint without further fine-tuning, and the recommended inference configuration. Our primary \method result uses predicted ego poses and predicted actor tracks from the same model, with rendering cameras obtained from the predicted trajectory. All \method configurations reuse the same checkpoint after 10k updates of joint fine-tuning; no per-clip optimization is performed.

\subsubsection{Evaluation protocol}

Our evaluation covers 99 clips from our \method internal reconstruction benchmark: 49 NRE and 50 PlanNet clips.
Both models are scored against the same recorded target exposures and RGB references, at native-wide resolution with a height of 720 pixels and a width of 1278 or 1280 pixels, preserving the camera aspect ratio. Scores average the five common cameras equally: front-wide, cross-left, cross-right, rear-left, and rear-right. Front-tele is excluded from scoring because the Instant NuRec model was not trained for that camera. \method uses six input cameras, whereas Instant NuRec uses five; each retains its own context-window recipe.

For \method, source camera poses, actor tracks, and rendering-camera poses are all constructed from model predictions. Known camera intrinsics and fixed camera-to-rig extrinsics are retained.
Target camera poses are queried from the predicted rig trajectory at each sensor timestamp and at the rolling-shutter exposure boundaries. No ground-truth ego poses, actor tracks, initial-pose anchors, or ground-truth-based trajectory or scale fits are used to construct the scene or rendering cameras.

Instant NuRec retains its original reference protocol, which requires ground-truth ego poses as input, alongside the RGB images.
Consequently, the comparison measures image reconstruction under the
respective inference configurations rather than an isolated Gaussian-decoder comparison.

We sample target frames at 1 Hz from the 30 Hz recordings. Each model predicts Gaussians in non-overlapping context windows. \cref{tab:gaussian_rgb} distinguishes three target sets. Overall reconstruction includes input frames and other targets, using the closest context window for each target. The matched interpolation subset excludes input frames and retains other targets inside both models' source intervals. The extrapolation set renders future frames from earlier context windows, rather than assigning them to the closest window.

For each camera, PSNR pools squared RGB errors and valid-value counts across clips before applying the logarithm; SSIM pools valid spatial-window statistics. We then average the five camera scores. Dynamic regions are defined by semantic masks for vehicles and other movable objects, including parked vehicles. This category does not isolate objects that are actually moving. The same mask is applied to both models. Since the models use different input-camera counts and context recipes, this benchmark compares their reconstruction systems rather than isolating the Gaussian decoder.

\subsubsection{Quantitative comparison}

As summarized in \cref{tab:gaussian_rgb}, \method demonstrates substantial improvements over Instant NuRec. Notably, even without ground-truth poses or tracks, our predicted input variant outperforms Instant NuRec (which consumes GT ego poses) by 1.28\,dB in overall PSNR (23.94 vs. 22.66\,dB) and 0.02 in SSIM. The performance boost is particularly pronounced in static regions across all evaluation splits (overall, interpolation, and extrapolation). While dynamic regions exhibit slightly lower overall PSNR (19.98 vs. 20.01\,dB), SSIM is consistently improved (0.73 vs. 0.72). Under a matching GT-ego pose setting, \method attains the highest score of 24.02\,dB. These findings highlight our method's robustness in reconstructing static environments, while pointing to dynamic novel-view extrapolation as an avenue for future exploration. All evaluation frames are rendered from the Gaussian predictions of the relevant context windows, rather than from Gaussians merged across the full scene. Full scene merging requires ego pose and is susceptible to pose prediction errors, similar to what is observed in \cref{sec:detection_ablations}.

\subsubsection{Qualitative results}

The multi-camera renderings in \cref{fig:gaussian_six_camera} illustrate the broad coverage and strong multi-view geometric consistency of \method Gaussians across all six native viewpoints. In addition, the side-by-side comparisons in \cref{fig:gaussian_kelvin_comparison_2} demonstrate noticeable visual improvements over Instant NuRec in road surface texture, distant background geometry, and sky appearance under diverse lighting conditions. Note that the per-frame PSNR annotations in \cref{fig:gaussian_kelvin_comparison_2} reflect individual scene performance, directly corroborating the overall pooled results in \cref{tab:gaussian_rgb}.

\FloatBarrier

\subsection{Unified model}
\label{sec:joint_ablation}
\begin{figure}[t]
\centering
\includegraphics[width=\textwidth]{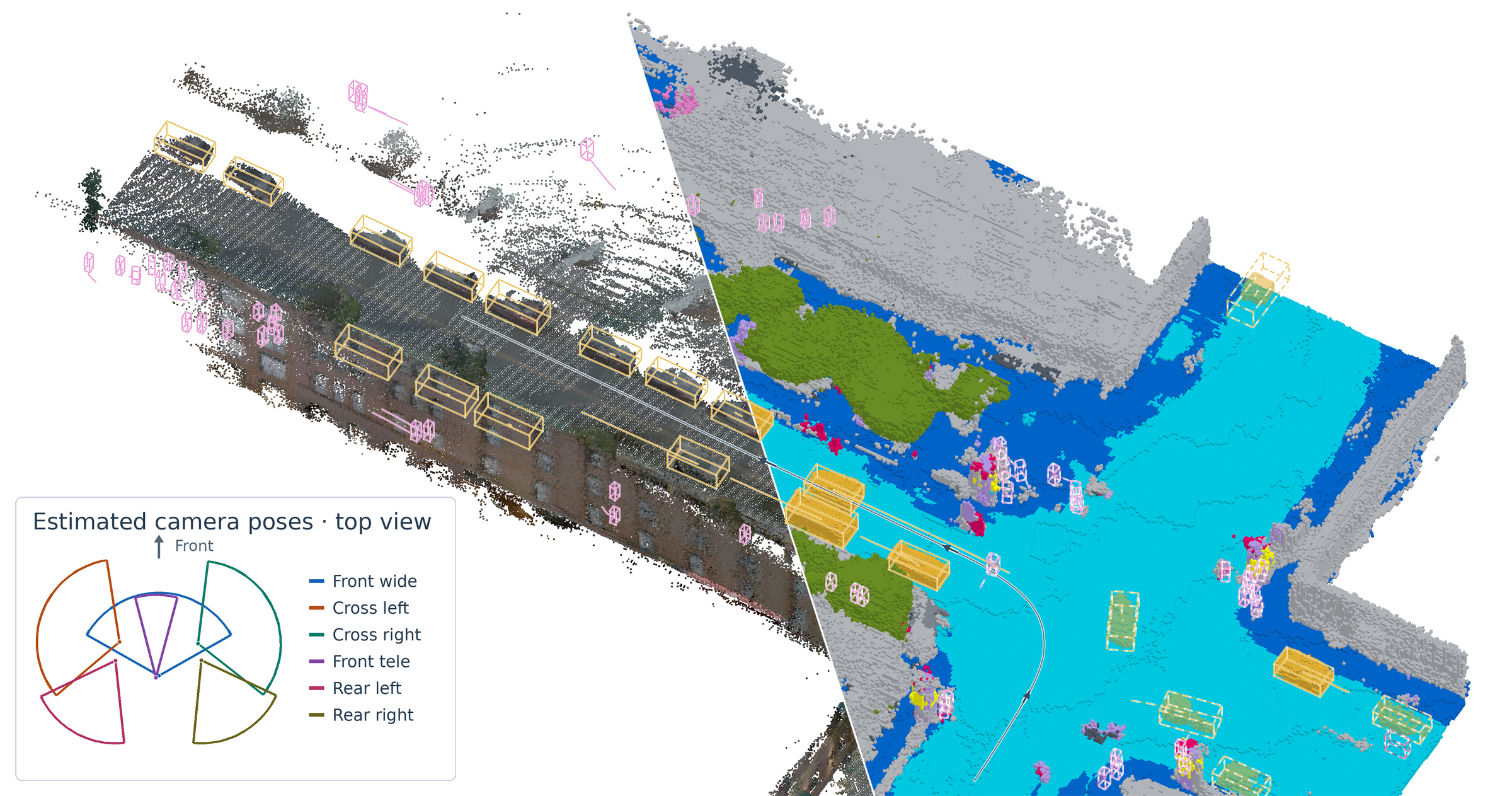}
\caption{\textbf{Scene-level geometry, occupancy, and tracks.} Pointmaps (left) and occupancy (right) span a 20-second clip, with car geometry and occupancy accumulated in tracked object frames. Solid cuboids mark the final scene at $\tau=19.97$\,s; dashed cuboids and translucent car and pedestrian occupancy show $\tau=1.50$\,s. The ego route uses optimized camera poses and fixed camera-to-rig calibration. 
}
\label{fig:teaser_3d}
\end{figure}

The scene-level visualization in \cref{fig:teaser_3d} brings together geometry, semantic occupancy, and tracked objects. We study interactions between geometry, detection, and occupancy in a four-camera setup, then profile the final six-camera unified model with all task heads enabled.

\subsubsection{Task-combination study}
We train all seven combinations of geometry (GEO), detection (DET), and occupancy (OCC), as summarized in \cref{tab:joint_task_ablation}. Each configuration is initialized from the same LingBot-Map stage-1 weights and trained in a single stage for 80,000 updates, using front-wide, front-tele, cross-left, and cross-right cameras. Data-to-task assignments and multi-task loss weights follow \cref{sec:training}; single-task runs use unit loss weight. OCC-only omits MLMCF, which lacks occupancy labels.

\label{sec:unified_evaluation_details}
All configurations share the same evaluation protocol for each task. We evaluate geometry on 301 CICD301 clips using the metrics defined in \cref{sec:geometry_metrics}, including rig mAA@3. Detection is evaluated on 300 class-stratified PlanNet validation clips using top-class mAP over automobile, person, rider, bus, and heavy truck, with 3D center-distance matching. For occupancy, we report empty-aware semantic mIoU on 99 NRE clips: empty voxels enter the confusion matrix, while the mean is taken over occupied semantic classes.

\paragraph{Findings.}

\begin{table*}[!tp]
\centering
\small
\setlength{\tabcolsep}{3pt}
\sisetup{detect-weight=true,table-number-alignment=center}
\caption{\textbf{Perception performance across training-task combinations.} All configurations use four cameras and are trained for 80,000 updates.}
\label{tab:joint_task_ablation}
\label{tab:joint_rig_ablation}
\begin{tabular*}{\linewidth}{@{\extracolsep{\fill}}ccc S[table-format=2.2] S[table-format=2.2] S[table-format=1.3] S[table-format=1.2] S[table-format=2.2] S[table-format=1.2] S[table-format=1.3]@{}}
\toprule
\multicolumn{3}{c}{Training tasks}
& \multicolumn{1}{c}{Detection} & \multicolumn{1}{c}{Occupancy} & \multicolumn{1}{c}{Depth} & \multicolumn{1}{c}{Trajectory} & \multicolumn{3}{c}{Rig pose} \\
\cmidrule(lr){1-3}\cmidrule(lr){4-4}\cmidrule(lr){5-5}\cmidrule(lr){6-6}\cmidrule(lr){7-7}\cmidrule(lr){8-10}
GEO & DET & OCC & \multicolumn{1}{c}{Top-mAP $\uparrow$} & \multicolumn{1}{c}{mIoU $\uparrow$} & \multicolumn{1}{c}{AbsRel $\downarrow$} & \multicolumn{1}{c}{ATE $\downarrow$} & \multicolumn{1}{c}{mAA@3 $\uparrow$} & \multicolumn{1}{c}{RTE $\downarrow$} & \multicolumn{1}{c}{RRE $\downarrow$} \\
& & & \multicolumn{1}{c}{(\%)} & \multicolumn{1}{c}{(\%)} & & \multicolumn{1}{c}{(cm)} & \multicolumn{1}{c}{(\%)} & \multicolumn{1}{c}{(cm)} & \multicolumn{1}{c}{($^\circ$)} \\
\midrule
$\checkmark$ & --- & --- & {---} & {---} & \bfseries 0.123 & 5.00 & 37.09 & 7.53 & \bfseries 0.898 \\
--- & $\checkmark$ & --- & 40.94 & {---} & {---} & {---} & {---} & {---} & {---} \\
--- & --- & $\checkmark$ & {---} & 11.00 & {---} & {---} & {---} & {---} & {---} \\
\midrule
$\checkmark$ & $\checkmark$ & --- & 41.30 & {---} & 0.127 & \bfseries 4.38 & \bfseries 43.17 & \bfseries 6.52 & 0.966 \\
$\checkmark$ & --- & $\checkmark$ & {---} & \bfseries 11.82 & 0.124 & 4.72 & 40.46 & 7.08 & 0.909 \\
--- & $\checkmark$ & $\checkmark$ & 41.64 & 10.49 & {---} & {---} & {---} & {---} & {---} \\
\midrule
$\checkmark$ & $\checkmark$ & $\checkmark$ & \bfseries 42.14 & 11.53 & 0.127 & 4.55 & 42.05 & 6.64 & 0.957 \\
\bottomrule
\end{tabular*}
\end{table*}

Detection benefits from co-training with geometry and occupancy, with top-class mAP increasing from $40.94\%$ for DET-only to $42.14\%$ when both tasks are included. Co-training with either task also improves detection. These consistent gains suggest that geometry and occupancy supervision help ground the shared representation in 3D structure, providing useful spatial cues for object detection.

Occupancy, in turn, benefits from co-training with geometry: mIoU rises from $11.00\%$ to $11.82\%$. The two tasks describe complementary aspects of the same scene. Depth and pointmaps provide dense, pixel-aligned supervision of observed surfaces, while occupancy extends this reconstruction to volumetric scene completion. Learning these surface constraints may therefore improve the shared features used for occupancy prediction. Detection alone does not provide the same benefit: DET+OCC reaches $10.49\%$ mIoU, whereas including geometry raises it to $11.53\%$.

Geometry also benefits in trajectory and rig translation accuracy. Co-training with detection reduces trajectory ATE from $5.00$ to $4.38$\,cm and rig RTE from $7.53$ to $6.52$\,cm, while achieving the highest rig mAA@3. The other combinations that include geometry improve these metrics as well. A plausible explanation is that detection and occupancy supervision strengthen the shared scene features used by the pose heads. These gains come with a trade-off: GEO-only retains the lowest depth AbsRel and rig rotation error.

\subsubsection{Runtime analysis}
\label{sec:unified_runtime}
We profile the final unified model on one NVIDIA H100 80GB GPU with the mixed-precision settings in \cref{sec:training_implementation}. All task heads run together on six-camera windows from one CICD301 scene, at $1036\times1036$ resolution and $T\in\{24,36,48\}$ timestamps per camera. We report median forward times over five runs after two warmup iterations. Timings exclude data loading and preparation, Gaussian rasterization, losses, metrics, and output export. Peak memory refers to PyTorch allocated GPU memory.

\begin{table}[!tp]
\centering
\small
\setlength{\tabcolsep}{8pt}
\renewcommand{\arraystretch}{1.08}
\caption{\textbf{Forward runtime and peak memory of the unified model.} Median times (s) are measured per six-camera window on one H100 80GB; $T$ counts timestamps per camera. Component medians need not sum to the independently measured full-forward median.}
\label{tab:unified_runtime}
\begin{threeparttable}
\begin{tabular*}{\linewidth}{@{\extracolsep{\fill}}lrrr@{}}
\toprule
Component & $T=24$ & $T=36$ & $T=48$ \\
\midrule
Image encoder & 2.218 & 3.322 & 4.455 \\
Spatiotemporal aggregator & 1.248 & 2.175 & 3.305 \\
Geometry heads & 1.786 & 2.673 & 3.559 \\
Shared DET/OCC neck & 0.014 & 0.021 & 0.028 \\
Detection head & 0.090 & 0.128 & 0.168 \\
Occupancy head$^{\dagger}$ & 1.060 & 1.576 & 2.104 \\
3DGS heads (without rendering) & 3.255 & 4.578 & 6.155 \\
\midrule
\textbf{Full forward (s)} & \textbf{9.673} & \textbf{14.480} & \textbf{19.782} \\
Peak allocated memory (GiB) & 32.47 & 45.44 & 58.41 \\
\bottomrule
\end{tabular*}
\begin{tablenotes}[flushleft]
\footnotesize
\item[$\dagger$] Dense occupancy outputs are decoded and released one timestamp at a time.
\end{tablenotes}
\end{threeparttable}
\end{table}

As shown in \cref{tab:unified_runtime}, the 3DGS heads account for about one third of forward time, followed by the image encoder and geometry heads. Doubling the window from 24 to 48 timestamps roughly doubles total forward time, and peak allocated memory increases from $32.47$ to $58.41$\,GiB.

\section{Application to lidarless NuRec}
\label{sec:application}
We apply \method to lidarless NuRec~\cite{nvidia2026nurec}, using its camera trajectories, dense geometry, and actor tracks as priors for scene reconstruction. This application complements the feed-forward Gaussian decoder in \cref{sec:gaussian_decoder}: NuRec optimizes a separate Gaussian representation for each scene against the recorded images.

\subsection{Reconstruction workflow}
\label{sec:application_pipeline}
\begin{figure}[!tp]
\centering
\begingroup
\definecolor{lfblue}{HTML}{1F5AA6}
\definecolor{lfteal}{HTML}{168A8A}
\definecolor{lforange}{HTML}{D97706}
\resizebox{\linewidth}{!}{%
\begin{tikzpicture}[
  box/.style={draw=#1,fill=#1!6,rounded corners=3pt,align=center,
    text width=2.65cm,minimum height=1.15cm,font=\small,inner sep=5pt},
  f4dbox/.style={draw=#1,fill=#1!6,rounded corners=3pt,align=center,
    text width=2.85cm,minimum height=2.4cm,font=\small,inner sep=5pt},
  nurecbox/.style={draw=#1,fill=#1!6,rounded corners=3pt,align=center,
    text width=2.75cm,minimum height=2.5cm,font=\small,inner sep=5pt},
  flow/.style={-{Stealth[length=2.2mm]},line width=.8pt,draw=black!65},
  label/.style={font=\scriptsize,align=center,text=black!80}]

  \node[box=lfblue] (input) at (0,0) {\textbf{Recorded inputs}\\[2pt]Multi-camera RGB\\GPS / IMU};

  \node[f4dbox=lfteal] (f4d) at (4.5,0) {\textbf{Scene priors}\\[3pt]Foundation4D\\+ video-level\\processing};

  \node[nurecbox=lforange] (nurec) at (10.4,0) {\textbf{NuRec}\\[3pt]Per-scene\\Gaussian\\optimization};

  \node[box=lfblue,minimum height=2.5cm] (out) at (14.6,0) {\textbf{Reconstructed 4D scene}\\[3pt]Novel-view rendering};

  \draw[flow] (input.east) -- (f4d.west);

  \draw[flow] ([yshift=0.7cm]f4d.east) -- node[label,above]{camera poses} ([yshift=0.7cm]nurec.west);
  \draw[flow] (f4d.east) -- node[label,above]{depth} (nurec.west);
  \draw[flow] ([yshift=-0.7cm]f4d.east) -- node[label,above]{actor tracks} ([yshift=-0.7cm]nurec.west);

  \draw[flow] (nurec.east) -- (out.west);

  \draw[flow] (input.south) -- (0,-1.85) -- node[label,above,pos=0.5]{RGB frames} (10.4,-1.85) -- (nurec.south);

\end{tikzpicture}}
\par\vspace{6pt}
\begin{minipage}[t]{.29\linewidth}\centering
  \includegraphics[width=\linewidth]{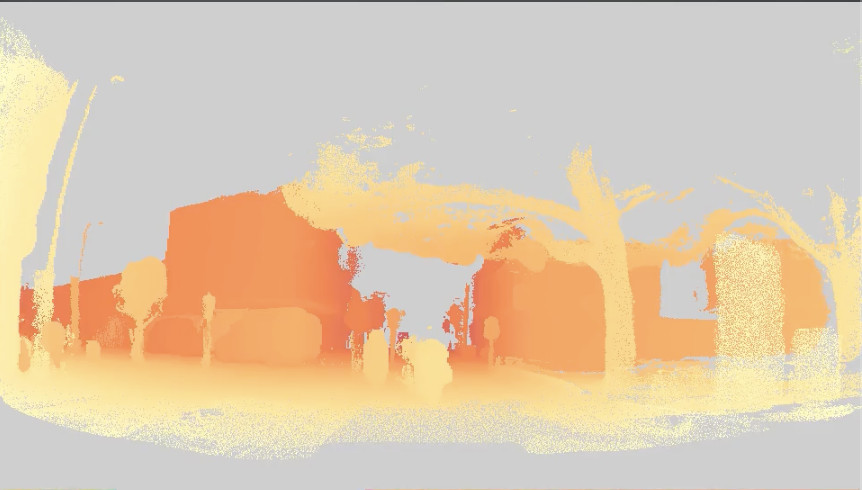}\\[-1pt]
  {\scriptsize Predicted depth}
\end{minipage}\hfill
\begin{minipage}[t]{.29\linewidth}\centering
  \includegraphics[width=\linewidth]{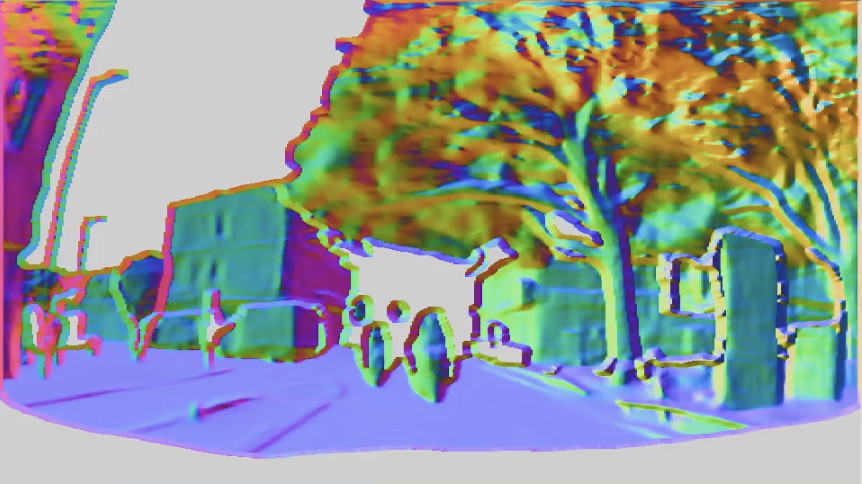}\\[-1pt]
  {\scriptsize Normal prior}
\end{minipage}\hfill
\begin{minipage}[t]{.29\linewidth}\centering
  \includegraphics[width=\linewidth]{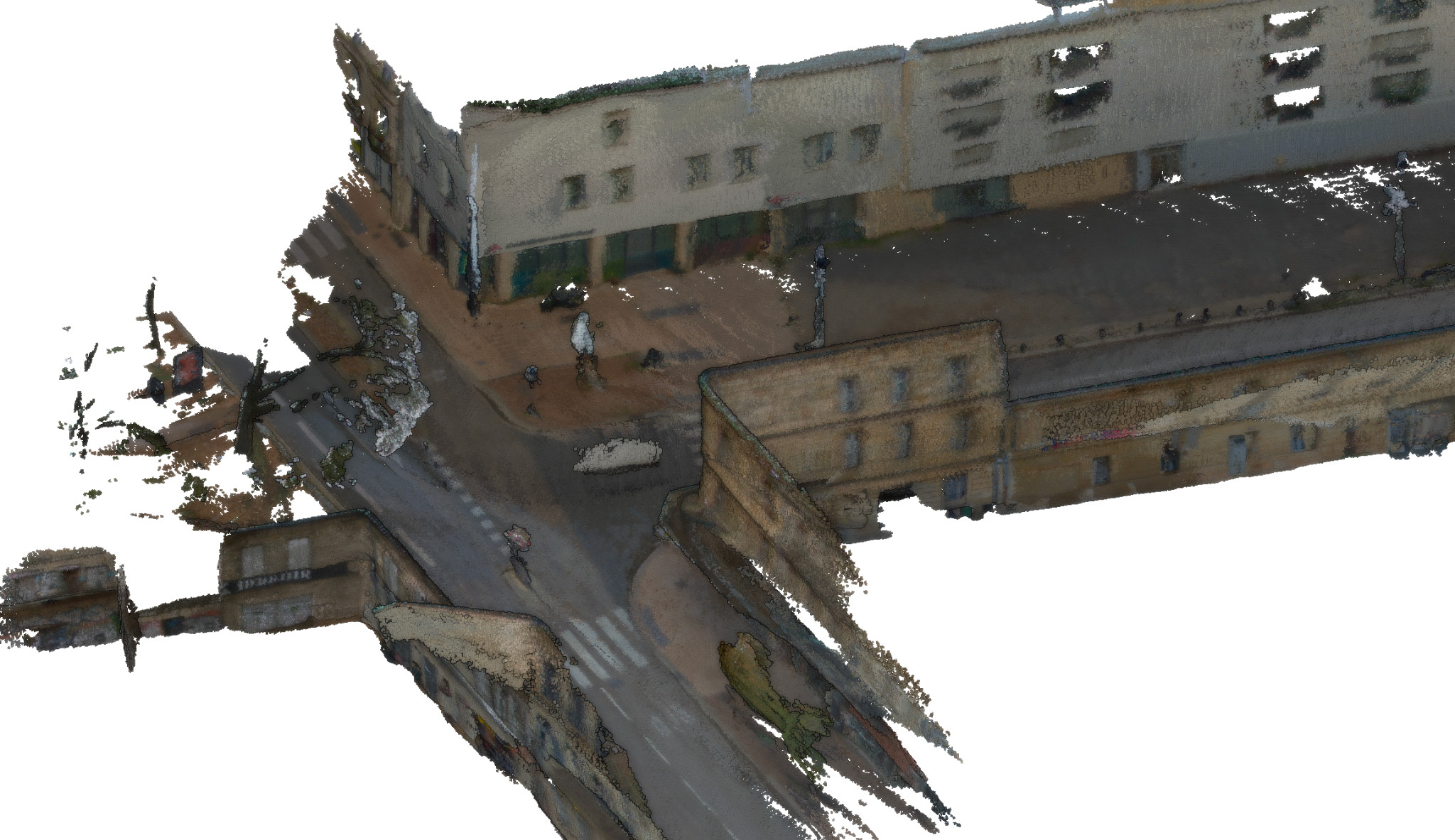}\\[-1pt]
  {\scriptsize Fused scene geometry}
\end{minipage}
\endgroup
\caption{\textbf{Foundation4D in lidarless NuRec.} Foundation4D and video-level processing supply the scene priors; NuRec then optimizes a Gaussian scene against the recorded RGB frames. Bottom: example depth, normal, and fused-geometry priors.}
\label{fig:lidarless_nurec_pipeline}
\end{figure}

The workflow in \cref{fig:lidarless_nurec_pipeline} takes multi-camera video and vehicle GPS/IMU trajectories as recorded inputs. Video-level processing assembles poses and depth predicted from images by \method into camera trajectories and a fused point cloud, which initializes reconstruction geometry. Detection and tracking provide oriented cuboids with persistent actor identities, separating moving objects from the static scene and associating their observations over time. These outputs replace priors previously obtained from LiDAR or separate reconstruction and perception pipelines. NuRec uses them to optimize a dynamic Gaussian scene against RGB observations for novel-view rendering~\cite{kerbl2023gaussiansplatting,wu2025dgut}.

\subsection{Reconstruction quality and efficiency}
\label{sec:application_quality}
\begin{figure}[!tp]
\centering
\begingroup
\setlength{\tabcolsep}{0pt}
\newcommand{\lfpanel}[2]{%
  \raisebox{-.5\height}{%
    \includegraphics[trim=#2,clip,width=.46\linewidth,height=.25875\linewidth]{assets/lidarless_nurec/#1}}}
\newcommand{\lfrowlabel}[1]{%
  \makebox[.035\linewidth][c]{%
    \raisebox{-.5\height}{\rotatebox{90}{\scriptsize #1}}}}
\begin{tabular}{@{}c@{\hspace{.015\linewidth}}c@{\hspace{.02\linewidth}}c@{}}
  & \small Detection and tracking priors & \small NuRec reconstruction\\[3pt]
  \lfrowlabel{w/o \method}
  & \lfpanel{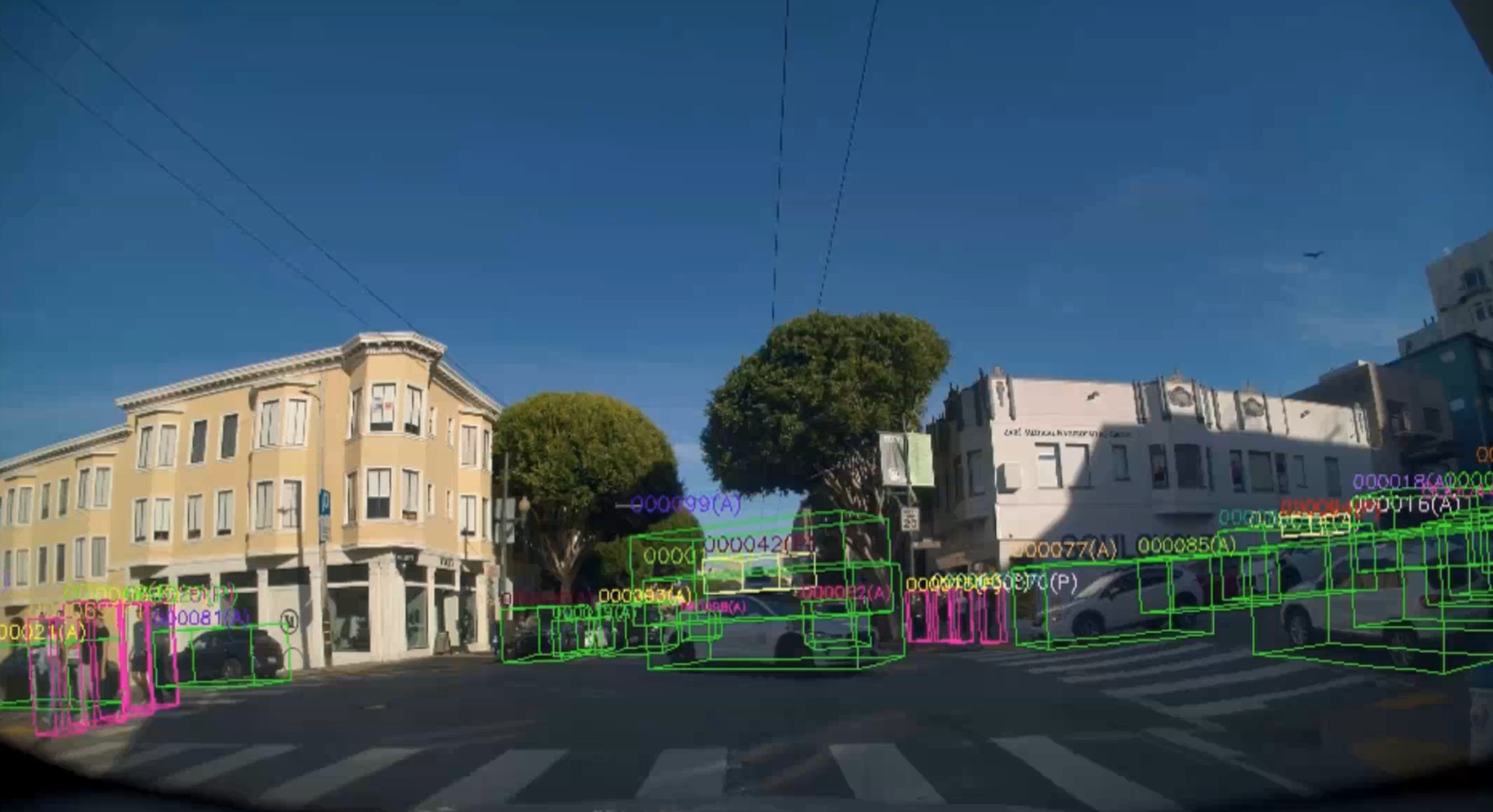}{33.7778bp 0bp 33.7778bp 0bp}
  & \lfpanel{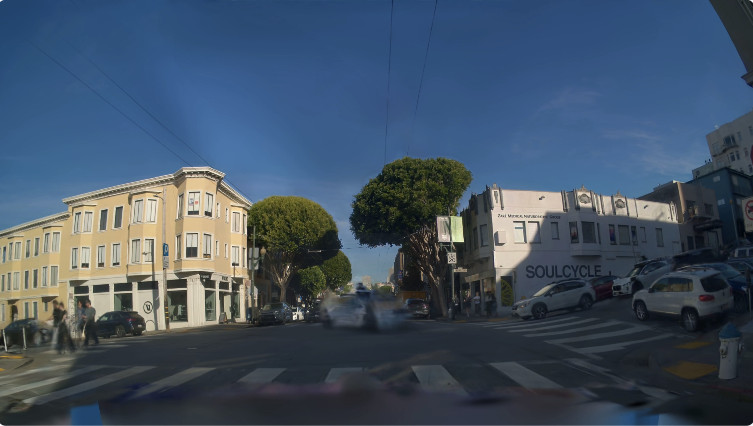}{0bp 1.21875bp 0bp 1.21875bp}\\
  \noalign{\vskip5pt}
  \lfrowlabel{w/ \method}
  & \lfpanel{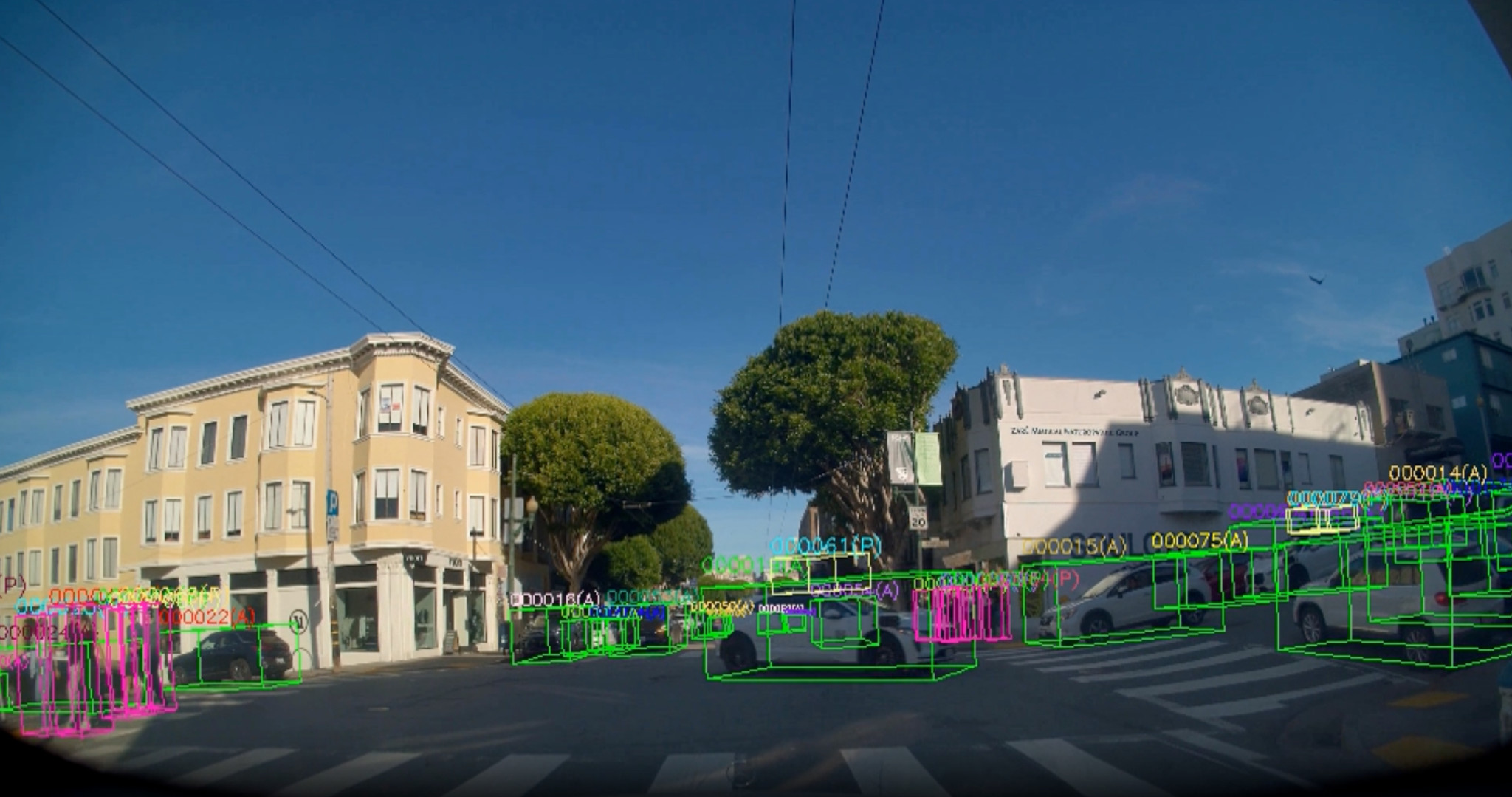}{64.8889bp 0bp 64.8889bp 0bp}
  & \lfpanel{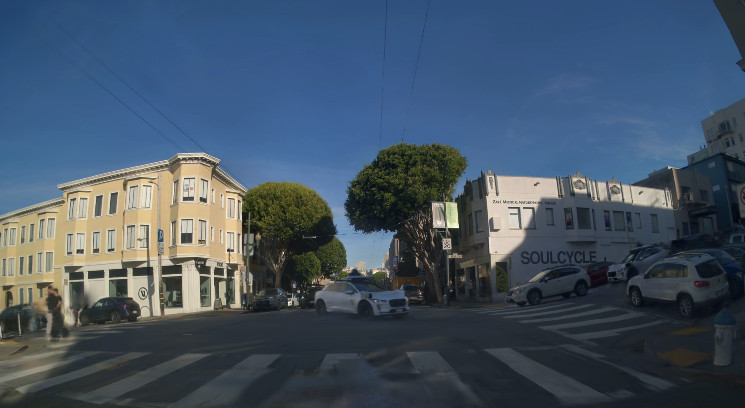}{9.833333bp 0bp 9.833333bp 0bp}
\end{tabular}
\endgroup
\caption{\textbf{NuRec reconstruction with and without \method predictions as priors.} Left: detection and tracking priors used for reconstruction; right: novel views of the optimized NuRec scene. The bottom row shows more consistent tracks and a more complete reconstruction of the crossing white vehicle. This comparison does not isolate individual model changes.}
\label{fig:lidarless_dynamic}
\end{figure}

The quality of the input priors affects how NuRec assigns image observations to scene components: inaccurate poses or inconsistent actor tracks can lead to misplaced geometry and ghosting. The example in \cref{fig:lidarless_dynamic} illustrates this connection between perception and reconstruction quality.
\cref{tab:lidarless_runtime} reports a reduction in end-to-end workflow turnaround from 16.0 to 3.7 hours per clip, a $4.3\times$ speedup. Camera-pose and depth estimation with fusion also speed up because feed-forward prediction replaces separate iterative reconstruction stages. These measurements describe the overall workflow improvement and do not isolate individual model optimizations.

\label{sec:application_runtime}

\begin{table}[!tp]
\centering
\small
\caption{\textbf{Workflow turnaround across releases.} Times are hours per clip, including startup and teardown. Camera configurations, software, and hardware differ across releases. Rows describe separate processing scopes and are not an additive breakdown.}
\label{tab:lidarless_runtime}
\begin{tabularx}{\linewidth}{@{}Xrrr@{}}
\toprule
Scope & \shortstack[r]{Without \method\\(hours/clip)} & \shortstack[r]{With \method\\(hours/clip)} & Speedup\\
\midrule
End-to-end lidarless NuRec & 16.0 & 3.7 & $4.3\times$\\
Camera pose estimation & 3.5 & 0.4 & $8.8\times$\\
Depth estimation and fusion & 8.0 & 1.0 & $8.0\times$\\
\bottomrule
\end{tabularx}
\par\vspace{3pt}
\end{table}

\section{Limitations and open questions}
\label{sec:limitations}

\paragraph{Unified model.}
Adding more tasks does not always improve performance. Geometry and occupancy trained together achieve $11.82\%$ occupancy mIoU, compared with $11.53\%$ for the three-task model that also includes detection (\cref{tab:joint_task_ablation}). Better task weighting and data sampling may help the model benefit more consistently from joint supervision, but testing new recipes is costly. For efficiency, our training schedule introduces Gaussian reconstruction after perception pretraining, deferring the cost of Gaussian decoding and rendering. Perception training took approximately 18 days, followed by another 9 days for Gaussian head training and joint fine-tuning. Although the final stage jointly optimizes the perception and Gaussian heads, we have not studied the benefits of training them together from the outset or systematically evaluated how geometry and rendering supervision affect each other.

\paragraph{Geometry.}
Depth errors increase at long range and around object boundaries (\cref{fig:geometry_pai_depth_ranges,fig:geometry_pai_far_range}). Distant objects and thin structures occupy few pixels, making their depth difficult to estimate reliably. Differences in camera capture times can also affect pose estimation: rig supervision can mix fixed camera calibration with ego motion between exposures. Interpolating the ego trajectory does not resolve this ambiguity.

\paragraph{Detection and tracking.}
Detection errors in crowded scenes and at long range also affect tracking. Missed objects and temporary confidence drops interrupt the observations available for association, leading to fragmented trajectories. Reliable long-video tracking therefore requires both more stable detections and better recovery across gaps in the observations.

\paragraph{Semantic occupancy.}
Occupancy accuracy decreases at long range, and rare classes are often missed or assigned incorrect semantic labels. Predictions also fluctuate between frames, particularly around object boundaries and thin structures. Improving the coverage of rare classes in the training data and encouraging consistent predictions across frames may help reduce these errors.

\section{Conclusion}
\label{sec:conclusion}

We presented \method, a unified model for perception and reconstruction from multi-camera driving video. Shared visual features support geometry, objects, semantic occupancy, and Gaussian reconstruction, while video-level alignment and tracking extend its predictions to long driving sequences. To support learning and evaluation, our annotation pipeline addresses sensor timing, motion, and visibility when generating dense depth and semantic occupancy labels. The released PAI--Foundation4D benchmark makes these annotations available for comparing geometry, detection, and occupancy methods on the same driving scenes.

Our evaluations show both the value of temporal observations and the limits of the current joint-training recipe. Longer input windows improve detection without retraining, and trajectory refinement improves pose estimation across windows. In our task-combination study, detection performs best with all three tasks, while occupancy performs best when paired with geometry. The predicted geometry and trajectories also provide priors for lidarless NuRec, demonstrating their use in reconstruction for simulation. Improving the balance between tasks and understanding the interaction between geometric and rendering supervision remain open directions.

\clearpage

\appendix
\section{Author list}
\label{AppA}

\textbf{Geometry:} Qunjie Zhou, Shengyu Huang, Mingxuan Liu, Bingxin Ke, Fei Xue

\noindent\textbf{Detection:} Guillem Braso, Sérgio Agostinho, Tim Meinhardt, Ilya Fedorov, Cristiano Saltori, Mingxuan Liu

\noindent\textbf{Occupancy:} Cristiano Saltori, Tim Meinhardt, Bingxin Ke, Maximilian Apodaca, Gang Huang

\noindent\textbf{Gaussian reconstruction:} Ning Xu, Shengyu Huang, Jiahui Huang

\noindent\textbf{Unified model:} Shengyu Huang, Guillem Braso, Tim Meinhardt

\noindent\textbf{Data:} Bingxin Ke, Aljoša Ošep, Michele Fenzi, Shengyu Huang, Mingxuan Liu, Sérgio Agostinho, Jiahui Huang

\noindent\textbf{Product team:} Emmanuel Attia, George Dong, Jin-Bey Yu, Kefei Ren, Keqiang Wu, Kuan-Ting Lee, Naveen Kumar Rai, Ning Xu, Tian Liu

\noindent\textbf{Product leadership:} Lei Zhang

\noindent\textbf{Leadership:} Laura Leal-Taix{\'e}

\section*{Acknowledgements}

We give special thanks to our product manager Martijn Tideman who worked tirelessly to help us bring this project to life. 
We thank Ning Xu and Jiaxin Zhang from the Dynamic autolabeling team for providing us with semantic occupancy annotations. 
We thank Hunter van Adelsberg and Xiangyu Chen for their contributions to the PAI dataset release. 
We thank Jiawei Ren, Zan Gojcic, Sanja Fidler, Nick Schneider and Matt Cragun, for the insightful discussions.

\clearpage
{
\small
\bibliographystyle{unsrtnat}
\bibliography{references}
}

\end{document}